\documentclass[final]{informs3}

\usepackage[autolanguage]{numprint}  
\usepackage{amsmath}
\usepackage{amssymb}
\usepackage{pgf}
\usepackage{tikz}
\usepackage{arydshln}
\usepackage{hyperref}
\usepackage{lscape}
\usepackage{verbatim}
\usepackage{ifthen}
\usepackage{listings}
\usepackage{moreverb}

\newboolean{TR}
\setboolean{TR}{true}

\ifthenelse{\boolean{TR}}{
\renewcommand{\theARTICLETOP}{}
\renewcommand{\theLRHSecondLine}{}
\renewcommand{\theRRHSecondLine}{}
}{ 
\renewcommand{\theARTICLETOP}{}
} 

\let\vec\relax
\DeclareMathAccent{\vec}{\mathord}{letters}{"7E}

\usepackage{accents}

\usetikzlibrary{backgrounds,arrows,calc}
\tikzset { domaine/.style 2 args={domain=#1:#2} } 

\newcommand*{\Cplusplus}{{C\nolinebreak[4]\hspace{-.05em}\raisebox{.4ex}{\tiny\bf ++}}}

\providecommand*{\Nset}{\mathbb{N}}            
\providecommand*{\Rset}{\mathbb{R}}            
\newcommand*{\Fset}{\mathbb{F}}               
\newcommand*{\Fsetsub}{\mathbb{F}^{\mathrm{sub}}} 

\newcommand*{\float}[2]{\ensuremath{\mathalpha{#1} \times \mathalpha{2^{#2}}}}
\newcommand*{\expf}[1]{e_{#1}}
\newcommand*{\lexpf}{\mathop{\mathrm{exp}}\nolimits}
\newcommand*{\midf}{\mathop{\mathrm{mid}}\nolimits}

\newcommand*{\feven}{\mathop{\mathrm{even}}\nolimits}
\newcommand*{\fodd}{\mathop{\mathrm{odd}}\nolimits}

\newcommand*{\fpzero}{+0}
\newcommand*{\fmzero}{-0}

\newcommand*{\fpinf}{+\infty} 
\newcommand*{\fminf}{-\infty} 

\newcommand*{\fmin}{f_\mathrm{min}}
\newcommand*{\fmax}{f_\mathrm{max}}
\newcommand*{\fnormin}{{f^\mathrm{nor}_\mathrm{min}}}
\newcommand*{\emin}{e_\mathrm{min}}
\newcommand*{\emax}{e_\mathrm{max}}

\newcommand{\st}{\mathrel{.}}
\newcommand{\itc}{\mathrel{:}}

\newcommand*{\fund}[3]{\mathord{#1}\colon#2\to#3}
\newcommand*{\pard}[3]{\mathord{#1}\colon#2\rightarrowtail#3}

\newcommand*{\ferrdown}[1]{\mathop{\Delta^{+}_{#1}}}
\newcommand*{\ferrup}[1]{\mathop{\Delta^{-}_{#1} }}

\newcommand*{\ulp}{\mathop{\mathrm{ulp}}\nolimits}

\newcommand*{\fadd}{\mathbin{\oplus}}
\newcommand*{\fsub}{\mathbin{\ominus}}
\newcommand*{\fmul}{\mathbin{\otimes}}
\newcommand*{\fdiv}{\mathbin{\oslash}}
\newcommand*{\fop}{\mathbin{\odot}}

\newcommand*{\rnear}{{\mathrm{n}}}

\newcommand*{\round}[2]{[#2]_{#1}}
\newcommand*{\roundnear}[1]{\round{\rnear}{#1}}
\newcommand*{\biground}[2]{\bigl[#2\bigr]_{#1}}
\newcommand*{\bigroundnear}[1]{\biground{\rnear}{#1}}
\newcommand*{\scaleround}[2]{\left[#2\right]_{#1}}
\newcommand*{\scaleroundnear}[1]{\scaleround{\rnear}{#1}}

\newcommand*{\var}[1]{\mathalpha{\texttt{\upshape{#1}}}} 
\newcommand*{\ub} [1]{\overline {\texttt{\upshape{#1}}}} 
\newcommand*{\lb} [1]{\underline{\texttt{\upshape{#1}}}} 

\newcommand*{\ulpmmax} [1]{\mathop{\bar{\delta}_{\mathord{#1}}}\nolimits}
\newcommand*{\ulpmmaxp}[1]{\mathop{\bar{\delta}'_{\mathord{#1}}}\nolimits}
\newcommand*{\ulpmmin} [1]{\mathop{\underaccent{\bar}{\delta}_{\mathord{#1}}}\nolimits}
\newcommand*{\ulpmminp} [1]{\mathop{\underaccent{\bar}{\delta}'_{\mathord{#1}}}\nolimits}
\newcommand*{\mufadd}[1]{\mathop{\mu_{\fadd}}\nolimits(#1)}
\newcommand*{\upperdiv}{\mathop{{\tilde{\delta}'_{\fdiv}}}\nolimits}
\newcommand*{\lowerdiv}{\mathop{{\underline{\tilde{\delta}}}'_{\fdiv}}\nolimits}

\usepackage[square,sort,comma,numbers]{natbib}
 \bibpunct[, ]{(}{)}{,}{a}{}{,}%
 \def\bibfont{\small}%
 \def\newblock{\ }%
\TheoremsNumberedThrough     

\EquationsNumberedThrough    

\begin{document}


\RUNAUTHOR{Bagnara et al.}

\RUNTITLE{Exploiting Binary Floating-Point Representations
          for Constraint Propagation}

\ifthenelse{\boolean{TR}}{
\TITLE{Exploiting Binary Floating-Point Representations \\
  for Constraint Propagation: \\
  The Complete Unabridged Version}
}{ 
\TITLE{Exploiting Binary Floating-Point Representations \\
       for Constraint Propagation}
} 

\ARTICLEAUTHORS{%
\AUTHOR{Roberto Bagnara}
\AFF{BUGSENG srl and Dept.\ of Mathematics and Computer Science,
     University of Parma, Italy, \\
     \EMAIL{\url{bagnara@cs.unipr.it}},
     \URL{\url{http://www.cs.unipr.it/~bagnara}}}
\AUTHOR{Matthieu Carlier}
\AFF{INRIA Rennes Bretagne Atlantique, France}
\AUTHOR{Roberta Gori}
\AFF{Dept.\ of Computer Science, University of Pisa, Italy, \\
     \EMAIL{\url{gori@di.unipi.it}},
     \URL{\url{http://www.di.unipi.it/~gori}}}
\AUTHOR{Arnaud Gotlieb}
\AFF{Certus Software V\&V Center, SIMULA Research Laboratory, Norway, \\
     \EMAIL{\url{arnaud@simula.no}},
     \URL{\url{http://simula.no/people/arnaud}}}
} 

\ABSTRACT{%
Floating-point computations are quickly finding their way in the
design of safety- and mission-critical systems, despite the fact
that designing floating-point algorithms is significantly
more difficult than designing integer algorithms.
For this reason, verification and validation of floating-point
computations is a hot research topic.
An important verification technique, especially in some industrial
sectors, is testing.
However, generating test data for floating-point intensive programs
proved to be a challenging problem.
Existing approaches usually resort to random or search-based test data
generation, but without symbolic reasoning it is almost impossible to
generate test inputs that execute complex paths controlled by
floating-point computations.
Moreover, as constraint solvers over the reals or the rationals do not
natively support the handling of rounding errors, the need arises for
efficient constraint solvers over floating-point domains.
In this paper, we present and fully justify improved algorithms for
the propagation of arithmetic IEEE~754 binary floating-point constraints.
The key point of these algorithms is a generalization of an idea
by B.~Marre and C.~Michel that exploits a property of the representation
of floating-point numbers.
}%


\KEYWORDS{software verification; testing; floating-point numbers;
          constraint solving}

\maketitle

\section{Introduction}
\label{sec:introduction}

During the last decade, the use of floating-point computations in the
design of critical systems has become increasingly acceptable.
Even in the civil and military avionics domain, which are among the most
critical domains for software, floating-point numbers are now seen as a
sufficiently-safe, faster and cheaper alternative to fixed-point arithmetic.
To the point that, in modern avionics, floating-point is the norm rather
than the exception \citep{BurdyDL12}.

Acceptance of floating-point computations in the design of critical systems
took a long time.
In fact, rounding errors can cause subtle bugs which are often missed by non
experts
\citep{Monniaux08}, and can lead to catastrophic failures.
For instance, during the first Persian Gulf War, the failure of a Patriot
missile battery in Dhahran was traced to an accumulating rounding error
in the continuous execution of tracking and guidance software:
this failure prevented the interception of an Iraqi Scud that hit the
barracks in Dhahran, Saudi Arabia, killing 28 US soldiers \citep{Skeel92}.
A careful analysis of this failure revealed that, even though the
rounding error obtained at each step of the floating-point computation
was very small, the propagation during a long loop-iterating path
could lead to dramatic imprecision.

Adoption of floating-point computations in critical systems involves
the use of thorough unit testing procedures that are able to exercise
complex chains of floating-point operations.
In particular, a popular practice among software engineers in charge of
the testing of floating-point-intensive computations consists in executing
carefully chosen loop-iterating paths in programs.
They usually pay more attention to the paths that are most likely to
expose the system to
unstable numerical computations.\footnote{A computation can be called
\emph{numerically stable} if it can be proven not to magnify
approximation errors.  It can be called \emph{(potentially) unstable}
otherwise.}
For critical systems, a complementary requirement is to demonstrate
the infeasibility of selected paths, in order to convince a third-party
certification authority that certain unsafe behaviors of the systems
cannot be reached.
As a consequence, software engineers face two difficult problems:
\begin{enumerate}
\item How to accurately predict the expected output of a given floating-point
      computation?\footnote{This is the well-known \emph{oracle problem}
      \citep[see][]{Weyuker82}.}
\item How to find a test input that is able to exercise a given path,
      the execution of which depends on the results of floating-point
      computations, or to guarantee that such a path is infeasible?
\end{enumerate}

The first problem has been well addressed in the literature
\citep{Kuliamin10} through several techniques.
\citet{AmmannK88} report on a technique known as the \emph{data
diversity} approach, which uses multiple related program executions
of a program to check their results.
\emph{Metamorphic testing} \citep{ChanCCLY98}
generalizes this technique by using known
numerical relations of the function implemented by a program to
check the results of two or more executions.
\citet{Goubault01} proposes using the \emph{abstract interpretation}
framework \citep{CousotC77} to estimate the deviation of the
floating-point results with respect to an interpretation over the reals.
\citet{ScottJDC07} propose using a probabilistic approach to estimate round-off
error propagation.
More recently, \citet{TangBLS10} propose to exploit
perturbation techniques to evaluate the stability of a numerical
program.
In addition to these approaches, it is possible
to use a (partial) specification, a prototype or an old implementation
in order to predict the results for a new implementation.

In contrast, the second problem received only little attention.
Beyond the seminal work of \citet{MillerS76},
who proposed to guide the search of floating-point
inputs to execute a selected path, few approaches try to exactly
reason over floating-point computations.
The work of \citet{MillerS76} paved the way to the development of
\emph{search-based test data generation} techniques, which consist
in searching test inputs by minimizing a cost function, evaluating
the distance between the currently executed path
and a targeted selected path \citep{Korel90a,McMinn04,LakhotiaHG10}.
Although these techniques enable quick and efficient coverage of
testing criteria such as ``all decisions,'' they are unfortunately
sensitive to the rounding errors incurred in the computation of the
branch distance \citep{Arcuri09}.
Moreover, search-based test data generation cannot be used to study
path feasibility, i.e., to decide whether a possible execution path
involving floating-point computations is feasible or not in the
program.
In addition, these techniques can be stuck in local minima without
being able to provide a meaningful result \citep{Arcuri09}.
An approach to tackle these problems combines program execution and
symbolic reasoning \citep{GodefroidKS05},
and requires solving constraints over floating-point
numbers in order to generate test inputs that exercise a selected behavior
of the program under test.
However, solving floating-point constraints is hard and requires dedicated
filtering algorithms \citep{MichelRL01,Michel02}.
According to our knowledge, this approach is currently implemented
in four solvers only:
ECLAIR\footnote{\url{http://bugseng.com/products/eclair}},
FPCS~\citep{BlancBGJ+06},
FPSE\footnote{\url{http://www.irisa.fr/celtique/carlier/fpse.html}}
\citep{BotellaGM06},
and GATeL, a test data generator for Lustre programs
\citep{MarreB05}.
It is worth noticing that existing constraint solvers dedicated to
continuous domains (such as, e.g., RealPaver \citep{GranvilliersB06},
IBEX and Quimper \citep{ChabertJ09} or ICOS \citep{Lebbah09})
correctly handle
real or rational computations, but they cannot preserve the
solutions of constraints over floating-point computations in all
cases (see Section~\ref{sec:discussion} for more on this subject).
``Surprising'' properties of floating-point computations such as
absorption and cancellation \citep{Goldberg91} show that the rounding
operations can severely compromise the preservation of the computation
semantics between the reals and the floats.
\begin{example}
Consider the C~functions \verb+f1+ and \verb+f2+:
\begin{lstlisting}
float f1(float x) {                                          float f2(float x) {
  float y = 1.0e12F;                                           float y = 1.0e12F;
  if (x < 10000.0F)                                            if (x > 0.0F)
     z = x + y;                                                  z = x + y;
  if (z > y)                                                   if(z == y)
    ...                                                          ...
\end{lstlisting}
For both functions, let us ask the question whether the paths
traversing lines 2-3-4-5-6 are feasible.
The condition that must be satisfied in order for a certain path to be
traversed is called \emph{path condition}.
For \verb+f1+, the path conditions
$\mathtt{x} < 10000.0$ and $\mathtt{x} + \mathrm{1.0e12} > \mathrm{1.0e12}$,
which on the reals are equivalent to $\mathtt{x} \in (0, 10000)$
whereas on the floats they have no solution.
Conversely, for \verb+f2+ the path conditions are
$\mathtt{x} > 0.0$ and $\mathtt{x} + \mathrm{1.0e12} = \mathrm{1.0e12}$, which
have no solutions on the reals but are satisfied by all IEEE~754
single precision floating-point numbers in the range $(0, 32767.99\cdots)$.
\end{example}

\subsection{A Real-World Example}

\begin{lstlisting}[float,floatplacement=tp,label=lst:cam,caption=Code excerpted from a real-world avionic library]
#define MAX_PPRZ 9600
#define MIN_PPRZ -MAX_PPRZ

#ifndef CAM_PAN_MAX
#define CAM_PAN_MAX 90
#endif
#ifndef CAM_PAN_MIN
#define CAM_PAN_MIN -90
#endif
#define M_PI 3.14159265358979323846
#define RadOfDeg(x) ((x) * (M_PI/180.))

#ifdef CAM_PAN_NEUTRAL
#if (CAM_PAN_MAX == CAM_PAN_NEUTRAL)
#error CAM_PAN_MAX has to be different from CAM_PAN_NEUTRAL
#endif
#if (CAM_PAN_NEUTRAL == CAM_PAN_MIN)
#error CAM_PAN_MIN has to be different from CAM_PAN_NEUTRAL
#endif
#endif

float cam_pan_c;

void cam_angles( void ) {
  float cam_pan = 0;
  if (cam_pan_c > RadOfDeg(CAM_PAN_MAX)) {
    cam_pan_c = RadOfDeg(CAM_PAN_MAX);
  } else {
    if (cam_pan_c < RadOfDeg(CAM_PAN_MIN))
      cam_pan_c = RadOfDeg(CAM_PAN_MIN);
  }

#ifdef CAM_PAN_NEUTRAL
  float pan_diff = cam_pan_c - RadOfDeg(CAM_PAN_NEUTRAL);
  if (pan_diff > 0)
    cam_pan = MAX_PPRZ * (pan_diff / (RadOfDeg(CAM_PAN_MAX - CAM_PAN_NEUTRAL)));
  else
    cam_pan = MIN_PPRZ * (pan_diff / (RadOfDeg(CAM_PAN_MIN - CAM_PAN_NEUTRAL)));
#else
  cam_pan = ((float)RadOfDeg(cam_pan_c - CAM_PAN_MIN))
              * ((float)MAX_PPRZ / (float)RadOfDeg(CAM_PAN_MAX-CAM_PAN_MIN) );
#endif

  if (cam_pan < MIN_PPRZ)
    cam_pan = MIN_PPRZ;
  else if (cam_pan > MAX_PPRZ)
    cam_pan = MAX_PPRZ;
}
\end{lstlisting}
To illustrate the concrete problem raised by floating-point computations
in program verification settings,
consider the code depicted in Listing~\ref{lst:cam}.
It is a somewhat reduced version of a real-world example extracted
from a critical embedded system.\footnote{The original source
code is available at \url{http://paparazzi.enac.fr}, file
\url{sw/airborne/modules/cam_control/cam.c}, last checked on November 29, 2013.}
In order to gain confidence in this code,
a test-suite should be created that contains
enough test cases to achieve a specified level of coverage.
The basic coverage criterion is ``all statements'', and
prescribes that each statement is reached at least once by at least
one test.\footnote{There exist more sophisticate and, correspondingly,
more challenging coverage criteria, such as the already-mentioned
``all decisions'' and \emph{Modified Condition Decision Coverage}
\citep[MCDC, see][]{AmmannOH03}.}
For each statement, a set of constraints is defined that encodes the
reachability of the statement and then solution is attempted:
if one solution is found, then such a solution, projected on the
explicit inputs (read parameters) and the implicit inputs (read global
variables) of the function, constitutes the input part of a test case;
if it is determined that a solution does not exist, then the statement
is dead code;
if the solution process causes a timeout, then we don't know.
For example, if the \verb"CAM_PAN_NEUTRAL" is defined to expand to the integer
literal \verb"5" (or, for that matter, \verb"45" or many other values),
then we can prove that the statements in lines~45 and~47 are
unreachable.\footnote{All the experiments
mentioned in this paper have been conducted using the ECLAIR system.}
The presence of dead code is not acceptable for several
industry standards such as MISRA C \citep{MISRA-C-2012},
MISRA \Cplusplus{} \citep{MISRA-CPP-2008},
and JSF \Cplusplus{} \citep{JSF-CPP-2005}.

Another application of the same technology is the proof of absence of
run-time anomalies, such as overflows or the unwanted generation of infinities.
For each operation possibly leading to such an anomaly, a constraint system
is set up that encodes the conditions under which the anomaly takes place.
A solution is then searched:
if it is found, then we have the proof that the code is unsafe;
if it can be determined that no solution exists, then we know the code
is safe;
otherwise we don't know.
For the code of Listing~\ref{lst:cam},
if the \verb"CAM_PAN_NEUTRAL" is defined \verb"5" or \verb"45",
then we can prove that no run-time anomaly is possible, whatever
is the value of variable \verb"cam_pan_c" when the function
is invoked.

Now, let us take another point of view and consider that
the macro \verb"CAM_PAN_NEUTRAL" is not defined in the same file,
as it is a configuration parameter.  Its definition is (partially)
validated by means of preprocessor directives as shown in the listing
at lines 13--20: these directives enough to protect against dangerous
definitions of \verb"CAM_PAN_NEUTRAL"?
We can provide an answer to this question by treating \verb"CAM_PAN_NEUTRAL"
as a variable of any type that is compatible with its uses in the code.
This way we discover that, if \verb"CAM_PAN_NEUTRAL" is defined to expand
to, e.g., \verb"-2147483558", then we will have an overflow in line~36 on
a 32-bit machine.  Most compilers will catch this particular mistake,
but this will not be the case if someone, someday, defines
\verb"CAM_PAN_NEUTRAL" as, e.g.,
\verb"+0x1ca5dc14c57550.p81" (roughly $1.94967\cdot 10^{40}$):
then in line 34 an infinity will be generated, something
that in the aviation and other industries is unacceptable.
One might also wonder whether one can define \verb"CAM_PAN_NEUTRAL"
as a double precision floating-point literal
so that the denominator of divisions in lines~36 and~38 can be so small
to cause an overflow: constraint solving over floating-point numbers
is able to answer negatively to this question.

\subsection{Contribution and Plan of the Paper}

A promising approach to improve the filtering capabilities of
constraints over floating-point variables consists in using some
peculiar numerical properties of floating-point numbers. For linear
constraints, this led to a relaxation technique where floating-point
numbers and constraints are converted into constraints over the reals
by using linear programming approaches \citep{BelaidMR12}.
For interval-based consistency approaches, \citet{MarreM10} identified
a property of the representation of floating-point numbers and proposed
to exploit it in filtering algorithms for addition and subtraction
constraints.
\citet{CarlierG11} proposed a reformulation of the Marre-Michel property
in terms of ``filtering by maximum ULP'' (\emph{Units in the Last Place})
that is generalizable to multiplication and division constraints.

\citet{BagnaraCGG13ICST} addressed the question of whether
the Marre-Michel property can be useful for the automatic solution of
realistic test input generation problems: they sketched (without proofs)
a reformulation and correction of the filtering algorithm proposed
in~\citep{MarreM10}, along with a uniform framework that generalizes
the property identified by Marre and Michel to the case of multiplication
and division.  Most importantly, \citep{BagnaraCGG13ICST}
presented the implementation of filtering by maximum ULP in FPSE
and some of its critical design choices, and an experimental evaluation
on constraint systems that
have been extracted from programs engaging into intensive floating-point
computations.
These results show that the Marre-Michel property
and its generalization are effective, practical properties for solving
constraints over the floats with an acceptable overhead.
The experiments reported in \citep{BagnaraCGG13ICST} showed that
improvement of filtering procedures with these techniques brings speedups
of the overall constraint solving process that can be substantial
(we have observed up to an order of magnitude);  in the cases
where such techniques do not allow significant extra-pruning,
the slowdowns are always very modest (up to a few percent on
the overall solution time).

The present paper is, on the one hand, the theoretical counterpart of
\citep{BagnaraCGG13ICST} in that all the results are thoroughly proved;
on the other hand, this paper generalizes and extends
\citep{BagnaraCGG13ICST} as far as the handling of subnormals and
floating-point division are concerned.
More precisely, the contributions of the paper are:
\begin{enumerate}
\item a uniform framework for filtering by maximum ULP is thoroughly defined
      and justified;
\item the framework is general enough to encompass all floating-point
      arithmetic operations and subnormals (the latter are not treated
      in~\citep{BagnaraCGG13ICST});
\item a second indirect projection by maximum ULP for division
      (not present in any previous work);
\item all algorithms only use floating-point machine arithmetic operations
      on the same formats used by the analyzed computations.
\end{enumerate}

The plan of the paper is as follows. Next section presents
the IEEE~754 standard of binary floating-point numbers and introduces the
notions and notations used throughout the paper.
Section~\ref{sec:background} recalls the basic principles of
interval-based consistency techniques over floating-point variables
and constraints.
Section~\ref{sec:max-ulp} presents our generalization of the Marre-Michel
property along with a precise definition and motivation of all the required
algorithms.
Section~\ref{sec:discussion} discusses related work.
\ifthenelse{\boolean{TR}}{
Section~\ref{sec:conclusion} concludes the main body of the paper.
The most technical proofs are available in the Appendix.
}{ 
Section~\ref{sec:conclusion} concludes.
The most technical proofs are available in the
Online Supplement to this paper \citep{BagnaraCGG15IJOC-OS}.
} 

\section{Preliminaries}

In this section we recall some preliminary concepts and
introduce the used notation.

\subsection{IEEE~754}

This section recalls the arithmetic model specified by the IEEE~754
standard for binary floating-point arithmetic \citep{IEEE-754-2008}.
Note that, although the IEEE~754 standard also specifies formats and
methods for decimal floating-point arithmetic, in this paper we only
deal with binary floating-point arithmetic.

IEEE~754 binary floating-point formats are uniquely identified by
quantities:
$p \in \Nset$, the number of significant digits (precision);
$\emax \in \Nset$, the maximum exponent;
$-\emin \in \Nset$, the minimum exponent.\footnote{Note that,
although the IEEE~754 formats have $\emin = 1 - \emax$,
we never use this property and decided to keep the extra-generality,
which might be useful to accommodate other formats.}
The \emph{single precision} format has $p = 24$ and $\emax = 127$,
the \emph{double precision} format has $p = 53$ and $\emax = 1023$
(IEEE~754 also defines extended precision formats).
A finite, non-zero IEEE~754 floating-point number $z$ has the form
$(-1)^s\float{b_1.m}{e}$ where
$s$ is the \emph{sign bit},
$b_1$ is the \emph{hidden bit},
$m$ is the $(p-1)$-bit \emph{significand} and
the \emph{exponent} $e$ is also denoted by $\expf{z}$ or $\lexpf(z)$.
Hence the number is positive when $s = 0$ and negative when $s = 1$.
$b_1$ is termed ``hidden bit'' because in the
\emph{binary interchange format encodings} it is not explicitly
represented, its value being encoded in the exponent \citep{IEEE-754-2008}.

Each format defines several classes of numbers: normal numbers,
subnormal numbers, signed zeros, infinities and NaNs (\emph{Not a Number}).
The smallest positive \emph{normal} floating-point number is
$\fnormin = \float{1.0\cdots0}{\emin} = 2^{\emin}$
and the largest is $\fmax = \float{1.1\cdots 1}{\emax} = 2^{\emax}(2 - 2^{1-p})$;
normal numbers have the hidden bit $b_1 = 1$.
The non-zero
floating-point numbers whose absolute value is less than $2^{\emin}$
are called \emph{subnormals}: they always have exponent equal to
$\emin$ and fewer than $p$ significant digits as their hidden bit is
$b_1 = 0$.
Every finite floating-point number is an integral multiple of
the smallest subnormal
$\fmin = \float{0.0\cdots01}{\emin} = 2^{\emin + 1 - p}$.
There are two infinities, denoted by $\fpinf$ and $\fminf$,
and two \emph{signed zeros}, denoted by $\fpzero$ and $\fmzero$:
they allow some algebraic properties to be maintained
\citep{Goldberg91}.\footnote{Examples of such properties are
$\sqrt{1/z} = 1/\sqrt{z}$ and $1/(1/x) = x$ for $x = \pm\infty$.}
NaNs are used to represent the results of invalid computations such as
a division of two infinities or a subtraction of infinities with the
same sign: they allow the program execution to continue without being
halted by an exception.

IEEE~754 defines five rounding directions:
toward negative infinity (\emph{roundTowardNegative} or, briefly, \emph{down}),
toward positive infinity (\emph{roundTowardPositive}, a.k.a.\ \emph{up}),
toward zero (\emph{roundTowardZero}, a.k.a.\ \emph{chop})
and toward the nearest representable value (a.k.a.\ \emph{near});
the latter comes in two flavors that depend on
different tie-break rules for numbers exactly halfway
between two representable numbers:
\emph{roundTiesToEven} (a.k.a.\ \emph{tail-to-even}) or
\emph{roundTiesToAway} (a.k.a.\ \emph{tail-to-away}) in which
values with even significand or values away from zero are preferred,
respectively.
This paper is only concerned with roundTiesToEven, which
is, by far, the most widely used.
The roundTiesToEven value of a real number $x$ will be
denoted by $\roundnear{x}$.

The most important requirement of
IEEE~754 arithmetic is the accuracy of floating-point computations:
add, subtract, multiply, divide, square root, remainder,
conversion and comparison operations must deliver to their destination
the exact result rounded as per the rounding mode in effect and the
format of the destination.
It is said that these operations are ``correctly rounded.''

The accuracy requirement of IEEE~754 can still surprise the average programmer:
for example the single precision, round-to-nearest addition of
$999999995904$ and $10000$ (both numbers can be exactly represented)
gives $999999995904$, i.e., the second operand is absorbed.
The maximum error committed by representing a real number with a
floating-point number under some rounding mode can be expressed in terms
of the function
$\fund{\ulp}{\Rset}{\Rset}$ \citep{Muller05}.
Its value on $1.0$ is about $10^{-7}$ for the single precision format.

\subsection{Notation}
The set of real numbers is denoted by $\Rset$ while
$\Fset_{p,\emax}$ denotes a subset of the binary floating-point numbers,
defined from a given IEEE~754 format, which
includes the infinities $\fminf$ and $\fpinf$,
the signed zeros $+0$ and $-0$, but
neither subnormal numbers nor NaNs.
Subnormals are introduced in the set
\(
  \Fsetsub_{p,\emax}
    = \Fset_{p,\emax}
        \cup \bigl\{\,
                (-1)^s\float{0.m}{\emin}
             \bigm|
                s \in \{0,1\}, m \neq 0
             \,\bigr\}
\).
In some cases, the exposition can be much simplified by allowing the
$\emax$ of $\Fset_{p,\emax}$ to be $\infty$, i.e., by considering an
idealized set of floats where the exponent is unbounded.
Among the advantages is the fact that subnormals in $\Fsetsub_{p,\emax}$
can be represented as normal floating-point numbers in $\Fset_{p,\infty}$.
Given a set of floating-point numbers $\Fset$, $\Fset^+$
denotes the ``non-negative''
subset of $\Fset$, i.e., with $s = 0$.

For a finite, non-zero floating-point number $x$, we will write $\feven(x)$
(resp., $\fodd(x)$)
to signify that the least significant digit of $x$'s significand
is~$0$ (resp.,~$1$).

When the format is clear from the context, a real decimal constant
(such as $10^{12}$) denotes the corresponding roundTiesToEven
floating-point value (i.e., $999999995904$ for $10^{12}$).

Henceforth, for $x \in \Rset$, $x^+$ (resp., $x^-$)
denotes the smallest (resp., greatest)
floating-point number strictly greater (resp., smaller) than $x$
with respect to the considered IEEE~754 format.
Of course, we have $\fmax^+ = \fpinf$ and $(-\fmax)^- = \fminf$.

Binary arithmetic operations over the floats will be denoted by
$\fadd$, $\fsub$, $\fmul$ and $\fdiv$, corresponding to $+$, $-$, $\cdot$
and $/$ over the reals, respectively.
According to IEEE~754, they are defined, under roundTiesToEven, by
\begin{align*}
  x \fadd y &= \roundnear{x + y},       & x \fsub y &= \roundnear{x - y}, \\
  x \fmul y &= \roundnear{x \cdot y},   & x \fdiv y &= \roundnear{x / y}.
\end{align*}
As IEEE~754 floating-point numbers are closed under negation,
we denote the negation of $x \in \Fsetsub_{p,\emax}$ simply by $-x$.
Note that negation is a bijection.
The symbol $\fop$ denotes any of $\fadd$, $\fsub$, $\fmul$ or $\fdiv$.
A floating-point variable $\var{x}$ is associated with an interval
of possible floating-point values;
we will write $\var{x} \in [\lb{x}, \ub{x}]$,
where $\lb{x}$ and $\ub{x}$ denote the smallest and greatest value of
the interval, $\lb{x} \leq \ub{x}$ and either $\lb{x} \neq \fpzero$
or $\ub{x} \neq \fmzero$.
Note that $[\fpzero, \fmzero]$ is not an interval,
whereas $[\fmzero, \fpzero]$ is the interval denoting the
set of floating-point numbers $\{ \fmzero, \fpzero \}$.

\section{Background on Constraint Solving over Floating-Point Variables}
\label{sec:background}

In this section, we briefly recall the basic principles of
interval-based consistency techniques over floating-point variables
and constraints.

\subsection{Interval-based Consistency on Arithmetic Constraints}

\begin{figure}
\begin{center}
\footnotesize
\renewcommand*\arraystretch{1.5}
\begin{tabular}{>{$}c<{$}|>{$}c<{$}}
  \var{z} = \var{x} \fadd \var{y}
    &\var{z} = \var{x} \fsub \var{y} \\
\hline
\begin{aligned}
\ub{z} &= \ub{x} \fadd \ub{y},
  & \text{(direct)}
  \\
\lb{z} &= \lb{x} \fadd \lb{y} &\\
\ub{x} &= \midf(\ub{z}, \ub{z}^{+}) \fsub \lb{y}
  & \text{(1\textsuperscript{st} indirect)}
  \\
\lb{x} &= \midf(\lb{z}, \lb{z}^{-}) \fsub \ub{y}
  \\
\ub{y} &= \midf(\ub{z}, \ub{z}^{+}) \fsub \lb{x}
  & \text{(2\textsuperscript{nd} indirect)}
  \\
\lb{y} &= \midf(\lb{z}, \lb{z}^{-}) \fsub \ub{x}
\end{aligned}
&
\begin{aligned}
\ub{z} &= \ub{x} \fsub \lb{y},
  & \text{(direct)}
  \\
\lb{z} &= \lb{x} \fsub \ub{y} & \\
\ub{x} &= \midf(\ub{z}, \ub{z}^{+}) \fadd \ub{y}
  & \text{(1\textsuperscript{st} indirect)}
  \\
\lb{x} &= \midf(\lb{z}, \lb{z}^{-}) \fadd \lb{y}
  \\
\ub{y} &= \ub{x} \fsub \textnormal{mid}(\lb{z}, \lb{z}^{-})
  & \text{(2\textsuperscript{nd} indirect)}
  \\
\lb{y} &= \lb{x} \fsub \textnormal{mid}(\ub{z}, \ub{z}^{+})
\end{aligned}
\end{tabular}
\medskip
\end{center}
\caption{Formulas for direct/indirect projections of addition/subtraction}
\label{fig:add-sub-projections}
\end{figure}

Program analysis usually starts with the generation of an intermediate code
representation in a form called \emph{three-address code} (TAC).
In this form, complex arithmetic expressions and assignments
are decomposed into sequences of assignment instructions of the form
\[
  \mathtt{result}
    \mathrel{:=}
      \mathtt{operand}_1 \,\mathbin{\mathtt{operator}} \,\mathtt{operand}_2.
\]
A further refinement consists in the computation of the
\emph{static single assignment form} (SSA) whereby,
labeling each assigned variable with a fresh name,
assignments can be considered as if they were equality constraints.
For example, the TAC form of the floating-point assignment
$\mathtt{z} \mathrel{:=} \mathtt{z}*\mathtt{z} + \mathtt{z}$
is
\(
  \mathtt{t} \mathrel{:=} \mathtt{z}*\mathtt{z}; \;
  \mathtt{z} \mathrel{:=} \mathtt{t} + \mathtt{z}
\),
which in SSA form becomes
\(
  \mathtt{t}_1 \mathrel{:=} \mathtt{z}_1*\mathtt{z}_1; \;
  \mathtt{z}_2 \mathrel{:=} \mathtt{t}_1 + \mathtt{z}_1
\),
which, in turn, can be regarded as the conjunction
of the constraints $t_1 = z_1 \fmul z_1$ and $z_2 = t_1 \fadd z_1$.

In an interval-based consistency approach to constraint solving over the
floats, constraints are used to iteratively narrow the intervals
associated with each variable:
this process is called \emph{filtering}.
A \emph{projection} is a function that,
given a constraint and the intervals associated with two of the variables
occurring in it, computes a possibly refined interval\footnote{That is,
tighter than the original interval.}
for the third
variable (the projection is said to be \emph{over} the third variable).
Taking $z_2 = t_1 \fadd z_1$ as an example, the projection over $z_2$
is called \emph{direct projection} (it goes in the same sense of the
TAC assignment it comes from), while the projections over $t_1$ and $z_1$
are called \emph{indirect projections}.\footnote{Note that direct and
indirect projections are idempotent, as their inputs and outputs do not
intersect.  Consider $z = x \fadd y$: direct projection propagates
information on $x$ and $y$ onto $z$, and doing it twice in a row
would not enable any further inference.  Likewise for indirect projections,
which propagate information on $z$ onto $x$ and $y$.}
Note that, for constraint propagation, both direct and indirect projections
are applied in order to refine the intervals for $t_1$, $z_1$ and $z_2$.
In this paper we propose new filtering algorithms
for improving indirect projections.

A projection is called \emph{optimal} if the interval constraints
it infers are as tight as possible, that is, if both bounds
of the inferred intervals are attainable (and thus cannot be pruned).

Figure~\ref{fig:add-sub-projections} gives non-optimal projections
for addition and subtraction.
For finite $x, y \in \Fset_{p,\emax}$, $\midf(x, y)$ denotes the
number that is exactly halfway between $x$ and $y$;
note that either $\midf(x, y) \in \Fset_{p,\emax}$
or $\midf(x, y) \in \Fset_{p+1,\emax}$.
Non-optimal projections for multiplication
and division can be found in \citep{Michel02,BotellaGM06}.
Optimal projections are known for monotonic functions over one argument
\citep{Michel02}, but they are generally not available for other functions.
Note, however, that optimality is not required in an interval-based
consistency approach to constraint solving, as filtering is just used
to remove some, not necessarily all, inconsistent values.

\subsection{The Marre-Michel Property}
\label{sec:marre-michel-property}

\citet{MarreM10} published an idea to improve the
filtering of the indirect projections for addition and subtraction.
This is based on a property of the distribution of floating-point
numbers among the reals: the greater a float, the greater the
distance between it and its immediate successor.
More precisely, for a given float $x$ with exponent $e_x$,
if $x^+ - x = \Delta$, then for $y$ of exponent $e_x + 1$ we have
$y^+ - y = 2\Delta$.

\begin{proposition} \textup{\cite[Proposition~1]{MarreM10}}
\label{prop:marre-michel-prop-1}
Let $z \in \Fset_{p,\infty}$ be such that $0 < z < \fpinf$; let also
\begin{align*}
  z
    &=
      \float{1. b_2 \cdots b_i \overbrace{0 \cdots 0}^k}{\expf{z}},
        &&\text{with $b_i = 1$;} \\
  \alpha
    &=
      \float{\overbrace{1.1 \cdots 1}^p}{\expf{z} + k},
        &&\text{with $k = p-i$;} \\
  \beta
    &=
      \alpha \fadd z.
\end{align*}
Then, for each $x, y \in \Fset_{p,\infty}$, $z = x \fsub y$
implies that $x \leq \beta$ and $y \leq \alpha$.
Moreover, $\beta \fsub \alpha = \beta - \alpha = z$.
\end{proposition}

\begin{figure}
\begin{center}
\begin{tikzpicture}[scale=1.0]

\draw (0, 0) -- (0.5, 0);
\draw[dashed] (0.5, 0) -- (2.0, 0);
\draw (2.0, 0) -- (3.0, 0);
\draw[dashed] (3.0, 0) -- (4.0, 0);
\draw (4.0, 0) -- (7.5, 0);
\draw[dashed] (7.5, 0) -- (8.5, 0);
\draw (8.5, 0) -- (11, 0);

\draw[dashed] (11, 0) -- (11.5, 0);
\draw[thick]  (0  , 0.125) -- (0  , -0.125);
\draw (0.125  , 0.0625) -- (0.125  , -0.0625);
\draw (0.250  , 0.0625) -- (0.250  , -0.0625);
\draw (0.375  , 0.0625) -- (0.375  , -0.0625);
\draw (0.500  , 0.0625) -- (0.500  , -0.0625);
% ...
\draw[thick]  (2.0 , 0.125 ) -- (2.0 , -0.125);
\draw (2.25, 0.0625) -- (2.25, -0.0625);
\draw (2.50, 0.0625) -- (2.50, -0.0625);
\draw (2.75, 0.0625) -- (2.75, -0.0625);
\draw (3.00, 0.0625) -- (3.00, -0.0625);
% ...
\draw (4.0 , 0.0625) -- (4.0 , -0.0625);
\draw (4.5 , 0.0625) -- (4.5 , -0.0625);
\draw (5.0 , 0.0625) -- (5.0 , -0.0625);
\draw (5.5 , 0.0625) -- (5.5 , -0.0625);
\draw[thick] (6.0 , 0.125 ) -- (6.0 , -0.125);
\draw (7.0 , 0.0625) -- (7.0 , -0.0625);
% ...
\draw ( 9.0, 0.0625) -- ( 9.0, -0.0625);
\draw (10.0, 0.0625) -- (10.0, -0.0625);
\draw (11.0, 0.0625) -- (11.0, -0.0625);


\draw (1.75, 0.5) node {$z$};
\draw (0  , 0.3 ) -- (3.5, 0.3 );
\draw (0  , 0.35) -- (0  , 0.25);
\draw (3.5, 0.35) -- (3.5, 0.25);

\draw[color=red] (5.5, 0.85) node {$\alpha$};
\draw (6.0, 0.35) node {$\alpha^+$};
\draw (7.0, 0.35) node {$\alpha^{++}$};
\draw[color=red] (9.0, 0.85) node {$\beta$};
\draw (10.0, 0.35) node {$\beta^+$};

\draw         (7.25, 0.8) node {$z$};
\draw[dotted] (5.5 , 0.6) --   (5.5, 0.0625);
\draw[->]     (7.25, 0.6) ->   (5.5, 0.6);
\draw[->]     (7.25, 0.6) ->   (9.0, 0.6);
\draw[dotted] (9.0 , 0.6) --   (9,0, 0.0625);
\draw[dotted] (9.0 , 0.6) --   (9.0, 0.0625);

\draw         (5.75, -0.50) node {$\Delta$};
\draw[<->]    (5.50, -0.25) -> (6.00, -0.25);

\draw       (6.50, -0.50) node {$2 \Delta$};
\draw[<->]  (6.00, -0.25) -> (7.00, -0.25);

\draw       (9.50, -0.50) node {$2 \Delta$};
\draw[<->]  (9.00, -0.25) -> (10.00, -0.25);

\draw (0, -0.5 ) node {$0$};

\end{tikzpicture}
\end{center}
\caption{An illustration of the Marre-Michel property: the segment $z$,
        if it has to represent the difference between two floats,
        cannot be moved past $\alpha$}
\label{fig:ULP-max}
\end{figure}
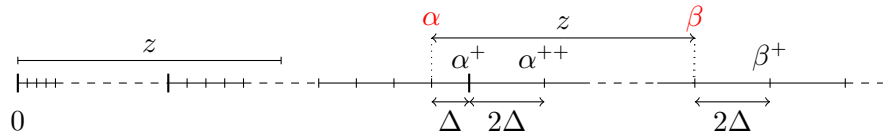

This property, which can be generalized to subnormals, can intuitively
be explained on Figure~\ref{fig:ULP-max} as follows.
Let $z \in \Fset_{p,\infty}$ be a strictly positive constant such that
$z = x \fsub y$, where $x, y \in \Fset_{p,\infty}$ are unknown.
The Marre-Michel property says that $y$ cannot be greater than
$\alpha$.  In fact, $\alpha$ is carefully positioned so that
$\alpha^{++} - \alpha^+ = 2(\alpha^+ - \alpha)$,
$\expf{\alpha} + 1 = \expf{\beta}$ and $z = \beta - \alpha$;
if we take $y = \alpha^+$ we need $x > \beta$ if we want $z = x - y$;
however, the smallest element of $\Fset_{p,\infty}$ that is greater than $\beta$,
$\beta^+$, is $2\Delta$ away from $\beta$, i.e., too much.
Going further with $y$ does not help: if we take $y \geq \alpha^+$,
then $y - \alpha$ is an odd multiple of $\Delta$ (one $\Delta$ step
from $\alpha$ to $\alpha^+$, all the subsequent steps being even multiples
of $\Delta$), whereas for each $x \geq \beta$, $x - \beta$ is an even multiple
of $\Delta$.
Hence, if $y > \alpha$,
$\bigl|z - (x - y)\bigr| \geq \Delta = 2^{\expf{z}+1-i}$.
However, since $k \neq p-1$, $z^+ - z = z - z^- = 2^{\expf{z}+1-p} \leq \Delta$.
The last inequality, which holds because $p \geq i$, implies
$z \neq x \fsub y$.
A similar reasoning allows one to see that $x$ cannot be greater
than $\beta$ independently from the value of $y$.
In order to improve the filtering of the addition/subtraction
projectors, \citet{MarreM10} presented an
algorithm to maximize the values of $\alpha$ and $\beta$ over
an interval.  As we will see, that algorithm is not correct for some inputs.
In Section~\ref{sec:ULP-add-sub},
the main ideas behind the work presented
in~\citep{MarreM10} will be revisited, corrected and discussed.

\section{Filtering by Maximum ULP}
\label{sec:max-ulp}

In this section we first informally present, by means of worked
numerical examples, the techniques that are precisely defined later.
We then reformulate the Marre-Michel property so as to generalize
it to subnormals and to multiplication and division operators.
The filtering algorithms that result from this generalization are
collectively called \emph{filtering by maximum ULP}.

\subsection{Motivating Example}

Consider the IEEE~754 single-precision constraint
$\var{z} = \var{x} \fadd \var{y}$ with initial intervals
$\var{z} \in [\fminf, \fpinf]$,
$\var{x} \in [-\float{1.0}{50}, \float{1.0}{50}]$ and
$\var{y} \in [-\float{1.0}{30}, \float{1.0}{30}]$.
Forward projection gives
\[
  \var{z} \in \bigl[-\float{1.\overbrace{0 \cdots 0}^{19}1}{50},
                     \float{1.\overbrace{0 \cdots 0}^{19}1}{50}\bigr],
\]
which is optimal, as both bounds are attainable.
Suppose now the interval for $\var{z}$ is further restricted to
$\var{z} \in [1.0, 2.0]$ due to, say, a constraint from an if-then-else
in the program or another indirect projection.

With the classical indirect projection we obtain
$\var{x}, \var{y} \in [-\float{1.0}{30}, \float{1.0}{30}]$, which,
however, is not optimal.
For example, pick $x = \float{1.0}{30}$:
for $y = -\float{1.0}{30}$ we have $x \fadd y = 0$
and $x \fadd y^+ = 64$.
By monotonicity of $\fadd$, for no $y \in [-\float{1.0}{30}, \float{1.0}{30}]$
we can have $x \fadd y \in [1.0, 2.0]$.

With our indirect projection, fully explained later,
we obtain, from $\var{z} \in [1.0, 2.0]$, the much tighter intervals
$\var{x}, \var{y} \in [-\float{1.1\cdots1}{24}, \float{1.0}{25}]$.
These are actually optimal as
\(
  -\float{1.1\cdots1}{24} \fadd \float{1.0}{25}
  =
  \float{1.0}{25} \fadd -\float{1.1\cdots1}{24}
  = 2.0
\).
This example shows that filtering by maximum ULP can be stronger
than classical interval-consistency based filtering.
However, the opposite phenomenon is also possible.  Consider again
$\var{z} = \var{x} \fadd \var{y}$ with $\var{z} \in [1.0, 2.0]$.
Suppose now the constraints for $\var{x}$ and $\var{y}$ are
$\var{x} \in [1.0, 5.0]$ and $\var{y} \in [-\fmax, \fmax]$.
As we have seen, our indirect projection gives
$\var{y} \in [-\float{1.1\cdots1}{24}, \float{1.0}{25}]$;
in contrast, the classical indirect projection exploits the available
information on $\var{x}$ to obtain $\var{y} \in [-4, 1]$.
Indeed, classical and maximum ULP filtering for addition
and subtraction are orthogonal: both should be applied in order to
obtain precise results.

For an example on multiplication,
consider the IEEE~754 single-precision constraint
$\var{z} = \var{x} \fmul \var{y}$ with initial intervals
$\var{z} \in [\float{1.0}{-50}, \float{1.0}{-30}]$ and
$\var{x}, \var{y} \in [\fminf, \fpinf]$.
In this case, classical projections do not allow pruning the intervals.
However, take $x = \float{1.1}{119}$:
for $y = 0$ we have $x \fmul y = 0$
and $x \fmul y^+ = \float{1.1}{-30}$.
By monotonicity of $\fmul$, for no $y \in [\fminf, \fpinf]$
we can have $x \fmul y \in [\float{1.0}{-50}, \float{1.0}{-30}]$.

On the same example,
$\var{x}, \var{y} \in [-\float{1.0\cdots0}{119}, \float{1.0\cdots0}{119}]$
are the constraints inferred by our indirect projection.
These are optimal because
\(
  \float{1.0}{-30}
    = -\float{1.0\cdots0}{119} \fmul -\float{1.0\cdots0}{149}
    =  \float{1.0\cdots0}{119} \fmul  \float{1.0\cdots0}{149}
\).
As is the case for addition, classical indirect projection can be
more precise.
Consider again $\var{z} = \var{x} \fmul \var{y}$ with
$\var{z} \in [\float{1.0}{-50}, \float{1.0}{-30}]$,
$\var{x} \in [2.0, 4.0]$ and
$\var{y} \in [-\fmax, \fmax]$.
Classical indirect projection infers
$\var{y} \in [\float{1.0}{-52}, \float{1.0}{-31}]$
by exploiting the information on $\var{x}$.

\subsection{Round-To-Nearest Tail-To-Even}

We now formally define the roundTiesToEven rounding mode.
To do that, we first introduce two functions:
$\ferrdown{z}$ and $\ferrup{z}$ give
the distance between $z^+$ and $z$ and
the distance between $z$ and $z^-$.

\begin{definition}
\label{def:rounding-error-functions}
The partial functions
$\pard{\ferrup{}}{\Fsetsub_{p,\emax}}{\Rset}$ and
$\pard{\ferrdown{}}{\Fsetsub_{p,\emax}}{\Rset}$
are defined as follows,
for each finite $z \in \Fsetsub_{p,\emax}$:
\begin{align*}
  \ferrdown{z}
    &=
      \begin{cases}
        2^{1-p+\emax},
          &\text{if $z = \fmax$;} \\
        \fmin,
          &\text{if $z = +0$ or $z = -0$;} \\
        z^+ - z,
          &\text{otherwise;}
      \end{cases} \\
  \ferrup{z}
    &=
      \begin{cases}
        2^{1-p+\emax}
          &\text{if $z = -\fmax$;} \\
        \fmin,
          &\text{if $z = +0$ or $z = -0$;} \\
        z - z^-,
          &\text{otherwise.}
      \end{cases}
\end{align*}
\end{definition}
Note the special cases when $z = \pm 0$:
since both $+0$ and $-0$ represent the real number $0$,
the distance between $z^+ = \fmin$ and $z = \pm 0$ is $\fmin$.
We can now define the function $\roundnear{\cdot}$
that captures roundTiesToEven.

\begin{definition}
\label{def:round-to-nearest-tail-to-even}
For $x \in \Rset$, $\roundnear{x}$ is defined
as follows:
\[
  \roundnear{x}
    =
      \begin{cases}
        +0,   &\text{if $0 \leq x \leq  \ferrdown{0} /2$;} \\
        -0,   &\text{if $ -\ferrup{0}/2 \leq x <  0$;} \\
         z,
            &\text{if $z\in \Fsetsub_{p,\emax} \setminus \{ -\infty, +\infty \}$ and either $\feven(z)$ and} \\
            &\text{$z - \ferrup{z}/2 \leq x \leq z+\ferrdown{z}/2$,
                   or $\fodd(z)$ and} \\
            &\text{$z - \ferrup{z}/2 < x < z + \ferrdown{z}/2; $} \\
        +\infty,
            &\text{if $x \geq \fmax +\ferrdown{\fmax}/2;$ } \\
        -\infty,
            &\text{if $x \leq -\fmax -\ferrup{-\fmax}/2.$} \\
      \end{cases}
\]
\end{definition}

\begin{figure}
\begin{center}
\begin{tikzpicture}[scale=1.0]

\draw[dashed] (0, 0) -- (1, 0);
\draw (1, 0) -- (11,0);
\draw[dashed] (11, 0) -- (12, 0);

\draw[dashed] (0, 2) -- (1, 2);
\draw (1, 2) -- (11, 2);
\draw[dashed] (11, 2) -- (12, 2);

\draw  (1.8 , 0.04 ) -- (1.8 , -0.04);
\draw  (1.9 , 0.04 ) -- (1.9 , -0.04);
\draw[thick]  (2.0 , 0.125 ) -- (2.0 , -0.125);
\draw  (2.1 , 0.04 ) -- (2.1 , -0.04);
\draw  (2.2 , 0.04 ) -- (2.2 , -0.04);
\draw (3.25, 0.0625) -- (3.25, -0.0625);
\draw  (4.3 , 0.04 ) -- (4.3 , -0.04);
\draw  (4.4 , 0.04 ) -- (4.4 , -0.04);
\draw[thick]  (4.5, 0.125 ) -- (4.5, -0.0625);
\draw  (4.6 , 0.04 ) -- (4.6 , -0.04);
\draw  (4.7 , 0.04 ) -- (4.7 , -0.04);

\draw (5.75, 0.0625) -- (5.75, -0.0625);
\draw  (6.8 , 0.04 ) -- (6.8 , -0.04);
\draw  (6.9, 0.04 ) -- (6.9 , -0.04);
\draw[thick]  (7, 0.125 ) -- (7, -0.0625);
\draw  (7.1 , 0.04 ) -- (7.1 , -0.04);
\draw  (7.2, 0.04 ) -- (7.2 , -0.04);
\draw (8.25, 0.0625) -- (8.25, -0.0625);
\draw  (9.3 , 0.04 ) -- (9.3 , -0.04);
\draw  (9.4, 0.04 ) -- (9.4 , -0.04);
\draw[thick]  (9.5, 0.125 ) -- (9.5, -0.125);
\draw  (9.6 , 0.04 ) -- (9.6 , -0.04);
\draw  (9.7, 0.04 ) -- (9.7 , -0.04);

\draw[thick] (3.25, 2.0625) -- (3.25, 1.9375);
\draw[thick](5.75, 2.0625) -- (5.75, 1.9375);
\draw[thick] (8.25, 2.0625) -- (8.25, 1.9375);

\draw[(-)]  (2.0, 0.5 ) -- (4.5, 0.5 );
\draw[<->]  (4.5, 1.1) -- (7, 1.1);
\draw[(-)]  (7.0, 0.5 ) -- (9.5, 0.5 );

\draw[dotted] (2.1, 0.6 ) --  (3.25, 1.9);
\draw[dotted] (4.4, 0.6 ) --  (3.25, 1.9);

\draw[dotted] (4.6, 1.2 ) --  (5.75, 1.9);
\draw[dotted] (6.9, 1.2 ) --  (5.75, 1.9);

\draw[dotted] (7.1, 0.6 ) --  (8.25, 1.9);
\draw[dotted] (9.4, 0.6 ) --  (8.25, 1.9);

\draw[color=red] (3.25, 0.3) node {$z^-$} ;
\draw[color=red] (5.75, 0.3) node {$z$} ;
\draw[color=red] (8.25, 0.3) node {$z^+$} ;

\draw[color=red] (3.25, 1.7) node {$z^-$} ;
\draw[color=red] (5.75, 1.7) node {$z$} ;
\draw[color=red] (8.25, 1.7) node {$z^+$} ;

\draw (2.0 , -0.8)  node {$z^- - \ferrup{z^-}/2$};
\draw (4.5 , -0.8)  node {$z^- + \ferrdown{z^-}/2$};
\draw (4.5 , -0.3)  node {$z - \ferrup{z}/2$};
\draw (7.0 , -0.3)  node {$z + \ferrdown{z}/2$};
\draw (7.0 , -0.8)  node {$z^+ - \ferrup{z^+}/2$};
\draw (9.5 , -0.8)  node {$z^+ + \ferrdown{z^+}/2$};

\draw (0.5, -0.5 ) node {$\Rset$};
\draw (0.5, 1.5 ) node {$\Fset$};
\end{tikzpicture}
\end{center}
\caption{Rounding of real numbers in the neighborhood of an even
         floating-point number $z$ under roundTiesToEven}
\label{fig:round-to-nearest-tail-to-even}
\end{figure}
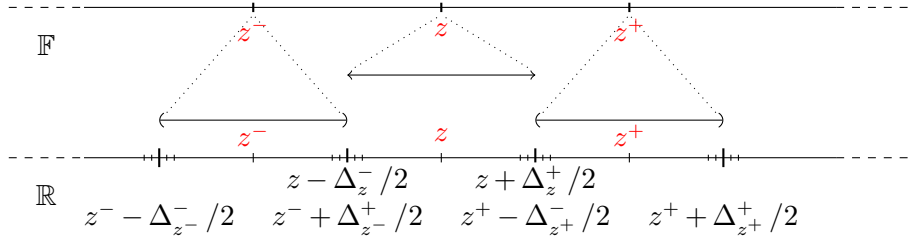
Figure~\ref{fig:round-to-nearest-tail-to-even} illustrates the
roundTiesToEven rounding mode;
if $z$ is even, each real number between
$z - \ferrup{z}/2$ and $z + \ferrdown{z}/2$, \emph{including} extremes,
is rounded to the same floating-point number $z$.
As $z$ is even, $z^-$ is odd,
and each real number between
$z^- - \ferrup{z^-}/2$ and $z^- + \ferrdown{z^-}/2$,
\emph{excluding} extremes, is rounded to $z^-$.
Similarly for $z^+$.
Note that point
$z - \ferrup{z}/2$ coincides with $z^- + \ferrdown{z^-}/2$ and
$z + \ferrdown{z}/2$ coincides with $z^+ - \ferrup{z^+}/2$.

All rounding modes are monotonic; in particular,
for each $x, y \in \Rset$,
$x \leq y$ implies $\roundnear{x} \leq \roundnear{y}$.
Moreover,
the \emph{chop} and \emph{near} rounding modes are \emph{symmetric},
i.e., the value after rounding does not depend on the sign:
for each $x \in \Rset$, $\roundnear{x} = -\roundnear{-x}$.

\subsection{Upper Bound}

It is worth pointing out that, while arithmetic operations on reals
are strictly monotone, that is if $x + y = z$ then $x_1 + y > z$ for
any $x_1 > x$, in floating-point arithmetic, operations are just
monotone. If $x + y = z$ then we may still have $x_1 + y = z$ for some
(or many) $x_1 > x$ since addition over the floats is
absorbing. Therefore, for determining the greatest (or the smallest)
$x_1$ satisfying $x_1 + y = z$ and correctly filter intervals over the
floats, we need to introduce an appropriate, duly justified,
framework.

For each IEEE~754 floating-point operation
$\mathord{\fop} \in \{ \fadd, \fsub, \fmul, \fdiv \}$,
in later sections we will define the sets
$\Fset_{\mathord{\fop}} \subseteq \Fsetsub_{p,\emax}$ and
$\bar{\Fset}_{\mathord{\fop}} \subseteq \Fset_{p,\infty}$.
Then we will define functions
$\fund{\ulpmmax{\fop}}{\Fset_{\mathord{\fop}}}{\bar{\Fset}_{\mathord{\fop}}}$
(see Definition~\ref{def:ulpmmax_fadd}
in Section~\ref{sec:ULP-add-sub} for $\fadd$ and, consequently, $\fsub$,
Definition~\ref{def:ulpmmax_fmul} in Section~\ref{sec:ULP-mul} for $\fmul$,
and Definition~\ref{def:mu-fdiv} in Section~\ref{sec:ULP-div} for $\fdiv$)
that satisfy the following property,
for each $z \in \Fset_{\fop} \setminus \{ -0, +0, -\infty \}$:
\begin{align}
\label{eq:ulpmmax1new}
  \ulpmmax{\fop}(z)
    =
      \max
        \{\,
          v \in \bar{\Fset}_{\mathord{\fop}}
        \mid
          \exists y \in \bar{\Fset}_{\mathord{\fop}} \st v \fop y &= z
        \,\}.
\end{align}
In words, $\ulpmmax{\fop}(z)$ is the greatest float in
$\bar{\Fset}_{\mathord{\fop}}$ that can be the left operand
of $\mathord{\fop}$ to obtain $z$.

Verifying that a function $\ulpmmax{\fop}$ satisfies~\eqref{eq:ulpmmax1new}
is equivalent to proving that it satisfies the following properties,
for each $z \in \Fset_{\fop} \setminus \{ -0, +0, -\infty \}$:
\begin{align}
\label{eq:ulpmmax1}
  \ulpmmax{\fop}(z)
    &\in
      \{\,
        v \in \bar{\Fset}_{\mathord{\fop}}
      \mid
        \exists y \in \bar{\Fset}_{\mathord{\fop}} \st v \fop y = z
      \,\};   \\
\label{eq:ulpmmax2}
  \forall z' \in\bar{\Fset}_{\fop}
    \itc
      z' > \ulpmmax{\fop}(z)
        \implies z'
          &\notin
            \{\,
              v \in \bar{\Fset}_{\mathord{\fop}}
            \mid
              \exists y \in \bar{\Fset}_{\mathord{\fop}} \st v \fop y = z
            \,\} .
\end{align}
Property~\eqref{eq:ulpmmax2} means $\ulpmmax{\fop}(z)$
is a \emph{correct} upper bound for the possible values of $x$,
whereas~\eqref{eq:ulpmmax1} implies that $\ulpmmax{\fop}(z)$
is the \emph{most precise} upper bound we could choose.

Note that we may have $\bar{\Fset}_{\fop} \nsubseteq \Fset_{p,\emax}$:
property~\eqref{eq:ulpmmax1new}
refers to an idealized set of floating-point numbers with unbounded exponents.

Since we are interested in finding the upper bound of
$\ulpmmax{\fop}(z)$ for $z \in [\lb{z}, \ub{z}]$, we need the
following

\begin{proposition}
\label{prop:ULP-max-upper}
Let $w, v_1, \dots, v_n \in \Fset_{\fop} \setminus \{ -0, +0, -\infty \}$
be such that, for each $i = 1$, \dots,~$n$,
$\ulpmmax{\fop}(w) \geq \ulpmmax{\fop}(v_i)$.
Then,
for each $w' \in \bar{\Fset}_{\mathord{\fop}}$ with $w' > \ulpmmax{\fop}(w)$
and each $z \in \Fset_{\fop} \setminus \{ -0, +0, -\infty \}$,
we have that
\(
  w'
    \notin
      \{\,
        v \in \bar{\Fset}_{\mathord{\fop}}
      \mid
        \exists y \in \bar{\Fset}_{\mathord{\fop}} \st v \fop y = z
      \,\}
\).
\end{proposition}
\proof{Proof.}
Follows directly from~\eqref{eq:ulpmmax1new}.
\endproof

Let $\var{z} = \var{x} \fop \var{y}$ be a floating-point constraint
where $-0, +0, -\infty \notin [\lb{z}, \ub{z}]$
and let $w \in [\lb{z}, \ub{z}]$ be such that
$\ulpmmax{\fop}(w) \geq \ulpmmax{\fop}(v)$
for each $v \in [\lb{z}, \ub{z}]$:
then no element of $\var{x}$ that is greater than $\ulpmmax{\fop}(w)$
can participate to a solution of the constraint.

Dually, in order to refine the upper bound of $\var{y}$
subject to $\var{z} = \var{x} \fop \var{y}$,
it is possible to define a function $\ulpmmaxp{\fop}$
satisfying the following property,
for each $z \in \Fset_{\fop} \setminus \{ -0, +0, -\infty \}$:
\begin{align}
\label{eq:ulpmmaxp1new}
  \ulpmmaxp{\fop}(z)
    =
      \max
        \{\,
          v \in \bar{\Fset}_{\mathord{\fop}}
        \mid
          \exists x \in \bar{\Fset}_{\mathord{\fop}} \st x \fop v &= z
        \,\}.
\end{align}
Due to~\eqref{eq:ulpmmaxp1new}, a result analogous to the one of
Proposition~\ref{prop:ULP-max-upper} holds for $\ulpmmaxp{\fop}$,
which allows refining the interval for $\var{y}$.
Note, though, that when $\fop$ is commutative
(i.e., it is $\fadd$ or $\fmul$),
$\ulpmmax{\fop} = \ulpmmaxp{\fop}$.

\subsection{Lower bound}

For computing the lower bound, we will introduce functions
$\fund{\ulpmmin{\fop}}{\Fset_{\mathord{\fop}}}{\bar{\Fset}_{\mathord{\fop}}}$
(defined in terms of the corresponding $\ulpmmax{\fop}$ functions
in Section~\ref{sec:ULP-add-sub} for $\fadd$ and $\fsub$,
in Section~\ref{sec:ULP-mul} for $\fmul$, and
in Section~\ref{sec:ULP-div} for $\fdiv$)
satisfying the following property,
for each $z \in \Fset_{\fop} \setminus \{ -0, +0, +\infty \}$:
\begin{align}
\label{eq:ulpmmin1new}
  \ulpmmin{\fop}(z)
    =
      \min
        \{\,
          v \in \bar{\Fset}_{\mathord{\fop}}
        \mid
          \exists y \in \bar{\Fset}_{\mathord{\fop}} \st v \fop y &= z
        \,\}.
\end{align}
This property entails a result similar to
Proposition~\ref{prop:ULP-max-upper}:
given constraint $\var{z} = \var{x} \fop \var{y}$
where $-0, +0, +\infty \notin [\lb{z}, \ub{z}]$
and $w \in [\lb{z}, \ub{z}]$ such that
$\ulpmmin{\fop}(w) \leq \ulpmmin{\fop}(v)$
for each $v \in [\lb{z}, \ub{z}]$,
the float $\ulpmmin{\fop}(w)$ is a possibly refined lower bound for~$\var{x}$.

In a dual way, in order to refine the lower bound of $\var{y}$
subject to $\var{z} = \var{x} \fop \var{y}$,
we will define functions $\ulpmminp{\fop}$ satisfying,
for each $z \in \Fset_{\fop} \setminus \{ -0, +0, +\infty \}$:
\begin{align}
\label{eq:ulpmminp1new}
  \ulpmminp{\fop}(z)
    =
      \min
        \{\,
          v \in \bar{\Fset}_{\mathord{\fop}}
        \mid
          \exists x \in \bar{\Fset}_{\mathord{\fop}} \st x \fop v &= z
        \,\}.
\end{align}
Property~\eqref{eq:ulpmminp1new} ensures that,
under $\var{z} = \var{x} \fop \var{y}$
where $-0, +0, +\infty \notin [\lb{z}, \ub{z}]$,
if $w \in [\lb{z}, \ub{z}]$ is such that
$\ulpmminp{\fop}(w) \leq \ulpmminp{\fop}(v)$
for each $v \in [\lb{z}, \ub{z}]$,
then the float $\ulpmminp{\fop}(w)$
is a possibly refined lower bound for~$\var{y}$.

Again, when $\fop$ is commutative $\ulpmmin{\fop} = \ulpmminp{\fop}$.

\subsection{Filtering by Maximum ULP on Addition/Subtraction}
\label{sec:ULP-add-sub}

In this section we introduce the functions
$\ulpmmax{\fadd}$, $\ulpmmin{\fadd}$,
$\ulpmmax{\fsub}$, $\ulpmmaxp{\fsub}$,
$\ulpmmin{\fsub}$ and $\ulpmminp{\fsub}$.
Note that, since $\fadd$ is commutative, we have
$\ulpmmaxp{\fadd} = \ulpmmax{\fadd}$ and
$\ulpmminp{\fadd} = \ulpmmin{\fadd}$.
Moreover, the function
$\fund{\ulpmmin{\fadd}}{\Fset_{\mathord{\fadd}}}{\bar{\Fset}_{\mathord{\fadd}}}$
can be defined in terms of the function $\ulpmmax{\fadd}$ as follows:
for each $z \in \Fset_{\fadd} \setminus \{ -0, +0, +\infty \}$,
$\ulpmmin{\fadd}(z) = -\ulpmmax{\fadd}(-z)$.
We see that, if $\ulpmmax{\fadd}$ satisfies Property~\eqref{eq:ulpmmax1new},
then $\ulpmmin{\fadd}$ satisfies Property~\eqref{eq:ulpmmin1new}.
Again, since $\fadd$ is commutative, $\ulpmminp{\fadd} = \ulpmmin{\fadd}$.

The first step for defining $\ulpmmax{\fadd}$ consists in extending
Proposition~\ref{prop:marre-michel-prop-1} in order to explicitly
handle subnormal numbers.
Such extension was already sketched by \citet{MarreM10}: here
we fully describe it and prove its correctness.
Subnormals, which in $\Fsetsub_{p,\emax}$ are represented
by numbers having the hidden bit $b_1 = 0$ and exponent ${\emin}$,
can be represented in $\Fset_{p,\infty}$ by numbers with
$b_1 = 1$ and exponent strictly smaller than ${\emin}$.
Namely, the element of $\Fsetsub_{p,\emax}$
\begin{align*}
  \mathalpha{0.0\cdots01b_{j+1}\cdots b_p}
    &\times 2^{\emin} \\
\intertext{%
can be represented in $\Fset_{p,\infty}$ by the (normal) float
}
  \mathalpha{1.b_{j+1}\cdots b_p \overbrace{0 \cdots 0}^{j-1}}
    &\times 2^{\emin-(j-1)}.
\end{align*}
Based on this observation we can state the following

\begin{proposition}
\label{prop:marre-michel-prop-ext}
Let $z \in \Fsetsub_{p,\emin}$ be such that $0 < z < \fnormin$;
define also
\begin{align*}
  z
    &=
      \float{0.0\cdots 0 1 b_{j+1} \cdots b_i \overbrace{0 \cdots 0}^k}{\emin},
        &&\text{with $b_i = 1$;} \\
  \alpha
    &=
      \float{\overbrace{1.1 \cdots 1}^p}{\emin + k},
        &&\text{with $k = p-i$;} \\
  \beta
    &=
      \alpha \fadd z.
\end{align*}
Then, for each $x, y \in \Fsetsub_{p,\emax}$, $z = x \fsub y$
implies that $x \leq \beta$ and $y \leq \alpha$.
Moreover, $\beta \fsub \alpha = \beta - \alpha = z$.
\end{proposition}
\proof{Proof.}
The subnormal $z$ is represented in $\Fset_{p,\infty}$
by the normal float
\[
  \hat{z}
    = \float{1.b_{j+1} \cdots b_i \overbrace{0 \cdots 0}^k
                                 \overbrace{0 \cdots 0}^{j-1}}{\emin-(j-1)}
    = \float{1.b_{j+1}\cdots b_i \overbrace{0 \cdots 0}^{k+j-1}}{\emin-(j-1)}.
\]
We can apply Proposition~\ref{prop:marre-michel-prop-1}
to $\hat{z}$ and obtain
\(
  \alpha
     = \float{1.1 \cdots 1}{\emin -(j-1)+ k+j-1}
     = \float{1.1 \cdots 1}{\emin + k}
\).
Moreover, Proposition~\ref{prop:marre-michel-prop-1} assures that
\[
  \beta
    =
      \alpha \fadd \float{1.b_{j+1}\cdots b_i \overbrace{0 \cdots 0}^{k+j-1}}
                         {\emin-(j-1)}
\]
is such that, for each $x, y \in \Fset_{p,\infty}$, $z = x \fsub y$
implies $x \leq \beta$ and $y \leq \alpha$ and
$\beta \fsub \alpha = \beta - \alpha = z$.
Since each number in $\Fsetsub_{p,\emax}$ has an equivalent representation
in $\Fset_{p,\infty}$, we only need to prove that
$\beta = \alpha \fadd z$, which holds, since
\begin{align*}
\notag
  \beta
    &=
      \alpha \fadd \float{1.b_{j+1}\cdots b_i \overbrace{0 \cdots 0}^{k+j-1}}
                         {\emin-(j-1)} \\
    &=
      \alpha \fadd \float{0.0\cdots01b_{j+1}\cdots b_i \smash{\underbrace{0 \cdots 0}_k}}
                         {\emin} \\
\notag
    &=
      \alpha \fadd z.
\end{align*}
\Halmos
\endproof

Using Propositions~\ref{prop:marre-michel-prop-1}
and~\ref{prop:marre-michel-prop-ext},
we formally define the function $\ulpmmax{\fadd}$ as follows.

\begin{definition}
\label{def:ulpmmax_fadd}
Let $\Fset_{\fadd} = \Fsetsub_{p,\emax}$,
$\bar{\Fset}_{\fadd} = \Fset^+_{p,\infty}$,
and $z \in \Fset_{\fadd}$ be such that
$|z| = \float{b_1. b_2 \cdots b_i 0 \cdots 0}{\expf{z}}$,
with $b_i = 1$.
Similarly to Propositions~\ref{prop:marre-michel-prop-1}
and~\ref{prop:marre-michel-prop-ext}, let
$k = p-i$,
$\alpha = \float{1.1 \cdots 1}{\expf{z} + k}$
and
$\beta = \alpha \fadd |z|$.
Then
$\fund{\ulpmmax{\fadd}}{\Fset_{\mathord{\fadd}}}{\bar{\Fset}_{\mathord{\fadd}}}$
is defined, for each $z \in \Fset_{\fadd}$, by
\[
  \ulpmmax{\fadd}(z)
    =
      \begin{cases}
        +\infty,
          &\text{if $z = -\infty$ or $z = +\infty$;} \\
        \alpha,
          &\text{if $-\infty < z < 0$;} \\
        +0,
          &\text{if $z = -0$ or $z = +0$;} \\
        \beta,
          &\text{if $0 < z < +\infty$.}
      \end{cases}
\]
\end{definition}

\begin{theorem}
\label{thm:mu-fadd-satisfies-properties}
Function $\ulpmmax{\fadd}$ is well-defined and satisfies~\eqref{eq:ulpmmax1}
and~\eqref{eq:ulpmmax2}.
\end{theorem}
\proof{Proof.}
We first show that $\ulpmmax{\fadd}(z)$ is well-defined,
i.e., that it is a total function from $\Fsetsub_{p,\emax}$ to $\Fset^+_{p,\infty}$.
To this aim note that $\alpha$ and $\beta$ are always non-negative
normal floating-point numbers belonging to $\Fset_{p,\infty}$,
and that $\ulpmmax{\fadd}(z)$ is defined for each $z \in \Fsetsub_{p,\emax}$.
Secondly, let us consider the following cases:

$z = +\infty$:
for each $y \neq -\infty$ we have $+\infty \fadd y = +\infty$;
thus, as $\ulpmmax{\fadd}(z) = +\infty$, \eqref{eq:ulpmmax1} holds
and~\eqref{eq:ulpmmax2} vacuously holds.

$\fnormin \leq z < +\infty$:
we can apply Proposition~\ref{prop:marre-michel-prop-1}
to obtain $z = \beta \fsub \alpha$.
Then note that
\(
 \beta \fsub \alpha
     = \roundnear{\beta - \alpha}
     = \roundnear{\beta + -\alpha}
     = \beta \fadd -\alpha
\).
Hence, $\beta \fadd -\alpha = z$.
Thus, $\ulpmmax{\fadd}(z) \fadd -\alpha = \beta \fadd -\alpha = z$
and~\eqref{eq:ulpmmax1} is satisfied with $y = -\alpha$.
For proving~\eqref{eq:ulpmmax2}, first
note that $\beta > -\alpha$ since $\beta > 0$ and $\alpha > 0$.
Moreover, by Proposition~\ref{prop:marre-michel-prop-1},
we know that there does not exist an $x \in \Fset_{p,\infty}$ with $x > \beta$
such that there exists $y \in \Fset_{p,\infty}$ that satisfies
$x \fsub y = z$.
Since $x \fsub y = x \fadd -y$ we can conclude that,
for each $z' > \beta = \ulpmmax{\fadd}(z)$,
it does not exist $y' \in \Fset_{p,\infty}$ such that
$z' \fadd y' = z$.
Hence also~\eqref{eq:ulpmmax2} holds.

$0< z < \fnormin$: by applying Proposition~\ref{prop:marre-michel-prop-ext}
instead of Proposition~\ref{prop:marre-michel-prop-1}
we can reason exactly as in the previous case.

$-\infty < z \leq -\fnormin $:
since $0 < -z < +\infty$ we can apply
Proposition~\ref{prop:marre-michel-prop-1} to $-z$ and obtain
$\beta \fsub \alpha = -z$ and thus $-(\beta \fsub \alpha) = z$.
As $\roundnear{\cdot}$ is a symmetric rounding mode, we have
\(
  -(\beta \fsub \alpha)
    = -\roundnear{\beta - \alpha}
    = \roundnear{\alpha - \beta}
    = \alpha \fadd -\beta
    = z
\).
Thus, $\ulpmmax{\fadd}(z) \fadd -\beta = \alpha \fadd -\beta = z$
and~\eqref{eq:ulpmmax1} is satisfied with $y = -\beta$.
For proving~\eqref{eq:ulpmmax2}, first
note that $\alpha > -\beta$ since $\alpha > 0$ and $\beta > 0$.
Moreover, by Proposition~\ref{prop:marre-michel-prop-1},
we know that there does not exist an $y \in \Fset_{p,\infty}$ with $y > \alpha$
such that there exists $x \in \Fset_{p,\infty}$ that satisfies
$x \fsub y = -z$.  Since $x \fsub y = -z$ is equivalent to $y \fadd -x = z$,
we can conclude that,
for each $z' > \alpha = \ulpmmax{\fadd}(z)$,
it does not exist $y' \in \Fset_{p,\infty}$ such that
$z' \fadd y' = z$.
Therefore, also in this case, \eqref{eq:ulpmmax2} holds.

$-\fnormin < z < 0$: by applying Proposition~\ref{prop:marre-michel-prop-ext}
instead of Proposition~\ref{prop:marre-michel-prop-1}
we can reason exactly as in the previous case.
\Halmos
\endproof

As we have already observed,
since $\fadd$ is commutative we have $\ulpmmaxp{\fadd} = \ulpmmax{\fadd}$,
that is, the same function $\ulpmmax{\fadd}$ is used to filter
both $x$ and $y$ with respect to $z = x \fadd y$.

We now need algorithms to maximize $\ulpmmax{\fadd}$ and minimize
$\ulpmmin{\fadd}$ over an interval of floating-point values.
Since the two problems are dual to each other, we will focus on
the maximization of $\ulpmmax{\fadd}$.
As $\ulpmmax{\fadd}$ is not monotonic, a nontrivial analysis of its range
over an interval is required.
When the interval contains only finite, nonzero and positive (resp., negative)
values, the range of $\ulpmmax{\fadd}$ has a simple shape.
We are thus brought to consider an interval $[\lb{z}, \ub{z}]$
such that $\lb{z} \notin \{ -\infty, -0, +0 \}$,
$\ub{z} \notin \{ -0, +0, +\infty \}$,
and $\lb{z}$ and $\ub{z}$ have the same sign.
We will now revisit, correct and extend to subnormal floating-point
numbers the algorithm originally proposed by \citet{MarreM10}
to maximize $\ulpmmax{\fadd}$ over $[\lb{z}, \ub{z}]$.

The idea presented in \citep{MarreM10} is the following.
When dealing with an interval $[\lb{z},\ub{z}]$ with $\lb{z} > 0$,
$\alpha$ (and thus $\beta$ and, therefore, our $\ulpmmax{\fadd}$)
grows (i) with the exponent and (ii) with the number of successive $0$ bits
to the right of the significand,
i.e., $k$ in Propositions~\ref{prop:marre-michel-prop-1}
and~\ref{prop:marre-michel-prop-ext}
and in Definition~\ref{def:ulpmmax_fadd}.
Thus, maximizing these two criteria allows one to maximize $\alpha$
over the interval.
\begin{definition}
\label{def:mu-fadd}
Let $\var{z}$ be a variable over $\Fsetsub_{p,\emax}$.
If we have $0 < \lb{z} < \ub{z} < +\infty$,
then $\mufadd{\var{z}} \in [\lb{z},\ub{z}]$ is given by:
\begin{enumerate}

\item
$\mufadd{\var{z}} = \float{1.0\cdots0} {\expf{\ub{z}}}$,
if $\expf{\lb{z}} \neq \expf{\ub{z}}$;

\item
$\mufadd{\var{z}} = \float{b_1.b_2\cdots b_{i-1} a0\cdots0}{\expf{\ub{z}}}$,
if $\expf{\lb{z}} = \expf{\ub{z}}$,
where, for some $b_{i} \neq b'_{i}$:
\begin{align*}
   \lb{z} &= \float{b_1.b_2\cdots b_{i-1} b_{i}\cdots}{\expf{\ub{z}}}; \\
   \ub{z} &= \float{b_1.b_2\cdots b_{i-1} b'_{i}\cdots}{\expf{\ub{z}}}; \\
        a &= \begin{cases}
               0, &\text{if $\float{b_1.b_2\cdots b_{i-1}0\cdots0}{\expf{\ub{z}}}
                             = \lb{z}$;} \\
               1, & \text{otherwise.}
             \end{cases}
\end{align*}
\end{enumerate}
If $0 < \lb{z} = \ub{z} < +\infty$,
then $\mufadd{\var{z}} = \lb{z}$.
If $-\infty < \lb{z} \leq \ub{z} < 0$,
then $\mufadd{\var{z}} \in [\lb{z},\ub{z}]$ is simply defined by
$\mufadd{\var{z}} = -\mufadd{\var{w}}$ where $\var{w} \in[-\ub{z}, -\lb{z}].$
We leave $\mufadd{\var{z}}$ undefined otherwise.
\end{definition}
Note that Definition~\ref{def:mu-fadd} cannot be usefully extended to
intervals containing zeros or infinities, as no interesting bounds can
be derived for $\var{x}$ and $\var{y}$ in such cases.
Consider, for example, the constraint $\var{x} \fadd \var{y} = \var{z}$
with $\var{z} = +0$:
for each $x \in [-\fmax, +\fmax]$ we have $x \fadd -x = +0$.
Hence, when the interval of $\var{z}$ contains zeros or infinities,
only the classical filtering~\citep{Michel02,BotellaGM06} is applied.

\begin{theorem}
\label{thm:mu-fadd}
Let $\var{z}$ be over $\Fsetsub_{p,\emax}$ with
$\lb{z} \notin \{ -\infty, -0, +0 \}$ and
$\ub{z} \notin \{ -0, +0, +\infty \}$ having the same sign.
Then, for each $z \in [\lb{z}, \ub{z}]$,
$\ulpmmax{\fadd}(z) \leq \ulpmmax{\fadd}\bigl(\mufadd{\var{z}}\bigr)$.
\end{theorem}
\proof{Proof.}
Without loss of generality, assume $\lb{z} > 0$.
If $\lb{z} = \ub{z}$ the result holds.
Let us now assume $\lb{z} < \ub{z}$.
We start proving that $\alpha$ and $\beta$ of
Definition~\ref{def:ulpmmax_fadd} computed over
$\mufadd{\var{z}}$ are greater than or equal to
the $\alpha$'s and $\beta$'s computed over any other value in
$[\lb{z}, \ub{z}]$.

We first prove that $\mufadd{\var{z}}$ maximizes $\alpha$.
For $z \in [\lb{z}, \ub{z}]$ we have
\[
  \alpha = \float{1.1 \cdots 1}{\expf{z} + k},
\]
where $k$ is the number of successive $0$'s
to the right of the significand of $z$.
Let us consider the maximum exponent of the values in $\var{z}$,
which is $\expf{\ub{z}}$.
Among the values in $[\lb{z}, \ub{z}]$ with such an exponent,
we want to select the one with the highest number of successive
zeros to the right of the significand.
Since $\lb{z}>0$, the maximum value for $\alpha$ would be attained
by the float $\float{1.0 \cdots 0}{{\expf{\ub{z}}}}$, if this belongs to
$[\lb{z}, \ub{z}]$.  This happens in three cases:
\begin{enumerate}

\item
$\expf{\lb{z}} \neq \expf{\ub{z}}$ and
$\mufadd{\var{z}} = \float{1. 0 \cdots 0}{\expf{\ub{z}}}$,
by the first case of Definition~\ref{def:mu-fadd}.

\item
$\expf{\lb{z}} = \expf{\ub{z}}$
and $\lb{z} = \float{1.0\cdots 0}{\expf{\ub{z}}}$;
in this case we have, again,
$\mufadd{\var{z}} = \float{1. 0 \cdots 0}{\expf{\ub{z}}}$,
so defined by the second case of Definition~\ref{def:mu-fadd};
in fact, for some $i \in \{ 2, \dots, p-1\}$ that depends on $\ub{z}$,
we have
\begin{align*}
  \ub{z} &= \float{1.b_2\cdots b_{i-1}10 \cdots 0}{\expf{\ub{z}}}, \\
  \lb{z} &= \float{1.b_2\cdots b_{i-1}00 \cdots 0}{\expf{\ub{z}}}
\end{align*}
with $b_2 = \cdots = b_{i-1} =0$,
and the algorithm gives
$\float{1.b_2\cdots b_{i-1} a0\cdots0}{\expf{\ub{z}}}$ with $a = 0$,
i.e., $\float{1. 0 \cdots 0}{\expf{\ub{z}}}$.

\item
$\expf{\lb{z}} = \expf{\ub{z}}$,
$\lb{z} = \float{0.b_2\cdots b_p}{\emin}$ and
$\ub{z} = \float{1.b'_2\cdots b'_p}{\emin}$;
thus we have,
$\mufadd{\var{z}} = \float{1. 0 \cdots 0}{\emin}$,
once again by the second case of Definition~\ref{def:mu-fadd}
where $i = 1$, hence $\mufadd{\var{z}} = \float{a.0\cdots 0}{\emin}$.
Moreover, since
$\lb{z} > 0$, necessarily
$\lb{z} \neq \float{0.0\cdots0}{\emin}$ and we must have $a = 1$.
\end{enumerate}
We are now left with the case when
$\float{1.0\cdots 0}{\expf{\ub{z}}} \notin [\lb{z}, \ub{z}]$.
This occurs when
$\expf{\lb{z}} = \expf{\ub{z}}$ but either
$\lb{z} > \float{1.0\cdots 0}{\expf{\ub{z}}}$ or
$\ub{z} < \float{1.0\cdots 0}{\expf{\ub{z}}}$.
In both cases, all the floats in $[\lb{z}, \ub{z}]$ have the same exponent
and the same most significant bit ($b_1$).
Therefore, in order to maximize $\alpha$, we need to choose among them
the one with the greatest number of successive zeros to the right
of the significand.
The first step is to find the index of the most significant
significand bit where $\lb{z}$ and $\ub{z}$ differ:
since $\lb{z} < \ub{z}$, such an index must exist.
Let then
\begin{align}
\notag
  \lb{z} &= \float{b_1.b_2\cdots b_{i-1} b_{i}\cdots}{\expf{\ub{z}}}, \\
\notag
  \ub{z} &= \float{b_1.b_2\cdots b_{i-1} b'_{i}\cdots}{\expf{\ub{z}}}, \\
\intertext{%
where $b_{i} = 0$ and $b'_{i} = 1$ for some $i > 1$.
The significand maximizing $\alpha$ is
$b_1.b_2\cdots b_{i-1}0 \cdots 0$.
Indeed, any float having a significand with a larger
number of consecutive zeros to the right does not belong to
$[\lb{z}, \ub{z}]$.
However, it is not always the case that
$\float{b_1.b_2\cdots b_{i-1}0 \cdots 0}{\expf{\ub{z}}}$
belongs to $[\lb{z}, \ub{z}]$:
we must have
}
\label{eq:thm:mu-fadd-special-lb-z}
  \lb{z} &= \float{b_1.b_2\cdots b_{i-1} b_{i}0\cdots0}{\expf{\ub{z}}}.
\end{align}
If~\eqref{eq:thm:mu-fadd-special-lb-z} is true,
then the second case of Definition~\ref{def:mu-fadd} gives
\begin{align*}
  \mufadd{\var{z}} &= \float{b_1.b_2\cdots b_{i-1} a0\cdots}{\expf{\ub{z}}},
    &\text{with $a = 0$,} \\
\intertext{%
which is indeed equal to~$\lb{z}$.
On the other hand,
if \eqref{eq:thm:mu-fadd-special-lb-z} is false,
then no float with significand $b_1.b_2\cdots b_{i-1}00\cdots0$ belongs
to $[\lb{z}, \ub{z}]$,
hence the significand maximizing $\alpha$ is necessarily the one with
one less zero to the right, i.e., $b_1.b_2\cdots b_{i-1}10\cdots0$,
which is guaranteed to belong to $[\lb{z}, \ub{z}]$.
This is consistent with the second case of Definition~\ref{def:mu-fadd},
which gives
}
  \mufadd{\var{z}} &= \float{b_1.b_2\cdots b_{i-1} a0\cdots0}{\expf{\ub{z}}},
    &\text{with $a = 1$.} \\
\end{align*}

We have proved that Definition~\ref{def:mu-fadd} gives a float
$\mufadd{\var{z}}$ that maximizes the value $\alpha$.
We now prove that $\mufadd{\var{z}}$ also maximizes the value of $\beta$.
By Propositions~\ref{prop:marre-michel-prop-1}
and~\ref{prop:marre-michel-prop-ext}
and Definition~\ref{def:ulpmmax_fadd},
$\beta = \alpha \fadd z$.
Note that $\mufadd{\var{z}}$ maximizes $\alpha$;
however, since $\beta$ also depends on $z$, we have to prove
that no $z \in [\lb{z}, \ub{z}]$ such that $z > \mufadd{\var{z}}$
results into a greater $\beta$.
Observe first that, by construction, $\mufadd{\var{z}}$
has the maximum exponent in $[\lb{z}, \ub{z}]$.
Therefore any $z > \mufadd{\var{z}}$ in $[\lb{z}, \ub{z}]$
must have a larger significand.
Assume that
$\mufadd{\var{z}} = \float{b_1.b_2\cdots b_j0\cdots0}{\expf{\ub{z}}}$
with $b_j = 1$ for some $j \in \{ 1, \dots, p \}$.
The exponent of the corresponding $\alpha$ is $\expf{\ub{z}} + p-j$.
Suppose now there exists $z > \mufadd{\var{z}}$ in $[\lb{z}, \ub{z}]$
with a larger significand:
this must have the form
$\float{b_1.b_2\cdots b_\ell0\cdots0}{\expf{\ub{z}}}$ with $b_\ell = 1$
and $j < \ell \leq p$.
The exponent of the corresponding $\alpha$ is $\expf{\ub{z}} + p - \ell$,
which is smaller than the $\alpha$ computed for $\mufadd{\var{z}}$
by at least one unit.
Hence, we can conclude that
\(
    \float{b_1.b_2\cdots b_j0\cdots0}{\expf{\ub{z}}}
      + \float{1.1 \cdots 1}{\expf{\ub{z}} + p - j}
  >
    \float{b_1.b_2\cdots b_l0\cdots0}{\expf{\ub{z}}}
      + \float{1.1 \cdots 1}{\expf{\ub{z}} + p - \ell}
\),
since $\ell > j$.
This shows that the float $\mufadd{\var{z}}$ also maximizes
the value of $\beta$.
We have proved that Definition~\ref{def:mu-fadd} gives
a float $\mufadd{\var{z}}$ that maximizes the value of both $\alpha$
and $\beta$ over $\var{z}$.
Since Definition~\ref{def:ulpmmax_fadd} defines
$\ulpmmax{\fadd}(z) = \alpha$ for $-\infty < z < 0$
and $\ulpmmax{\fadd}(z) = \beta$ for $0 < z < +\infty$, we can
conclude that,
for each $z \in [\lb{z}, \ub{z}]$,
$\ulpmmax{\fadd}(z) \leq \ulpmmax{\fadd}(\mufadd{\var{z}})$.
\Halmos
\endproof

As we have already pointed out, the algorithm of Definition~\ref{def:mu-fadd},
if restricted to normal numbers, is similar to the algorithm presented in
\citep{MarreM10}.
There is an important difference, though, in the case when
$\lb{z} = \float{b_1.b_2\cdots b_{i-1} b_{i}0\cdots0}{\expf{\ub{z}}}$,
$\ub{z} = \float{b_1.b_2\cdots b_{i-1} b'_{i}\cdots}{\expf{\ub{z}}}$
and $\lb{z} > 0$.
In this case the algorithm of \citet{MarreM10} erroneously returns
$\float{b_1.b_2\cdots b_{i-1}10\cdots0}{\expf{\ub{z}}}$ instead of
the value that maximizes $\alpha$, i.e., $\lb{z}$, which is
correctly computed by our algorithm.

For efficiency reasons, filtering by maximum ULP might be applied
only when $\ulpmmax{\fadd}\bigl(\mufadd{\var{z}}\bigr) \leq \fmax$
so as to avoid the use of wider floating-point formats.

In order to define
$\ulpmmax{\fsub}$, $\ulpmmaxp{\fsub}$,
$\ulpmmin{\fsub}$ and $\ulpmminp{\fsub}$,
we can use the following observation.
Since $x \fsub y = \roundnear{x - y} = \roundnear{x + -y} = x \fadd -y$,
the constraints
$\var{z} = \var{x} \fsub \var{y}$ and
$\var{z} = \var{x} \fadd -\var{y}$ are equivalent.
Thus we have
$\ulpmmax{\fsub} = \ulpmmax{\fadd}$ and
$\ulpmmin{\fsub} = \ulpmmin{\fadd}$, while
$\ulpmmaxp{\fsub} = -\ulpmmin{\fadd}$ and
$\ulpmminp{\fsub} = - \ulpmmax{\fadd}$ since,
if $-y \in [\ulpmmin{\fadd}(z), \ulpmmax{\fadd}(z)]$,
then $y\in [-\ulpmmax{\fadd}(z), -\ulpmmin{\fadd}(z)]$.
Moreover, since $\mufadd{\var{z}}$
maximizes $\ulpmmax{\fadd}$ and
minimizes $\ulpmmin{\fadd}$
over an interval of floating-point values $\var{z}$,
$\mufadd{\var{z}}$ can be used as well to
maximize $\ulpmmaxp{\fsub}$ and
minimize $\ulpmminp{\fsub}$ on $\var{z}$.

\subsection{Filtering by Maximum ULP on Multiplication}
\label{sec:ULP-mul}

For filtering multiplication constraints of the form $z = x \fmul y$
(and similarly for division), we cannot rely on the same maximum
ULP property identified by Marre and Michel upon which the treatment
of addition and subtraction rests.
This is because the ULP property of $z$ is only loosely related to the
ULP property of $x$ and $y$ when they are being multiplied.
Our generalized property, instead, covers also multiplication (and division,
as we will see in Section~\ref{sec:ULP-div}).
As indicated in~\eqref{eq:ulpmmax1new} and~\eqref{eq:ulpmmin1new},
we have to the determine the maximum and minimum values for $x$
satisfying $z = x \fmul y$.

Consider a strictly positive constant $z \in \Fset_{p,\emax}$
and two unknowns $x, y \in \Fsetsub_{p,\emax}$ such that $z = x \fmul y$.
If $z \leq \fmax/\fmin$,
there exists a greatest float $x_\mathrm{m} \in \Fsetsub_{p,\emax}$
such that there exists $y \in \Fsetsub_{p,\emax}$
satisfying $z = x_\mathrm{m} \fmul y$.
More precisely, $x_\mathrm{m}$ must satisfy $z = x_\mathrm{m} \fmul \fmin$ and
it turns out that we can take $x_\mathrm{m} = z \fdiv \fmin$.
Since, for $z \leq \fmax/\fmin$,
division of $z$ by $\fmin = 2^{\emin + 1 - p}$
amounts to an exponent shifting,
we have that $\Fsetsub_{p,\emax} \ni x_\mathrm{m} = z / \fmin$.
Moreover, we have that $x_\mathrm{m} = z/\fmin$ is the greatest float
\ifthenelse{\boolean{TR}}{
such that $z = x_\mathrm{m} \fmul \fmin$.%
\footnote{See the proof of forthcoming
Theorem~\ref{thm:mu-fmul-satisfies-properties} in the Appendix.}
}{ 
such that $z = x_\mathrm{m} \fmul \fmin$.%
\footnote{See the proof of forthcoming
Theorem~\ref{thm:mu-fmul-satisfies-properties}
available in the Online Supplement \citep{BagnaraCGG15IJOC-OS}.}
} 

On the other hand,
there is no other float $y < \fmin$ such that $z = x \fmul y$,
since $y$ must be greater than $+0$, for otherwise
$x \fmul y$ would not be strictly positive.
However, for no $y \in \Fsetsub_{p,\emax}$ we have $+0 < y < \fmin$.
Therefore, the greatest value $x_\mathrm{m}$ such that $z = x_\mathrm{m} \fmul \fmin$
is the greatest value for $x$ that can satisfy $z = x \fmul y$
for some $y \in \Fsetsub_{p,\emax}$.

When dealing with subnormal floating-point numbers a similar argument
applies.
In fact, also in this case there exists a greatest float
$x_\mathrm{m} \in \Fsetsub_{p,\emax}$
satisfying $z = x_\mathrm{m} \fmul y$ for some $y \in \Fsetsub_{p,\emax}$.
As before, such $x_\mathrm{m}$ must satisfy $z = x_\mathrm{m} \fmul \fmin$.
However, it turns out that, when $z$ is subnormal,
there may exist values for $x_\mathrm{m}$ greater than $z/\fmin$
that still satisfy $z = x_\mathrm{m} \fmul \fmin$.
This is because the distance between subnormal numbers, being fixed to
$\fmin$, does not depend on $z$.

Based on the previous reasoning, we can define
$\ulpmmax{\fmul}$ and $\ulpmmin{\fmul}$.

\begin{definition}
\label{def:ulpmmax_fmul}
Let
\(
  \Fset_{\fmul}
     =
       \bigl\{\,
         z \in \Fsetsub_{p,\emax}
       \bigm|
         |z| / \fmin \leq \fmax
       \,\bigr\}
\)
and $\bar{\Fset}_{\fmul} = \Fset_{p,\emax}$.
Then
$\fund{\ulpmmax{\fmul}}{\Fset_{\mathord{\fmul}}}{\bar{\Fset}_{\mathord{\fmul}}}$
is defined, for each $z \in \Fset_{\fmul}$, by
\[
  \ulpmmax{\fmul}(z)
    =
      \begin{cases}
        \phantom{\Bigl(\bigl(}|z| \fdiv \fmin,
          &\text{if $|z| \geq \fnormin$;} \\
        \phantom{\Bigl(}\bigl(|z| \fdiv \fmin\bigr) \fadd 2^{-1},
          &\text{if $0 < |z| < \fnormin$ and $\feven(z)$;} \\
        \Bigl(\bigl(|z| \fdiv \fmin\bigr) \fadd 2^{-1}\Bigr)^-,
          &\text{if $0 < |z| < \fnormin$ and $\fodd(z)$.}
      \end{cases}
\]
\end{definition}

\begin{theorem}
\label{thm:mu-fmul-satisfies-properties}
Function $\ulpmmax{\fmul}$ is well-defined and satisfies~\eqref{eq:ulpmmax1}
and~\eqref{eq:ulpmmax2}.
\end{theorem}
\proof{Proof.}
\ifthenelse{\boolean{TR}}{
Given in the Appendix.
}{ 
Given in the Online Supplement \citep{BagnaraCGG15IJOC-OS}.
} 
\endproof

A monotonicity property of $\ulpmmax{\fmul}$ simplifies the identification
an element of the interval $\var{z}$ that maximizes the value of
$\ulpmmax{\fmul}$ over $\var{z}$.
\begin{proposition}
\label{prop:mu-fmul-monotonicity}
Let $z \in \Fset_{\fmul}$ be nonzero.
If $z > 0$, then $\ulpmmax{\fmul}(z^+) \geq \ulpmmax{\fmul}(z)$;
on the other hand,
if $z < 0$, then $\ulpmmax{\fmul}(z^-) \geq \ulpmmax{\fmul}(z)$.
\end{proposition}
\proof{Proof.}
\ifthenelse{\boolean{TR}}{
Given in the Appendix.
}{ 
Given in the Online Supplement \citep{BagnaraCGG15IJOC-OS}.
} 
\endproof

Since $\fmul$ is commutative, $\ulpmmaxp{\fmul}=\ulpmmax{\fmul}$, and
the same bounds can be used to filter both $x$ and $y$ in the
constraint $z = x \fmul y$.

The function
$\fund{\ulpmmin{\fmul}}{\Fset_{\mathord{\fmul}}}{\bar{\Fset}_{\mathord{\fmul}}}$
is defined dually:
for each $z \in \Fset_{\fmul} \setminus \{ -0, +0\}$,
$\ulpmmin{\fmul}(z) = -\ulpmmax{\fmul}(z)$.
We can see that properties~\eqref{eq:ulpmmax1}
and~\eqref{eq:ulpmmax2} of $\ulpmmax{\fmul}$ entail
property~\eqref{eq:ulpmmin1new} of $\ulpmmin{\fmul}$.
Again, since $\fmul$ is commutative
we have $\ulpmminp{\fmul} = \ulpmmin{\fmul}$.

Thanks to Proposition~\ref{prop:mu-fmul-monotonicity}
we know that the value $M \in [\lb{z}, \ub{z}]$
that maximizes $\ulpmmax{\fmul}$ is
the one with the greatest absolute value,
i.e., $M = \max \bigl\{ |\lb{z}|, |\ub{z}| \}$.
Since $\ulpmmin{\fmul}$ is defined as $-\ulpmmax{\fmul}(z)$, the
value that minimizes $\ulpmmin{\fmul}$ is again $M$.
Hence, if $[\lb{z}, \ub{z}]$ does not contain zeros,
$\ulpmmax{\fmul}(M)$ (resp., $\ulpmmin{\fmul}(M)$)
is an upper bound (resp., a lower bound) of $\var{x}$
with respect to the constraint $\var{z} = \var{x} \fmul \var{y}$.

The restriction to intervals $\var{z}$ not containing zeros
is justified by the fact that, e.g., if $z = 0$ then
$z = x \fmul y$ holds with $x = \fmax$ and $y = 0$, hence, in this case,
no useful filtering can be applied to $x$.
The same thing happens when
$\max \bigl\{ |\lb{z}|, |\ub{z}| \} / \fmin > \fmax$.
Moreover, whenever the interval of $\var{y}$ does not contain zeros,
filtering by maximum ULP for multiplication, in order to refine $\var{x}$,
is subsumed by the standard indirect projection,
which, in this case, can usefully exploit the information on $\var{y}$.
In contrast, when the interval of $\var{y}$ does contain
zeros, our filter is able to derive bounds that cannot be obtained with
the standard indirect projection, which, in this case, does not allow
any refinement of the interval.
Thus, for multiplication (and, as we will see, for division as well),
the standard indirect projection and filtering by maximum ULP are
mutually exclusive: one applies when the other cannot derive anything useful.
Commenting on a previous version of the present paper, Claude Michel observed
that one could modify the standard indirect projections with interval
splitting so that indirect projections are always applied to source
intervals not containing zeros.
This idea rests on the observation that, for $\var{z} = \var{x} \odot \var{y}$
with $\odot \in \{\otimes, \oslash\}$,
when the interval of $\var{z}$ is a subset of the finite non zero floats
neither $\var{x}$ nor $\var{y}$ do have any support for
$\pm0$ and $\pm\infty$.
For multiplication, ordinary standard indirect projection would be
modified as follows,
assuming that $\var{z}$ is positive and that we want to apply the standard
indirect projection to $\var{z}$ and $\var{y}$ in order to refine $\var{x}$
(the other cases being similar):
\begin{itemize}
\item
we apply the ordinary standard indirect projection to $\var{z}$ and
$\var{y} \cap [-\fmax, -\fmin]$,
intersecting the resulting interval with $[-\fmax, -\fmin]$;
\item
we apply the ordinary standard indirect projection to $\var{z}$ and
$\var{y} \cap [\fmin, \fmax]$,
intersecting the resulting interval with $[\fmin, \fmax]$;
\item
finally, we use the convex union of the two intervals so computed to
refine $\var{x}$.
\end{itemize}
We believe that, when the applied ordinary (i.e., non-splitting)
standard indirect projection is as precise as the one specified
by \citet{Michel02},
the refining interval computed for $\var{x}$
by the modified procedure is very precise:
it coincides with the result of the ordinary standard indirect projection,
when $0 \notin \var{y}$ and thus filtering by maximum ULP is not applicable,
or it coincides with the result of filtering by maximum ULP, when
$0 \in \var{y}$ and therefore the ordinary standard
indirect projection would not help.\footnote{We are indebted to Claude Michel
for this observation.}
This approach has the advantage to be applicable to any rounding mode.
On the other hand the standard indirect projections specified
in~\citep{Michel02} require working on rationals or on larger
floating-point formats, whereas one of our aims is to always work with machine
floating-point numbers of the same size of those used in the analyzed
computation.

\begin{example}
\label{ex:mu-fmul}
Consider the IEEE~754 single-precision constraint
$\var{z} = \var{x} \fmul \var{y}$ with $\var{z}$ subnormal,
\(
   \var{z} \in [-\float{0.00000000000000010001001}{-126},
                -\float{0.00000000000010000000000}{-126}]
\),
and $\var{x}$ and $\var{y}$ unconstrained,
$\var{x}, \var{y} \in [\fminf, \fpinf]$.
Our indirect projection infers the constraints
\(
  \var{x}, \var{y} \in [-\float{1.00000000001}{10},
                         \float{1.00000000001}{10}]
\),
while classical inverse projections do not allow pruning
the intervals for $x$ and $y$, no matter what they are.
\end{example}

\subsection{Filtering by Maximum ULP on Division}
\label{sec:ULP-div}

We now define filtering by maximum ULP for floating-point
constraints of the form $\var{z} = \var{x} \fdiv \var{y}$.
We begin defining the first indirect projection.
We will then tackle
the problem of defining the second indirect projection, which,
as we will see, is significantly more involved than the first one:
the solution we propose is new to this paper.

\subsubsection{The First Indirect Projection}
\label{sec:first-indirect-division}

A role similar to the one of $\fmin $ in the definition of filtering
by maximum ULP on multiplication is played by $\fmax$
in the definition of the first indirect projection for division.

\begin{definition}
\label{def:mu-fdiv}
Let us define the sets
\(
 \Fset'_{\fdiv} = \bigl\{\,
                  z \in \Fsetsub_{p,\emax}
                \bigm|
                  |z| \fmul \fmax \leq \fmax
                \,\bigr\}
\)
and $\bar{\Fset}'_{\fdiv} = \Fset_{p,\emax}$.
Let also $q = 1 - p + \emin + \emax$.\footnote{In the very common case
where $\emin = 1 - \emax$ we have $q = 2 - p$.}
Then
$\fund{\ulpmmax{\fdiv}}{\Fset'_{\mathord{\fdiv}}}{\bar{\Fset}'_{\mathord{\fdiv}}}$
is defined, for each $z \in \Fset'_{\fdiv}$, by
\[
  \ulpmmax{\fdiv}(z)
    =
      \begin{cases}
        \phantom{\Bigl(\bigl(}|z| \fmul \fmax,
          &\text{if $\fnormin \leq |z| \leq 1$;} \\
        \phantom{\Bigl(}\bigl(|z| \fmul \fmax\bigr) \fadd 2^q,
          &\text{if $0 \leq |z| < \fnormin$} \\
          &\text{$\quad \mathord{}
                   \land \bigl(|z| \neq \float{1}{\expf{z}}
                               \lor \expf{z} = \emin-1\bigr)$;} \\
        \Bigl(\bigl(|z| \fmul \fmax\bigr) \fadd 2^q\Bigr)^-,
          &\text{ otherwise.}
      \end{cases}
\]
\end{definition}

Observe that we have $|z| \fmul \fmax \leq \fmax$ if and only if $|z| \leq 1$.
In fact, for $z = 1^+ = 1 + 2^{1-p}$, we obtain
\begin{align}
\notag
  |z| \fmul \fmax
    &= (1 + 2^{1-p}) \fmul \fmax \\
\notag
    &= \bigroundnear{(1 + 2^{1-p}) \fmax} \\
\notag
    &= \bigroundnear{\fmax + (2 - 2^{1-p})2^{\emax+1-p}} \\
\label{eq:after-def:mu-fdiv-1}
    &= +\infty,
\end{align}
where~\eqref{eq:after-def:mu-fdiv-1} holds
by Definition~\ref{def:round-to-nearest-tail-to-even},
since $(2 - 2^{1-p})2^{\emax+1-p} > \ferrdown{\fmax}/2 = 2^{\emax-p}$.
By monotonicity of $\fmul$ we can conclude that $z \in \Fset'_{\fdiv}$
if and only if $|z| \leq 1$.

\begin{theorem}
\label{thm:mu-fdiv-satisfies-properties}
$\ulpmmax{\fdiv}$ is well-defined and
satisfies~\eqref{eq:ulpmmax1} and~\eqref{eq:ulpmmax2}.
\end{theorem}
\proof{Proof.}
\ifthenelse{\boolean{TR}}{
Given in the Appendix.
}{ 
Given in the Online Supplement \citep{BagnaraCGG15IJOC-OS}.
} 
\endproof

The function $\ulpmmin{\fdiv}$ is defined,
for each $z \in \Fset'_{\fdiv}$, by $\ulpmmin{\fdiv} = -\ulpmmax{\fdiv}(z)$.

As for multiplication, a monotonicity property of $\ulpmmax{\fdiv}$
enables quickly identifying the value of $\var{z}$ that maximizes the
function.
\begin{proposition}
Let $z \in \Fset_{\fdiv}$ be nonzero.
If $z > 0$, then $\ulpmmax{\fdiv}(z^+) \geq \ulpmmax{\fdiv}(z)$;
on the other hand,
if $z < 0$, then $\ulpmmax{\fdiv}(z^-) \geq \ulpmmax{\fdiv}(z)$.
\end{proposition}
\proof{Proof.}
\ifthenelse{\boolean{TR}}{
Given in the Appendix.
}{ 
Given in the Online Supplement \citep{BagnaraCGG15IJOC-OS}.
} 
\endproof

By monotonicity, the value $M \in [\lb{z}, \ub{z}]$ that maximizes
$\ulpmmax{\fdiv}$ is the one that has the greatest
absolute value, i.e., $M = \max \bigl\{ |\lb{z}|, |\ub{z}| \bigr\}$.
Since $\ulpmmin{\fdiv}$ is defined as $-\ulpmmax{\fdiv}(z)$,
$M$ is also the value that minimizes $\ulpmmin{\fdiv}$.
Hence, if $[\lb{z}, \ub{z}]$ does not contain zeros,
$\ulpmmax{\fdiv}(M)$ (resp., $\ulpmmin{\fdiv}(M)$)
is an upper bound (resp. a lower bound) of $\var{x}$ with respect to
the constraint $\var{z} = \var{x} \fdiv \var{y}$.
The restriction to intervals not containing zeros
is justified by the fact that, e.g.,
if $z = 0$ then $z = x \fdiv y$ holds with $x = \fmax$ and $y = \infty$;
hence, in this case, no useful filtering can be applied to $x$.
The same happens when
$\max \bigl\{ |\lb{z}|, |\ub{z}| \} \fmul \fmax > \fmax$.
In addition, the same phenomenon we saw for multiplication manifests
itself here:
whenever the interval of the variable $\var{y}$ does not
contain infinities, filtering by maximum ULP for division in order
to refine $\var{x}$ is subsumed by the standard indirect projection.
On the other hand, when the interval of $\var{y}$
does contain infinities, the standard indirect projection gives nothing
whereas filtering by maximum ULP provides nontrivial bounds.
Thus, the standard indirect projection and filtering by maximum ULP
for division are mutually exclusive: one applies when the other cannot
derive anything useful.
And, just as for multiplication, if using rationals or extended
floating-point formats is an option, then a pruning variant
(one that cuts off infinities) of the
indirect projection specified in~\citep{Michel02} will be equally precise.
\begin{example}
\label{ex:mu-fdiv1}
Consider the IEEE~754 single-precision constraint
$\var{z} = \var{x} \fdiv \var{y}$ with initial intervals
$\var{z} \in [-\float{1.0}{-110}, -\float{1.0}{-121}]$ and
$\var{x}, \var{y} \in [\fminf, \fpinf]$.
We have
\begin{align*}
  \ulpmmax{\fdiv}(\float{1.0}{-110})
    &= \float{1.0}{-110} \cdot \float{1.1\cdots 1}{127}\\
    &= \float{1.1\cdots1}{17}, \\
  \ulpmmin{\fdiv}(\float{1.0}{-110})
    &= -\float{1.0}{-110} \cdot \float{1.1\cdots 1}{127}\\
    &= -\float{1.1\cdots1}{17}.
\end{align*}
Filtering by maximum ULP improves upon classical filtering,
which would not restrict any interval,
with $\var{x} \in [-\float{1.1\ldots1}{17},\float{1.1\ldots1}{17}]$.

For an example involving subnormals, consider
$\var{z} = \var{x} \fdiv \var{y}$ with initial interval for $\var{z}$
equal to $[\float{0.00000000000000000000001}{-126}, \float{0.01}{-126}]$
and $\var{x}, \var{y} \in [\fminf, \fpinf]$:
our algorithm produces the constraint
\(
  \var{x} \in [-\float{1.00000000000000000000001}{-46},
                \float{1.00000000000000000000001}{-46}]
\)
whereas classical filtering is unable to infer anything on $x$.
\end{example}

\subsubsection{The Second Indirect Projection}

The discussion in Section~\ref{sec:first-indirect-division}
shows that, for $|z| \leq 1$, we have $\ulpmmaxp{\fdiv}(z) = \fmax$.
We thus need to study $\ulpmmaxp{\fdiv}(z)$ for $|z| > 1$.
It turns out that, due to rounding, the restriction of
$\ulpmmaxp{\fdiv}$ over that subdomain is not a simple function.
Given $z \in \Fsetsub_{p,\emax}$, $\ulpmmaxp{\fdiv}(z)$ is the maximum $y$
such that $x \fdiv y = z$.
Note that, in order to maximize $y$, $x$ must be maximized as well.
A qualitative reasoning on the reals tells us that, since
$\fmax / (\fmax / z) = z$, $y$ should be roughly equal to $\fmax / |z|$.
Indeed, it can be proved that, for $|z| > 1$,
$\fmax\fdiv\bigl(\fmax\fdiv|z|\bigr) $ is equal to $z$, $z^-$ or $z^+$
depending on the value of $z$.
This allows the determination of a rather small upper bound to the values
that $\var{z}$ may take,
which is ultimately our goal for filtering $\var{y}$ values.
To this aim we define the function $\upperdiv$.

\begin{definition}
\label{def:upperbound_div}
The function $\fund{\upperdiv}{\Fsetsub_{p,\emax}}{\Fset^+_{p,\emax}}$
is defined, for each $z \in \Fsetsub_{p,\emax}$, as follows:
\[
  \upperdiv(z)
    =
      \begin{cases}
        \fmax \fdiv |z|^{-\,-},
          &\text{if $1^+ < |z| \leq \fmax$;} \\
        \fmax,
          &\text{otherwise.}
       \end{cases}
\]
\end{definition}
It turns out that $\upperdiv(z)$ satisfies the dual of
Property~\eqref{eq:ulpmmax2}, i.e., it is a correct upper bound,
while it does not satisfy the dual of Property~\eqref{eq:ulpmmax1},
i.e., smaller correct upper bounds might exist.

\begin{theorem}
\label{th:upperbound}
Let $\Fset''_{\fdiv} = \Fsetsub_{p,\emax}$
and $\bar{\Fset}''_{\fdiv} = \Fset^+_{p,\emax}$.
Let
\(
  \fund{\ulpmmaxp{\fdiv}}{\Fset''_{\mathord{\fdiv}}}
                         {\bar{\Fset}''_{\mathord{\fdiv}}}
\)
be a function satisfying~\eqref{eq:ulpmmaxp1new}.
Then, for $0 < |z| \leq 1^+$ or $z = +\infty$,
$\ulpmmaxp{\fdiv}(z) \leq \upperdiv(z)$;
moreover, for $1^+ < |z| \leq \fmax$,
$\ulpmmaxp{\fdiv}(z) < \upperdiv(z)$.
\end{theorem}
\proof{Proof.}
\ifthenelse{\boolean{TR}}{
Given in the Appendix.
}{ 
Given in the Online Supplement \citep{BagnaraCGG15IJOC-OS}.
} 
\endproof

Dually, a lower bound for the function $\ulpmminp{\fdiv}$ can be
obtained by means of the function $\lowerdiv$,
defined by $\lowerdiv(z) = -\upperdiv(z)$.

The value $N \in [\lb{z}, \ub{z}]$ that maximizes $\upperdiv$
is the one that has the smallest
absolute value, i.e., $N = \min \bigl\{ |\lb{z}|, |\ub{z}| \bigr\}$.
Since $\lowerdiv$ is defined as $-\upperdiv(z)$,
$N$ is also the value that minimizes $\lowerdiv$.
Thus, if $[\lb{z}, \ub{z}]$ does not contain zeros,
$\upperdiv(N)$ (resp., $\lowerdiv(N)$)
is an upper bound (resp. a lower bound) for $\var{x}$
with respect to the constraint
$\var{z} = \var{x} \fdiv \var{y}$.
The restriction to intervals not containing zeros
is justified by the fact that if, e.g.,
$z = 0$, then the equality $z = x \fdiv y$ holds with $y = \infty$
for each $x$ such that $0 \leq x \leq \fmax$.
Hence, as in the case of the first projection,
no useful filtering can be applied to $\var{y}$.
Analogously to the case of the filter for the first projection,
this filter is useful whenever the interval of $\var{x}$
contains infinities.
In this case, in fact, it is able to derive useful bounds for $\var{y}$
where the standard indirect projection does not allow any refinement
of the interval.
Just as is the case for multiplication and the first indirect projection
of division,
the standard indirect projection and filtering by maximum ULP are
mutually exclusive: one applies when the other cannot derive anything useful.

Note that, only for this projection, we have chosen to compute a (very small)
upper bound that, in general, is not the least upper bound.
We did so in order to trade precision for efficiency: this way we have
an algorithm that only uses floating-point machine arithmetic operations
on the same format used by the analyzed constraint
$\var{z} = \var{x} \fdiv \var{y}$.
When using rationals or larger floating-point formats is an option,
a pruning variant (as in the previous case, one that cuts off infinities)
of a second indirect projection satisfying the precision
constraints set forth in \citep{Michel02} may result in extra precision
at a higher computational cost.

\begin{example}
\label{ex:mu-fdiv2}
Consider the IEEE~754 single-precision division constraint
$\var{z} = \var{x} \fdiv \var{y}$ with initial intervals
$\var{z} \in [\float{1.0 \cdots 010}{110}, \float{1.0}{121}]$ and
$\var{x}, \var{y} \in [\fminf, \fpinf]$.
We have
\begin{align*}
  \upperdiv(\float{1.0\cdots01}{110})
    &= \float{1.1\cdots 1}{127}
         \fdiv \bigl((\float{1.0\cdots01}{110})^-\bigr)^- \\
    &= \float{1.1\cdots 1}{127}\fdiv \float{1.1\cdots1}{109} \\
    &= \float{1.0}{18}, \\
  \lowerdiv(\float{1.0\cdots01}{110})
    &= -\float{1.1\cdots 1}{127}
          \fdiv \bigl((\float{1.0\cdots01}{110})^-\bigr)^- \\
    &= -\float{1.0}{18}.
\end{align*}
Filtering by maximum ULP improves upon classical filtering,
which gives nothing, with the constraint
$\var{y} \in [-\float{1.0}{18}, \float{1.0}{18}]$.
\end{example}

\subsection{Synthesis}
\label{sec:ULP-synthesis}

Table~\ref{tab:max-ulp-synopsis} provides a compact presentation
of filtering by maximum ULP.

\begin{landscape}
\begin{table}
\caption{Filtering by maximum ULP synopsis}
\label{tab:max-ulp-synopsis}
\bigskip
\begin{center}
\renewcommand*\arraystretch{1.8}
\begin{tabular}{>{$}l<{$}|>{$}l<{$}|>{$}l<{$}|l}
\multicolumn{1}{c|}{Constraint}
  & \multicolumn{1}{c|}{$\var{x} \subseteq \mathord{\cdot}$}
  & \multicolumn{1}{c|}{$\var{y} \subseteq \mathord{\cdot}$}
  & \multicolumn{1}{c}{Condition(s)} \\
\hline
\var{z} = \var{x} \fadd \var{y}, \; 0 < \var{z} \leq \fmax
  & [\phantom{-}\ulpmmin{\fadd}(\zeta\phantom{'}),
     \phantom{-}\ulpmmax{\fadd}(\zeta\phantom{'})]
  & [\phantom{-}\ulpmmin{\fadd}(\zeta\phantom{'}),
     \phantom{-}\ulpmmax{\fadd}(\zeta\phantom{'})]
  & $\zeta = \mufadd{\var{z}}$, \;
    $-\fmax \leq \ulpmmin{\fadd}(\zeta)$, \;
    $\ulpmmax{\fadd}(\zeta) \leq \fmax$ \\
\var{z} = \var{x} \fadd \var{y}, \; -\fmax \leq \var{z} < 0
  & [-\ulpmmax{\fadd}(\zeta'),
     -\ulpmmin{\fadd}(\zeta')]
  & [-\ulpmmax{\fadd}(\zeta'),
     -\ulpmmin{\fadd}(\zeta')]
  & $\zeta' = \mufadd{-\var{z}}$, \;
    $-\fmax \leq \ulpmmin{\fadd}(\zeta')$, \;
    $\ulpmmax{\fadd}(\zeta') \leq \fmax$ \\
\var{z} = \var{x} \fsub \var{y}, \; 0 < \var{z} \leq \fmax
  & [\phantom{-}\ulpmmin{\fadd}(\zeta\phantom{'}),
     \phantom{-}\ulpmmax{\fadd}(\zeta\phantom{'})]
  & [-\ulpmmax{\fadd}(\zeta\phantom{'}),
     -\ulpmmin{\fadd}(\zeta\phantom{'})]
  & $\zeta = \mufadd{\var{z}}$, \;
    $-\fmax \leq \ulpmmin{\fadd}(\zeta)$, \;
    $\ulpmmax{\fadd}(\zeta) \leq \fmax$ \\
\var{z} = \var{x} \fsub \var{y}, \; -\fmax \leq \var{z} < 0
  & [-\ulpmmax{\fadd}(\zeta'),
     -\ulpmmin{\fadd}(\zeta')]
  & [\phantom{-}\ulpmmin{\fadd}(\zeta'),
     \phantom{-}\ulpmmax{\fadd}(\zeta')]
  & $\zeta' = \mufadd{-\var{z}}$, \;
    $-\fmax \leq \ulpmmin{\fadd}(\zeta')$, \;
    $\ulpmmax{\fadd}(\zeta') \leq \fmax$ \\
\var{z} = \var{x} \fmul \var{y}, \; |z|\leq 2^{2-p}(2-2^{1-p})
  & [\phantom{-}\ulpmmin{\fmul}(m),
     \phantom{-}\ulpmmax{\fmul}(m)]
  & [\phantom{-}\ulpmmin{\fmul}(m),
     \phantom{-}\ulpmmax{\fmul}(m)]
  & $m = \max \bigl\{ |\lb{z}|, |\ub{z}| \bigr\}$ \;
     \\
\var{z} = \var{x} \fdiv \var{y}, \; 0 < |\var{z}| \leq 1
  & [\phantom{-}\ulpmmin{\fdiv}(m),
     \phantom{-}\ulpmmax{\fdiv}(m)]
  & [- \fmax,
     \phantom{++} + \fmax]
  & $m = \max \bigl\{ |\lb{z}|, |\ub{z}| \bigr\}$ \\
  \var{z} = \var{x} \fdiv \var{y}, \; 1< |\var{z}| \leq \fmax
  &
  & [\phantom{-}\lowerdiv(n),
     \phantom{-}\upperdiv(n)]
  & $n = \min \bigl\{ |\lb{z}|, |\ub{z}| \bigr\}$ \\

\hline
\end{tabular}
\end{center}
\vfill
\begin{align*}
  \ulpmmax{\fadd}(z)
    &=
      \begin{cases}
        \beta,  &\text{if $0 < z < +\infty$,} \\
        \alpha, &\text{if $-\infty < z < 0$;}
      \end{cases}
                          &\ulpmmin{\fadd}(z) &= -\ulpmmax{\fadd}(-z); \\
  \ulpmmax{\fmul}(z)
    &=
      \begin{cases}
        \phantom{\Bigl(\bigl(}|z| \fdiv \fmin,
          &\text{if $z \geq \fnormin$;} \\
        \phantom{\Bigl(}\bigl(|z| \fdiv \fmin\bigr) \fadd 2^{-1},
          &\text{if $0 < z < \fnormin$ and $\feven(z)$;} \\
        \Bigl(\bigl(|z| \fdiv \fmin\bigr) \fadd 2^{-1}\Bigr)^-,
          &\text{if $0 < z < \fnormin$ and $\fodd(z)$;}
      \end{cases}
                          &\ulpmmin{\fmul}(z) &= -\ulpmmax{\fmul}(z); \\
  \ulpmmax{\fdiv}(z)
    &=
      \begin{cases}
        \phantom{\Bigl(\bigl(}|z| \fmul \fmax,
          &\text{if $\fnormin \leq |z| \leq 1$;} \\
        \phantom{\Bigl(}\bigl(|z| \fmul \fmax\bigr) \fadd 2^q,^{(\ast)}
          &\text{if $0 \leq |z| < \fnormin
                       \land \bigl(|z| \neq \float{1}{\expf{z}}
                                   \lor \expf{z} = \emin-1\bigr)$;} \\
        \Bigl(\bigl(|z| \fmul \fmax\bigr) \fadd 2^q\Bigr)^-,
          &\text{otherwise;}
      \end{cases}
                          &\ulpmmin{\fdiv}(z) &= -\ulpmmax{\fdiv}(z); \\
  \upperdiv(z)
    &=
      \begin{cases}
        \fmax \fdiv |z|^{-\,-},
          &\text{if $1^+ < |z| \leq \fmax$;} \\
        \fmax,
          &\text{otherwise;}
       \end{cases}
                          & \lowerdiv(z)      &= -\upperdiv(z);
\end{align*}
\vfill
$(\ast) \quad q = 1 - p + \emin + \emax$.
\end{table}
\end{landscape}

\section{Discussion}
\label{sec:discussion}

This work is part of a long-term research effort concerning the
correct, precise and efficient handling of floating-point constraints
\citep{BelaidMR12,Belaid13TH,BotellaGM06,CarlierG11,MichelRL01,Michel02,MarreM10}
for software verification purposes.

Restricting the attention to test data generation
other authors have considered using search-based techniques
with a specific notion of distance in their fitness
function \citep{LakhotiaHG10,LakhotiaTHDH10}.
For instance, search-based tools like AUSTIN and FloPSy can generate
a test input for a specific path by evaluating the path covered by some
current input with respect to a targeted path in the program.
However, they cannot \emph{solve} the constraints of path conditions,
since: 1) they cannot determine unsatisfiability when the path is infeasible,
and 2) they can fail to find a test input while the set of constraints
is satisfiable \citep{BagnaraCGG13ICST}.

Recently, \citet{BorgesAABP12} combined a search-based test
data generation engine with the RealPaver \citep{GranvilliersB06}
interval constraint solver,
which is well-known in the Constraint Programming community.
Even though constraint solvers over continuous domains
(e.g., RealPaver \citep{GranvilliersB06},
Quimper \citep{ChabertJ09} or ICOS \citep{Lebbah09})
and the work described in the present paper are based on similar
principles, the treatment of intervals is completely different.
While our approach preserves all the solutions over the floats,
it is not at all concerned with solutions over the reals.
In contrast, RealPaver preserves solutions over the reals
by making the appropriate choices in the rounding modes used for
computing the interval bounds, but RealPaver can lose solutions over
the floats.  For instance, a constraint like
$(\var{x} > 0.0 \land \var{x} \oplus 10000.0 \leq 10000.0)$
is shown to be unsatisfiable on the reals by RealPaver,
while it is satisfied by many IEEE~754 floating-point values
of single or double precision format for $\var{x}$ \citep{BotellaGM06}.
Note that RealPaver has recently been used to tackle test input
generation in presence of transcendental functions
\citep{BorgesAABP12}, but this approach,
as mentioned by the authors of the cited paper,
is neither correct nor complete due to the error
rounding of floating-point computations.

The CBMC model checker \citep{ClarkeKL04} supports floating-point
arithmetic using a bit-precise floating-point decision procedure
based on propositional encoding.  According to \citep{DSilvaHKT12}
``CBMC translates the floating-point arithmetic to large propositional
circuits which are hard for SAT solvers.''
The technique presented in this paper is orthogonal to decision
procedures over floating-point computations, such as those
used in CBMC \citep{ClarkeKL04} or CDFL \citep{DSilvaHKT12}.
Of course, implementing the filtering procedures suggested here
would require dedicated bitwise encodings, but this would
enable to perform more constraint-based reasoning over these
computations.

\section{Conclusion}
\label{sec:conclusion}

This paper concerns constraint solving over binary floating-point numbers.
Interval-based consistency techniques are very effective for the
solution of such numerical constraints, provided precise and efficient
filtering algorithms are available.
We reformulated and corrected the filtering algorithm proposed by
\citet{MarreM10} for addition and subtraction.
We proposed a uniform framework that generalizes the property identified
by Marre and Michel to the case of multiplication and division.
We also revised, corrected and extended our initial ideas, sketched in
\citet{CarlierG11}, to subnormals and to the effective treatment
of floating-point division.
All algorithms have been proved correct.
In order to gain further confidence on the algorithms, we have
exhaustively tested a first prototype, symbolic implementation
on floating-point numbers with a small number of bits (e.g.,
$p = 6$ and $\emax = 3$).
The implementation working on IEEE~754 formats was also
tested with a variety of methodologies with the help of test-suites
like the one by the \citet{IBM-FPTS-2008}.

An important objective of this work has been to allow maximum
efficiency by defining all algorithms in terms of IEEE~754 elementary
operations on the same formats as the ones of the filtered constraints.
Indeed, the computational cost of filtering by maximum ULP as defined
in the present paper and properly implemented is negligible.
In fact, all the filters defined in this paper can be directly translated
into constant-time algorithms (as IEEE 754 formats have fixed size)
based on IEEE 754 elementary operations and simple bitwise manipulations.
Moreover, for multiplication and division, precise conditions are given
in order to decide whether standard filtering or our filtering by maximum ULP
is applied: it is one or the other, never both of them.
As shown in \citep{BagnaraCGG13ICST}, the improvement of filtering
procedures with these techniques brings significant speedups of the overall
constraint solving process, with only occasional, negligible slowdowns.
Note that  the choice of different heuristics concerning the selection
of constraints and variables to subject to filtering and the labeling
strategy has a much more dramatic effect on solution time, even though
the positive or negative effects of such heuristics change wildly from
one analyzed program to the other.
Filtering by maximum ULP contributes to reducing this variability.
To understand this, consider the elementary constraint
$\var{z} = \var{x} \fop \var{y}$: if $\var{x}$ and $\var{y}$ are subject
to labeling before $\var{z}$, then filtering by maximum ULP will not help.
However, $\var{z}$ might be labeled before $\var{x}$ or $\var{y}$:
this can happen under \emph{any} labeling heuristic and constitutes
a performance bottleneck.
In the latter case, filtering by maximum ULP may
contribute to a much improved pruning of the domains
of $\var{x}$ and $\var{y}$ and remove the bottleneck.

Future work includes coupling filtering by maximum ULP with sophisticated
implementations of classical filtering based on multi-intervals and
with dynamic linear relaxation algorithms \citep{DenmatGD07} using
linear relaxation formulas such as the ones proposed by \citet{BelaidMR12}.
Another extension, by far more ambitious, concerns the correct handling of
transcendental functions (i.e., $\sin$, $\cos$, $\exp$, \dots):
as IEEE~754 only provides recommendations
rather than formal requirements for these functions,
solutions will be dependent on the particular implementation
and/or be imprecise; in other words, generated test inputs will not
be applicable to other implementations and/or may fail to exercise
the program paths they were supposed to traverse.


\ifthenelse{\boolean{TR}}{

}{ 

} 

\ACKNOWLEDGMENT{%
We are grateful to Abramo Bagnara (BUGSENG srl, Italy) for the many
fruitful discussions we had on the subject of this paper, and
to Paul Zimmermann (INRIA Lorraine, France) for the help he gave
us proving a crucial result.  We are also indebted to Claude Michel
for several constructive remarks that allowed us to improve the paper.
Finally, we wish to express our gratitude to the anonymous reviewers
for the many useful suggestions they contributed.
}

%
%
%

\ifthenelse{\boolean{TR}}{
\clearpage
\begin{APPENDIX}{Technical Proofs}

\setcounter{theorem}{2}
\begin{theorem}
Function $\ulpmmax{\fmul}$ is well-defined and satisfies~\eqref{eq:ulpmmax1}
and~\eqref{eq:ulpmmax2}.
\end{theorem}
\proof{Proof.}
First note that $\Fset_{\fmul}$ is the set of all $z \in \Fsetsub_{p,\emax}$
such that
\[
  |z| \leq \fmax \cdot \fmin = (2-2^{1-p})2^{\emax+\emin+1-p}
\]
and that the range of $\ulpmmax{\fmul}$ is the positive subset of
$\Fset_{p,\emax}$.
This is because its domain is $\Fset_{\fmul}$
and multiplication by $2^{-(\emin+1-p)}$, for $z \in \Fset_{\fmul}$,
boils down to summing exponents.
Moreover,
$\bigl(|z|/\fmin\bigr) \fadd 2^{-1} = |z|/\fmin + 2^{-1}$.
In fact, let $|z| = m 2^{\expf{z}}$ for some $1 \leq m < 2$.
We have
\begin{equation}
\label{ineq:denormalized-significand}
  m < 2 - 2^{\emin-\expf{z}} 2^{1-p},
\end{equation}
since $z$ is subnormal and $m$ is a normalized significand.
Hence,
\begin{align}
\notag
  \bigl(|z|/\fmin\bigr) \fadd 2^{-1}
    &= \roundnear{m 2^{\expf{z}}/\fmin + 2^{-1}} \\
\notag
    &= \roundnear{m 2^{\expf{z} - \emin - 1 + p} + 2^{-1}} \\
\notag
    &= \bigroundnear{(m + 2^{\emin - \expf{z} - 1}2^{1 - p} )2^{\expf{z} - \emin - 1 + p}} \\
\label{eq:mu-fmul-sub}
    &= (m + 2^{\emin - \expf{z} -1}2^{1 - p})2^{\expf{z} - \emin - 1 + p}  \\
\notag
    &= |z|/\fmin +2^{-1},
\end{align}
where~\eqref{eq:mu-fmul-sub} holds
because of~\eqref{ineq:denormalized-significand}.

Consider now the following cases:
\begin{description}

\item[$\fnormin \leq z \leq (2-2^{1-p})2^{\emax+\emin+1-p}:$]
We have $\ulpmmax{\fmul}(z)= |z| 2^{-(\emin+1-p)}$,
hence $y = \fmin = 2^{\emin+1-p}$ satisfies~\eqref{eq:ulpmmax1}:
\begin{align}
\notag
  \ulpmmax{\fmul}(z) \fmul y
    &= \bigl(|z| 2^{-(\emin+1-p)}\bigr) \fmul 2^{\emin+1-p} \\
\label{eq:thm:mu-fmul-satisfies-properties-1}
    &= \bigroundnear{|z| 2^{-(\emin+1-p)} 2^{\emin+1-p}} \\
\notag
    &= |z| \\
\notag
    &= z.
\end{align}
Eq.~\eqref{eq:thm:mu-fmul-satisfies-properties-1} holds because,
since $z$ is normal,
we have $z 2^{-(\emin + 1 - p)} \leq \fmax$.
In order to prove~\eqref{eq:ulpmmax2}, we have to show that,
for each $z' > \ulpmmax{\fmul}(z)$ there does not exist
$y \in \Fsetsub_{p,\emax}$ such that $z' \fmul y = z$.
By monotonicity of $\fmul$, a $y$ satisfying $z' \fmul y = z$
should be smaller than or equal to $\fmin $ and greater than $+0$.
However, the smallest float in $\Fsetsub_{p,\emax}$ that is greater than
$+0$ is $\fmin $.
Hence we are left to prove that
$\forall z' > \ulpmmax{\fmul}(z) \itc z' \fmul \fmin > z$.
Since $z' \geq \ulpmmax{\fmul}(z)^+$, we have two cases:
\begin{description}

\item[$\ulpmmax{\fmul}(z)^+ = +\infty:$]
In this case, $z' \fmul \fmin=+\infty>z$.

\item[$\ulpmmax{\fmul}(z)^+ \neq +\infty:$]
Letting $z = \float{m}{\expf{z}}$ we have
\begin{align}
\notag
  \ulpmmax{\fmul}(z)^+
    &= (\float{m}{\expf{z} - \emin - 1 + p})^+ \\
\notag
    &= (m + 2^{1-p})2^{\expf{z} - \emin - 1 + p}  \\
\notag
    &= m 2^{\expf{z} - \emin - 1 + p} + 2^{\expf{z} - \emin} \\
\notag
    &= \ulpmmax{\fmul}(z)+2^{\expf{z}-\emin}, \\
\intertext{%
hence
}
\notag
  z' \fmul \fmin
    &= \roundnear{z' \fmin} \\
\notag
    &\geq \bigroundnear{(\ulpmmax{\fmul}(z) + 2^{\expf{z}-\emin}) \fmin} \\
\notag
    &= \bigroundnear{(z \fmin^{-1} + 2^{\expf{z}-\emin}) \fmin} \\
\notag
    &= \roundnear{z + 2^{\expf{z}-\emin} \fmin} \\
\notag
    &= \roundnear{z + 2^{\expf{z}-\emin} 2^{\emin+1-p}} \\
\notag
    &= \roundnear{z + 2^{\expf{z}+1-p}} \\
\label{eq:thm:mu-fmul-satisfies-properties-2}
    &= z^+ \\
\notag
    &> z,
\end{align}
\end{description}
where~\eqref{eq:thm:mu-fmul-satisfies-properties-2}
holds because $z \geq \fnormin$.
In any case, \eqref{eq:ulpmmax2} holds.

\item[$0 < z <\fnormin$ and $\feven(z):$]
We have $\ulpmmax{\fmul}(z) = |z|2^{-(\emin + 1 - p)} + 2^{-1}$,
hence $y = \fmin = 2^{\emin + 1 - p}$ satisfies~\eqref{eq:ulpmmax1}:
\begin{align}
\notag
  \ulpmmax{\fmul}(z) \fmul \fmin
    &= \scaleroundnear{\bigl((z / \fmin) +2^{-1}\bigr) \fmin} \\
\notag
    &= \roundnear{z + 2^{-1} 2^{\emin+1-p}} \\
\notag
    &= \roundnear{z + 2^{\emin-p}} \\
\notag
    &= \roundnear{ z+\ferrdown{z}/2} \\
\label{eq:thm:mu-fmul-satisfies-properties-3}
    &= z.
\end{align}
Note that, as we have $\feven(z)$,
\eqref{eq:thm:mu-fmul-satisfies-properties-3}
holds by Definition~\ref{def:round-to-nearest-tail-to-even}

In order to prove~\eqref{eq:ulpmmax2}, we have to show that,
for each $z' > \ulpmmax{\fmul}(z)$, $z' \fmul \fmin > z$.
Of course, as observed in the previous case,
$y$ cannot be smaller than $\fmin$.
However, for each $z' \geq \bigl(\ulpmmax{\fmul}(z)\bigr)^+$,
we have
\begin{align}
\label{ineq:thm:mu-fmul-satisfies-properties-4}
  z' \fmul \fmin
    &\geq \bigl(\ulpmmax{\fmul}(z)\bigr)^+ \fmul \fmin \\
\label{ineq:thm:mu-fmul-satisfies-properties-5}
    &> \scaleroundnear{
         \bigl((z / \fmin) +2^{-1} + 2^{1-p}2^{\expf{z}-\emin-1+p} \bigr) \fmin
       } \\
\notag
    &= \roundnear{  z+   2^{\emin-p}+ 2^{1-p+\expf{z}} } \\
\notag
    &> \roundnear{ z+\ferrdown{z}/2} \\
\label{eq:thm:mu-fmul-satisfies-properties-6}
    &\geq z^+,
\end{align}
where~\eqref{ineq:thm:mu-fmul-satisfies-properties-4} holds by
monotonicity of $\fmul$,
\eqref{ineq:thm:mu-fmul-satisfies-properties-5}
holds because
\(
  \lexpf\bigl(\ulpmmax{\fmul}(z)\bigr)
    = \lexpf(z/\fmin +2^{-1})
    \geq \expf{z} - \emin - 1 + p
\),
and~\eqref{eq:thm:mu-fmul-satisfies-properties-6} holds
by Definition~\ref{def:round-to-nearest-tail-to-even}.

\item[$0 < z < \fnormin$ and $\fodd(z):$]
We have $\ulpmmax{\fmul}(z)= (|z| 2^{-(\emin+1-p)}+ 2^{-1})^-$
and we prove that~\eqref{eq:ulpmmax1}
is satisfied with $y = \fmin = 2^{\emin+1-p}$.
To this aim we show that
$\ulpmmax{\fmul}(z) \fmul \fmin = \bigroundnear{\ulpmmax{\fmul}(z) \fmin} = z$.
In order to prove the latter equality,
by Definition~\ref{def:round-to-nearest-tail-to-even},
we need to show that
$z - 2^{\emin-p} \leq \ulpmmax{\fmul}(z) \fmin \leq z + 2^{\emin-p}$.
In fact, on the one hand we have
\begin{align}
\label{eq:thm:mu-fmul-satisfies-properties-7}
  \ulpmmax{\fmul}(z) \fmin
    &\leq (z/\fmin + 2^{-1} - 2^{1-p}2^{\expf{z}-\emin-1+p} )\fmin \\
\notag
    &= z + 2^{-1} 2^{\emin+1-p} - 2^{1-p+\expf{z}} \\
\notag
    &= z + 2^{\emin-p} - 2^{1-p+\expf{z}} \\
\notag
    &< z + 2^{\emin-p},
\end{align}
where~\eqref{eq:thm:mu-fmul-satisfies-properties-7} holds because
\(
  \lexpf\bigl(\ulpmmax{\fmul}(z) +2^{-1}\bigr)
    \leq \lexpf(z\fmin)
    = \expf{z} - \emin - 1 + p
\).
On the other hand,
we can prove that $\ulpmmax{\fmul}(z) \fmin \geq z - 2^{\emin-p}$:
\begin{align}
\label{eq:thm:mu-fmul-satisfies-properties-8}
  \ulpmmax{\fmul}(z) \fmin
    &\geq (z/\fmin + 2^{-1} - 2^{1-p}2^{\expf{z} - \emin + p}) \fmin \\
\notag
    &= z + 2^{-1}2^{\emin+1-p} - 2^{-p + \expf{z}} \\
\notag
    &= z + 2^{\emin-p} -2^{-p+\expf{z}} \\
\label{eq:thm:mu-fmul-satisfies-properties-9}
    &> z -2^{\emin-p},
\end{align}
where~\eqref{eq:thm:mu-fmul-satisfies-properties-8} holds because
\(
  \lexpf(\ulpmmax{\fmul}(z) + 2^{-1})
    \geq \lexpf(z\fmin + 1)
    = \expf{z} - \emin + p
\)
and, since $z$ is subnormal,
\eqref{eq:thm:mu-fmul-satisfies-properties-9} holds because
$2^{-p+\expf{z}} < 2^{\emin-p}$.
By Definition~\ref{def:round-to-nearest-tail-to-even},
we can conclude that
$\ulpmmax{\fmul}(z) \fmul \fmin = \bigroundnear{\ulpmmax{\fmul}(z) \fmin} = z$,
as we have $\fodd(z)$.

In order to prove~\eqref{eq:ulpmmax2}, we have to show that,
for each $z' > \ulpmmax{\fmul}(z)$, $z' \fmul \fmin > z$.
Again, $y$ cannot be smaller than $\fmin$ and
for $z' \geq \bigl(\ulpmmax{\fmul}(z)\bigr)^+$ we have:
\begin{align}
\notag
  z' \fmul \fmin
    &\geq \bigl(\ulpmmax{\fmul}(z)\bigr)^+ \fmul \fmin \\
\notag
    &= \Bigl(\bigl(\roundnear{z/\fmin + 2^{-1}}\bigr)^-\Bigl)^+ \\
\notag
    &= \roundnear{z/\fmin + 2^{-1}} \\
\notag
    &= \roundnear{z + 2^{-1}2^{\emin+1-p}} \\
\notag
    &= \roundnear{  z+   2^{\emin-p}} \\
\notag
    &= \roundnear{z + \ferrdown{z}/2} \\
\label{eq:thm:mu-fmul-satisfies-properties-10}
    &= z^+.
\end{align}
Note that~\eqref{eq:thm:mu-fmul-satisfies-properties-10} holds
by Definition~\ref{def:round-to-nearest-tail-to-even},
since we have $\fodd(z)$.

\item[$-(2-2^{1-p})2^{\emax+\emin+1-p}\leq z < 0:$]
Choosing $y = -\fmin$ we can reason, depending on the value of $|z|$,
as in the previous cases.
\Halmos
\end{description}
\endproof

\setcounter{proposition}{3}
\begin{proposition}
Let $z \in \Fset_{\fmul}$ be nonzero.
If $z > 0$, then $\ulpmmax{\fmul}(z^+) \geq \ulpmmax{\fmul}(z)$;
on the other hand,
if $z < 0$, then $\ulpmmax{\fmul}(z^-) \geq \ulpmmax{\fmul}(z)$.
\end{proposition}
\proof{Proof.}
Assume $z > 0$, the other case being symmetric.
For $z \geq \fnormin$ the property holds by monotonicity of division
on the dividend.
The following cases remain:
\begin{description}

\item[$0 < z < (\fnormin)^-$ and $\feven(z):$]
We need to show that $\ulpmmax{\fmul}(z^+) \geq \ulpmmax{\fmul}(z)$.
Since $z$ is subnormal, by Definition~\ref{def:ulpmmax_fmul}
and the observation that all the floating-point operations that occur in it
are exact, we have
\begin{align}
\notag
  \ulpmmax{\fmul}(z^+)
    &= \bigl((z+2^{1-p+\emin})/\fmin+2^{-1}\bigr)^- \\
\label{ineq:prop:mu-fmul-monotonicity-1}
    &\geq (z+2^{1-p+\emin})/\fmin+2^{-1}-2^{\expf{z}-1+p-\emin} \\
\notag
    &= z/\fmin + 1 + 2^{-1} - 2^{\expf{z}-1+p-\emin} \\
\label{ineq:prop:mu-fmul-monotonicity-2}
    &\geq z/\fmin +2^{-1} \\
\notag
    &= \ulpmmax{\fmul}(z),
\end{align}
where~\eqref{ineq:prop:mu-fmul-monotonicity-1} holds
because
\(
  \lexpf\bigl(\ulpmmax{\fmul}(z) +2^{-1}\bigr)
    \geq \lexpf(z\fmin + 1)
    = \expf{z} - \emin + p
\),
whereas~\eqref{ineq:prop:mu-fmul-monotonicity-2} holds because
$2^{\expf{z} - 1 + p - \emin} \leq1 $.

\item [$0 < z < (\fnormin)^-$ and $\fodd(z):$]
In this case the result holds because
\begin{align*}
  \ulpmmax{\fmul}(z)
    &= (z/\fmin + 2^{-1})^- \\
    &< z/\fmin + 2^{-1} \\
    &< (z^+)/\fmin +2^{-1} \\
    &= \ulpmmax{\fmul}(z^+).
\end{align*}

\item [$z = (\fnormin)^-$:]
Note that in this case we have $\fodd(z)$, hence,
\begin{align*}
  \ulpmmax{\fmul}(z)
    &= (z/\fmin + 2^{-1})^- \\
    &< z/\fmin + 2^{-1} \\
    &< z/\fmin + 1 \\
    &= (z + 2^{1-p+\emin})/\fmin \\
    &= \ulpmmax{\fmul}(z^+) \\
    &= \ulpmmax{\fmul}(\fnormin).
   \end{align*}
\end{description}
\Halmos
\endproof

\begin{lemma}
\label{lem:mul-div-by-fmax}
If $z \in \Fset'_{\fdiv}$, then
$(z \fmul \fmax) \fdiv \fmax = z$.
\end{lemma}
\proof{Proof.\footnotemark}
\footnotetext{The main idea of this proof is due
to Paul Zimmermann, INRIA, France.}As $\roundnear{\cdot}$ is a symmetric rounding mode we can focus on
the cases where $+0 \leq z \leq 1$: the cases where $-1 \leq z \leq -0$ are
symmetric.
We thus consider the following cases:

\begin{description}
\item[$z = 1:$]
We have
$z \fmul \fmax = \roundnear{z \fmax} = \fmax$,
hence,
\begin{align*}
  (z \fmul \fmax) \fdiv \fmax
    &= \bigroundnear{(z \fmul \fmax ) / \fmax} \\
    &= \roundnear{\fmax / \fmax } \\
    &= 1 \\
    &= z.
\end{align*}

\item[$z = 1/2:$]
As
\(
  z \fmul \fmax
    = \roundnear{2^{-1} \fmax}
    = \bigroundnear{(2 - 2^{1-p}) 2^{\emax-1}}
    = (2 - 2^{1-p}) 2^{\emax-1}
\),
we have
\begin{align*}
  (z \fmul \fmax) \fdiv \fmax
    &= \bigroundnear{(z \fmul \fmax) / \fmax} \\
    &= \scaleroundnear{\frac{(2 - 2^{1-p}) 2^{\emax-1}}
                            {(2 - 2^{1-p}) 2^{\emax}}} \\
    &= 1/2 \\
    &= z.
\end{align*}

\item[$1/2 < z < 1:$]
In this case we have
\begin{align}
\label{eq:mul-div-by-fmax-first}
  z \fmul \fmax
    &= \roundnear{z \fmax} \\
    &= \bigroundnear{z (2 - 2^{1-p}) 2^{\emax}} \\
    &= \bigroundnear{z (1 - 2^{-p}) 2^{\emax+1}} \\
\label{eq:mul-div-by-fmax-mid}
    &= \bigroundnear{z (1 - 2^{-p})} 2^{\emax+1} \\
    &= \roundnear{z - z 2^{-p}} 2^{\emax+1} \\
\label{eq:mul-div-by-fmax-last}
    &= z^- \cdot 2^{\emax+1}.
\end{align}
Note that equality~\eqref{eq:mul-div-by-fmax-mid} holds because the
multiplication by $2^{\emax+1}$ can give rise neither to an
overflow, since $z \fmax < \fmax$,
nor to an underflow,
since $z (1 - 2^{-p}) > 2^{-1} (1 - 2^{-p}) \gg \fmin$.
To see why equality~\eqref{eq:mul-div-by-fmax-last} holds,
recall Definition~\ref{def:round-to-nearest-tail-to-even}
and consider that $\ferrup{z} = \ferrdown{z} = 2^{-p}$ for
$1/2 < z < 1$; we thus have
\(
  z^- - \ferrup{z^-}/2
    = (z - 2^{-p}) - 2^{-p-1}
    <  z - z 2^{-p}
    <  z - 2^{-p-1}
    = z^- + \ferrdown{z^-}/2
\).
Now we can write
\begin{align*}
  (z \fmul \fmax) / \fmax
    &= (z^- \cdot 2^{\emax+1}) / \fmax \\
    &= \frac{(z-2^{-p}) 2^{\emax+1}}{(1-2^{-p})2^{\emax+1}} \\
    &= (z-2^{-p})/(1-2^{-p}) \\
    &< z,
\end{align*}
and, since $z \geq 1/2+2^{-p}$, whence $1 - z \leq 1/2 - 2^{-p}$,
\begin{align*}
  z - \bigl((z \fmul \fmax) / \fmax)\bigr)
   &= z - \bigl((z-2^{-p}) / (1-2^{-p})\bigr)\\
   &= (z-z 2^{-p} -z+ 2^{-p}) / (1-2^{-p})\\
   &= \bigl(2^{-p}(1-z)\bigr) / (1-2^{-p})\\
   &\leq \bigl(2^{-p} (1/2 - 2^{-p})\bigr) / (1-2^{-p}) \\
   &= 2^{-p} \bigl((1/2 - 2^{-p}) / (1-2^{-p})\bigr) \\
   &< 2^{-p} \cdot 1/2 \\
   &= \ferrup{z}/2.
\end{align*}
As $0 < z - \bigl((z \fmul \fmax) / \fmax\bigr) < \ferrup{z}/2$,
we have
$z - \ferrup{z}/2 < (z \fmul \fmax) \fdiv \fmax < z$.
Hence, by Definition~\ref{def:round-to-nearest-tail-to-even},
we can conclude that
$\bigroundnear{(z \fmul \fmax) / \fmax} = z$.

\item[$\fnormin \leq z < 1/2:$]
In this case $z$ is such that $2^{-\ell} \leq z<2^{-\ell+1}$
with $-\emin \leq \ell \leq 2$,
and we can apply the same reasoning of the last two cases above
by substituting the exponent $-1$ with the exponent $-\ell$;
this is because $z \fmul \fmax$ does never generate
an overflow (a fortiori, since $z$ is now smaller)
nor an underflow,
because $z (1 - 2^{-p}) \ge 2^\emin (1 - 2^{-p}) > \fmin$.

\item[$2^{\emin-1} < z < \fnormin:$]
In this case we have
\begin{align}
\label{eq:mul-div-by-fmax-first-sub}
  z \fmul \fmax
    &= \roundnear{z \fmax} \\
    &= \bigroundnear{z (2 - 2^{1-p}) 2^{\emax}} \\
    &= \bigroundnear{(2z-z2^{1-p}) 2^{\emax}} \\
\label{eq:mul-div-by-fmax-last-but-one-sub}
    &= \bigroundnear{(z - z 2^{-p}) 2^{\emax+1}} \\
\label{eq:mul-div-by-fmax-last-sub}
    &= (z 2^{\emax+1})^-.
\end{align}
To see why~\eqref{eq:mul-div-by-fmax-last-sub} holds,
note that we can express $z$ as $\float{m}{\expf{z}}$
with $1 < m <2 $ and $\expf{z} = \emin - 1$. Then
$z 2^{\emax+1}=m 2^{\expf{z}+\emax+1}$.
Since $m > 1$,
\begin{align}
\notag
  \ferrup{z 2^{\emax+1}}
    &= z 2^{\emax+1} - (z 2^{\emax+1})^- \\
\notag
    &= m2^{\expf{z}+\emax+1} - (m-2^{1-p}) 2^{\expf{z}+\emax+1} \\
\label{eq:mul-div-by-fmax-sub-delta-1}
    &= 2^{1-p}2^{\expf{z}+\emax+1}.
\intertext{%
Similarly,
}
\notag
  \ferrdown{(z 2^{\emax+1})^-}
    &= \bigl((z 2^{\emax+1})^-\bigr)^+ - (z 2^{\emax+1})^- \\
\notag
    &= z 2^{\emax+1} - (z 2^{\emax+1})^- \\
\label{eq:mul-div-by-fmax-sub-delta-2}
    &= 2^{1-p}2^{ \expf{z}+\emax+1}.
\intertext{%
Finally, exploiting once again the fact that $m > 1$,
}
\notag
  \ferrup{(z 2^{\emax+1})^-}
    &= (z 2^{\emax+1})^- - \bigl((z 2^{\emax+1})^-\bigr)^- \\
\label{eq:mul-div-by-fmax-sub-delta-3}
    &\leq (m-2^{1-p}) 2^{\expf{z}+\emax+1} - (m-2^{2-p}) 2^{\expf{z}+\emax+1} \\
\label{eq:mul-div-by-fmax-sub-delta-4}
    &= 2^{1-p}2^{ \expf{z}+\emax+1}.
 \end{align}
For~\eqref{eq:mul-div-by-fmax-sub-delta-3},
note that  $m > 1$
implies that $(z 2^{\emax+1})^- = (m-2^{1-p}) 2^{\expf{z}+\emax+1}$.
Applying the same reasoning to
$\bigl((z 2^{\emax+1})^-\bigr)^- = \bigl((m-2^{1-p}) 2^{\expf{z}+\emax+1}\bigr)^-$,
we have two cases:
\begin{description}

\item[$(m-2^{1-p})>1:$]
then, as before, we have $\ferrup{(z 2^{\emax+1})^-}= 2^{1-p} 2^{\expf{z}+\emax+1}$
and thus
\begin{align*}
  \bigl((z 2^{\emax+1})^-\bigr)^-
    &= (m-2^{1-p}) 2^{\expf{z}+\emax+1} - 2^{1-p} 2^{\expf{z}+\emax+1} \\
    &= (m-2^{2-p}) 2^{\expf{z}+\emax+1};
\end{align*}
as a consequence, \eqref{eq:mul-div-by-fmax-sub-delta-3}
holds with the equality;

\item[$(m-2^{1-p})=1:$]
in this case $\ferrup{(z 2^{\emax+1})^-} = 2^{1-p} 2^{\expf{z}+\emax}$,
hence
\begin{align*}
  \bigl((z 2^{\emax+1})^-\bigr)^-
    &= (m-2^{1-p}) 2^{\expf{z}+\emax+1} - 2^{1-p} 2^{\expf{z}+\emax} \\
    &= (m-2^{1-p}-2^{-p}) 2^{\expf{z}+\emax+1};
\end{align*}
as a consequence, \eqref{eq:mul-div-by-fmax-sub-delta-3} holds
with the inequality.
\end{description}

In order to prove~\eqref{eq:mul-div-by-fmax-last-sub},
by Definition~\ref{def:round-to-nearest-tail-to-even},
we have to show
\begin{align}
\label{eq:mul-div-by-fmax-lb}
   (z 2^{\emax+1})^- - \frac{\ferrup{(z 2^{\emax+1})^-}}{2}
     &< (z - z 2^{-p}) 2^{\emax + 1} \\
\label{eq:mul-div-by-fmax-ub}
     &< (z 2^{\emax+1})^- + \frac{\ferrdown{(z 2^{\emax+1})^-}}{2}.
 \end{align}
To prove~\eqref{eq:mul-div-by-fmax-lb} observe that,
by~\eqref{eq:mul-div-by-fmax-sub-delta-1},
\begin{equation}
\label{eq:mul-div-by-fmax-sub-delta-5}
  (z 2^{\emax+1})^- = z 2^{\emax+1}-\ferrup{z 2^{\emax+1}}
    = z2^{\emax+1} - 2^{1-p+\expf{z}+\emax+1}.
\end{equation}
Hence, by~\eqref{eq:mul-div-by-fmax-sub-delta-4}, we have
\begin{align}
\notag
  (z 2^{\emax+1})^- - \frac{\ferrup{(z 2^{\emax+1})^-}}{2}
    &\leq(z 2^{\emax+1}) - 2^{1-p+\expf{z}+\emax+1} - 2^{-p+\expf{z}+\emax+1} \\
\notag
    &< (z - 2^{1-p+\expf{z}}) 2^{\emax+1} \\
\label{eq:mul-div-by-fmax-2}
    &< (z - m 2^{-p+\expf{z}}) 2^{\emax+1} \\
\notag
    &= (z - z 2^{-p}) 2^{\emax+1},
\end{align}
where~\eqref{eq:mul-div-by-fmax-2} holds because $1 < m < 2$.
We are left to prove~\eqref{eq:mul-div-by-fmax-ub}.
To this aim, we write the following sequence of inequalities,
which are all equivalent:
\begin{align}
\label{eq:mul-div-by-fmax-sub-delta-6}
  (z - z 2^{-p}) 2^{\emax+1}
    &<
      (z 2^{\emax+1})^- + \ferrdown{(z 2^{\emax+1})^-}/2 \\
\label{eq:mul-div-by-fmax-sub-delta-7}
  (z - z 2^{-p}) 2^{\emax+1}
    &<
      (z2^{\emax+1}-2^{1-p+\expf{z}+\emax+1})+2^{-p+\expf{z}+\emax+1} \\
  z - z 2^{-p}
    &<
      (z-2^{1-p+\expf{z}})+2^{-p+\expf{z}} \notag\\
  -z 2^{-p}
    &<
      -2^{-p+\expf{z}} \notag\\
  2^{-p+\expf{z}}
    &<
      z 2^{-p} \notag\\
  2^{-p+\expf{z}}
    &<
      (m 2^{\expf{z}}) 2^{-p} \notag \\
   1
    &<
      m \notag
\end{align}
where~\eqref{eq:mul-div-by-fmax-sub-delta-6} is equivalent to~\eqref{eq:mul-div-by-fmax-sub-delta-7}
because of~\eqref{eq:mul-div-by-fmax-sub-delta-5} and~\eqref{eq:mul-div-by-fmax-sub-delta-2}.
Moreover, since we have decomposed $z$ so that $1 < m < 2$, the last inequality holds
and we can conclude that $z \fmul \fmax = (z 2^{\emax+1})^-$.
Now we can write
\begin{align*}
  (z \fmul \fmax) / \fmax
    &= (z 2^{\emax+1})^- / \fmax \\
    &= \frac{(z-2^{1-p+\expf{z}}) 2^{\emax+1}}{(1-2^{-p})2^{\emax+1}} \\
    &= \frac{z-2^{1-p+\expf{z}}}{1-2^{-p}}.
\end{align*}
As in the previous case, we want to show that
$ z - (z \fmul \fmax) / \fmax < \ferrup{z}/2$,
since this will guarantee that
$(z \fmul \fmax)\fdiv \fmax = z$.
In fact,
\begin{align}
  z - (z \fmul \fmax) / \fmax
\notag
    &= \frac{z - (z-2^{1-p+\expf{z}})}{1-2^{-p}} \\
\notag
    &= \frac{z-z 2^{-p} -z+ 2^{1-p+\expf{z}}}{1-2^{-p}} \\
\notag
    &= \frac{-z 2^{-p} + 2^{1-p+\expf{z}}}{1-2^{-p}} \\
\label{eq:sub1}
    &= \frac{2^{\emin-p} - z2^{-p}}{1-2^{-p}} \\
\label{eq:sub2}
    &< \frac{2^{\emin-p} - 2^{\emin-p-1}}{1-2^{-p}} \\
\notag
    &= \frac{2^{\emin-p-1}}{1-2^{-p}} \\
\notag
    &< 2^{\emin-p} \\
\label{eq:sub3}
    &= \ferrup{z}/2,
\end{align}
where Eq.~\eqref{eq:sub1} holds as $\expf{z} = \emin-1$;
moreover, \eqref{eq:sub2} holds as $2^{\emin-1} < z < \fnormin$;
and~\eqref{eq:sub3} holds because, since $z$ is subnormal,
$\ferrup{z} = \fmin$.
From $0 < z - (z \fmul \fmax) / \fmax < \ferrup{z}/2$
we get $z - \ferrup{z}/2 < (z \fmul \fmax) / \fmax < z$.
Thus, by Definition~\ref{def:round-to-nearest-tail-to-even},
we can conclude
$(z \fmul \fmax)\fdiv \fmax = \bigroundnear{(z \fmul \fmax) / \fmax} = z$.

\item[$z = 2^{\emin-1}:$]
We have
\begin{align*}
  z \fmul \fmax
    &= \bigroundnear{2^{\emin-1} 2^{\emax}(2 - 2^{1-p})} \\
    &= \bigroundnear{(2 - 2^{1-p}) 2^{\emax+\emin-1}} \\
    &= (2 - 2^{1-p}) 2^{\emax+\emin-1},
\end{align*}
hence
\begin{align*}
  \bigroundnear{(z \fmul \fmax) / \fmax}
    &= \scaleroundnear{\frac{(2 - 2^{1-p}) 2^{\emax+\emin-1}}
                            {(2 - 2^{1-p}) 2^{\emax}}} \\
    &= 2^{\emin-1} \\
    &= z.
\end{align*}

\item[$\fmin \leq z < 2^{\emin-1}:$]
In this case $z$ is such that $2^{-\ell} \leq z <2^{-\ell+1}$
provided that $-(\emin - p + 1) \leq \ell \leq -\emin + 2$, hence,
we can apply the same reasoning of the last two cases above
by substituting the exponent $\emin - 1$ with $\ell$.

\item[$z = 0:$]
Note that,
for $z = +0$, we have
$(z \fmul \fmax) \fdiv \fmax = +0 \fdiv \fmax = +0$ while,
for $z = -0$,
we have $(z \fmul \fmax) \fdiv \fmax = -0 \fdiv \fmax = -0$.
\end{description}
\Halmos
\endproof

\begin{lemma}
\label{lem:mu-fdiv-satisfies-properties}
The restriction of $\ulpmmax{\fdiv}$ to $\Fset'_{\fdiv} \cap \Fset_{p,\emax}$
is well-defined and satisfies~\eqref{eq:ulpmmax1} and~\eqref{eq:ulpmmax2}.
\end{lemma}
\proof{Proof.}
Note that the range of $\ulpmmax{\fdiv}$ is constituted
by non negative elements of $\Fset_{p,\emax}$.

Consider first the case where $z > 0$.
By definition, $\ulpmmax{\fdiv}(z) = z \fmul \fmax $;
hence, choosing $y = \fmax$ and applying Lemma~\ref{lem:mul-div-by-fmax},
we get
\(
  \ulpmmax{\fdiv}(z) \fdiv y
    = (z \fmul \fmax) \fdiv \fmax
    = z
\),
so that~\eqref{eq:ulpmmax1} holds.
In order to prove~\eqref{eq:ulpmmax2}, we have to show that,
for each $z' \in \Fsetsub_{p,\emax}$ with $z' > \ulpmmax{\fdiv}(z)$,
there is no $y \in \Fsetsub_{p,\emax}$ such that $z' \fdiv y = z$.
We first prove that $z' \fdiv \fmax > z$.
Let $\hat{z}$ be the smallest floating-point number strictly greater than
$\ulpmmax{\fdiv}(z)= z \fmul \fmax$,
i.e., $\hat{z} = z \fmul \fmax  + 2^{1-p+\lexpf(z \fmul \fmax)}$.
We have two cases:
\begin{description}

\item[$\lexpf(z \fmul \fmax) = \expf{z} + \emax + 1:$]
Then
\begin{align*}
  \hat{z} / \fmax
    &= \frac{(z \fmul \fmax) + 2^{1-p+ \expf{z}+\emax +1}}{\fmax} \\
\intertext{%
and, following the
steps~\eqref{eq:mul-div-by-fmax-first}--\eqref{eq:mul-div-by-fmax-last}
of the proof of Lemma~\ref{lem:mul-div-by-fmax}, we obtain
}
  \hat{z} / \fmax
    &= \frac{(z \fmul \fmax) + 2^{1-p} 2^{\expf{z}+\emax +1}}{\fmax} \\
    &= \frac{(z - 2^{1-p+\expf{z}}) 2^{\emax+1} + 2^{2-p+\expf{z}+\emax}}{\fmax} \\
    &= \frac{(2z - 2^{2-p+\expf{z}}) 2^{\emax} + 2^{2-p+\expf{z}+\emax}}{(2 - 2^{1-p})2^{\emax}} \\
    &= \frac{2z }{2 - 2^{1-p}} \\
    &= \frac{z}{1 - 2^{-p}}.
  \end{align*}
We now want to show that $\hat{z} \fdiv \fmax = \roundnear{\hat{z} / \fmax} \geq z^+$.
Hence, by Definition~\ref{def:round-to-nearest-tail-to-even},
we need to prove that
\(
  z / (1 - 2^{-p})
    > z + \ferrdown{z}/2
    = z^+ - \ferrup{z^+}
    = z + 2^{-p+\expf{z}}
\).
To this aim we write the following sequence of inequalities, which are all equivalent:
\begin{align*}
  \frac{z}{1 - 2^{-p}}
    &>
      z + 2^{-p+\expf{z}} \\
  z
    &>
      z + 2^{-p+\expf{z}} - z 2^{-p} - 2^{-2p+\expf{z}} \\
  0
    &>
      2^{-p+\expf{z}} - z 2^{-p} - 2^{-2p+\expf{z}} \\
  0
    &>
      2^{-p+\expf{z}} - m 2^{-p+\expf{z}} - 2^{-2p+\expf{z}} \\
  0
    &>
      (1 - m)2^{-p+\expf{z}} - 2^{-2p+\expf{z}}.
   \end{align*}
Since $z \in \Fset_{p,\emax}$, $z = \float{m}{\expf{z}}$
with $1 \leq m <2$.
Hence, the last inequality holds
and, by Definition~\ref{def:round-to-nearest-tail-to-even},
roundTiesToEven gives
\(
  \roundnear{\hat{z} / \fmax} = \hat{z} \fdiv \fmax \geq z^+ > z
\).

\item[$\lexpf(z \fmul \fmax) = \expf{z} + \emax:$]
This implies that $z = \float{1.0\ldots0}{-\ell}$
for some $\ell$ such that $-\emin \leq \ell \leq 0$.
In fact, $z \geq \fnormin$ as $z \in \Fset'_{\fdiv} \cap \Fset_{p,\emax}$.
We thus have that $z \fmul \fmax = (2 - 2^{1-p}) 2^{\emax-\ell}$ and
\begin{align*}
  \hat{z} / \fmax
    &= \frac{(z \fmul \fmax) + 2^{1-p-\ell+\emax}}{\fmax} \\
    &= \frac{(2 - 2^{1-p}) 2^{\emax-\ell} +2^{1-p- \ell+\emax}}{\fmax} \\
    &= \frac{2^{1+\emax-\ell}}{(2 - 2^{1-p})2^{\emax}} \\
    &= \frac{2^{1-\ell}}{2 - 2^{1-p}} \\
    &=\frac{2^{-\ell}}{1 - 2^{-p}}.
\end{align*}
As in the previous case, we want to show that
$\hat{z} \fdiv \fmax = \roundnear{\hat{z} / \fmax} \geq z^+$.
Hence, by Definition~\ref{def:round-to-nearest-tail-to-even},
we need  to prove that
\(
  2^{-\ell}/(1 - 2^{-p})
    > z + \ferrdown{z}/2
    = z^+ - \ferrup{z^+}
    = 2^{-\ell}+2^{-\ell-p}
\).
To this aim we write the following sequence of inequalities,
which are all equivalent:
\begin{align*}
  \frac{2^{-\ell}}{1 - 2^{-p}} &> 2^{-\ell}+2^{-\ell-p} \\
                    2^{-\ell} &> 2^{-\ell}+2^{-\ell-p}-2^{-\ell-p}-2^{-2p-\ell} \\
                          0  &> -2^{-2p-\ell}.
\end{align*}
Since the last inequality holds, we can conclude
that round-to-nearest gives
\(
  \roundnear{\hat{z} / \fmax} = \hat{z} \fdiv \fmax \geq z^+ > z
\).
\end{description}
In both cases an $y \in \Fset^+_{p,\emax}$ satisfying $z' \fdiv y = z$
should be greater than $\fmax$ and less than $+\infty$:
as such $y$ does not exist, \eqref{eq:ulpmmax2} holds.

For the case where $z < 0$ we can reason as before choosing $y = -\fmax$.
\Halmos
\endproof

\begin{lemma}
\label{lem:mu-fdiv-satisfies-properties-sub}
The restriction of $\ulpmmax{\fdiv}$ to $\Fset'_{\fdiv} \setminus \Fset_{p,\emax}$
is well-defined and satisfies~\eqref{eq:ulpmmax1} and~\eqref{eq:ulpmmax2}.
\end{lemma}
\proof{Proof.}
As already observed, the range of $\ulpmmax{\fdiv}$ is constituted
by non negative elements of $\Fset_{p,\emax}$.

Consider first the case where $z > 0$.
Choosing $y = \fmax$ and applying Lemma~\ref{lem:mul-div-by-fmax},
we obtain
\(
  (z \fmul \fmax) \fdiv y
    = (z \fmul \fmax) \fdiv \fmax
    = z
\),
but this is not enough.
In order to prove that~\eqref{eq:ulpmmax1} holds, we have to show that
$\ulpmmax{\fdiv}(z) \fdiv \fmax = z$.
We first show that
\begin{equation}
\label{eq:mu-fdiv-satisfies-properties-sub00}
  z \fmul \fmax = (z 2^{\emax+1})^-.
\end{equation}
We have two cases on the value of $z$:
\begin{description}

\item[$z = \float{1}{\expf{z}}$
      with $\emin - p + 1 \leq \expf{z} \leq \emin-1:$]
In this case
\begin{align*}
  z \fmul \fmax
    &= \bigroundnear{(2-2^{1-p}) 2^{\expf{z}+\emax}} \\
    &= \roundnear{\float{1}{\expf{z}+\emax+1} - 2^{1-p+\expf{z}+\emax} } \\
    &= \roundnear{z 2^{\emax+1} - \ferrup{z 2^{\emax+1}}} \\
    &= (z 2^{\emax+1})^-.
\end{align*}

\item[$z = \float{m}{\expf{z}}$ with $m > 1:$]
Following exactly the same
steps~\eqref{eq:mul-div-by-fmax-first-sub}--\eqref{eq:mul-div-by-fmax-last-but-one-sub}
of the proof of Lemma~\ref{lem:mul-div-by-fmax}, we obtain
$z \fmul \fmax = (z 2^{\emax+1})^-$.
\end{description}
In order to prove $\ulpmmax{\fdiv}(z) \fdiv \fmax = z$,
observe that
$(z \fmul \fmax) \fdiv \fmax \leq \ulpmmax{\fdiv}(z) \fdiv \fmax$,
since $\fdiv$ is monotonically non-decreasing in its first argument.
By Lemma~\ref{lem:mul-div-by-fmax}, we have $(z \fmul \fmax) \fdiv \fmax = z$,
therefore $z \leq \ulpmmax{\fdiv}(z) \fdiv \fmax$.
Hence, by Definition~\ref{def:round-to-nearest-tail-to-even},
we are left to prove $\ulpmmax{\fdiv}(z) / \fmax < z + \ferrdown{z}/2$.
We now distinguish three cases on $z$:
\begin{description}

\item[$z \neq \float{1}{\expf{z}}:$]
Recall that $q = 1 - p + \emin + \emax$.
We begin by proving that we have
\(
  \ulpmmax{\fdiv}(z)
    = \bigl(z \fmul \fmax\bigr) \fadd 2^q
    = \bigl(z \fmul \fmax\bigr) + 2^q
    = (z 2^{\emax+1})^- + 2^q
\).
Let $z= m 2^{\expf{z}}$, for some $m$ with $1 \leq m <2$.
It is worth to observe that, for $z = m 2^{\expf{z}}$,
\begin{equation}
\label{denormalized-significand2}
  m < 2 - 2^{\emin-\expf{z}} 2^{1-p},
\end{equation}
since the normalized significand $m$ was obtained from a denormalized significand
$m' = 0.0 \cdots 0b_{\emin-\expf{z}+1} \cdots b_p$
with $b_{\emin-\expf{z}+1} = 1$.
Then we can write
\begin{align}
\notag
  (z \fmul \fmax) \fadd 2^q
    &= \bigroundnear{\bigl(z \fmul \fmax\bigr) + 2^q} \\
\label{eq:fadd-2powq-is-exact-1}
    &= \bigroundnear{(z 2^{\emax+1})^- + 2^q} \\
\notag
    &= \bigroundnear{(m 2^{\expf{z}} 2^{\emax+1})^- + 2^q} \\
\notag
    &= \bigroundnear{(m-2^{1-p}) 2^{\emax+1+ \expf{z}} + 2^q} \\
\notag
    &= \scaleroundnear{\bigl((m-2^{1-p}) + 2^{1-p} 2^{\emin -\expf{z}-1}\bigr)
                         2^{\emax+1+ \expf{z}}} \\
\label{eq:fadd-2powq-is-exact-2}
   &= \bigl((m-2^{1-p}) + 2^{1-p} 2^{\emin -\expf{z}-1}\bigr) 2^{\emax+1+ \expf{z}} \\
\notag
   &= (m 2^{\expf{z}} 2^{\emax+1})^-+2^q \\
\label{eq:fadd-2powq-is-exact-3}
    &= (z 2^{\emax+1})^-+2^q \\
\notag
   &=\bigl(z \fmul \fmax\bigr) + 2^q,
\end{align}
where~\eqref{eq:fadd-2powq-is-exact-1} holds
because of~Eq.\eqref{eq:mu-fdiv-satisfies-properties-sub00}.
For~\eqref{eq:fadd-2powq-is-exact-2} observe that, by~\eqref{denormalized-significand2},
we have $(m-2^{1-p}) + 2^{1-p} 2^{\emin -\expf{z}-1} < 2-2^{1-p}$,
hence the left-hand side of the latter inequality can be expressed by
a normalized significand without resorting to a greater exponent.

Now in order to prove that~\eqref{eq:ulpmmax1} holds, note that
the following inequalities are all equivalent:
\begin{align}
\label{eq:mu-fdiv-satisfies-properties-sub000pre}
  \frac{(z \fmul \fmax) \fadd 2^q}{\fmax}
    &< z + \frac{\ferrdown{z}}{2} \\
\label{eq:mu-fdiv-satisfies-properties-sub000}
  \frac{ (z 2^{\emax+1})^-+2^q}{\fmax}
    &< z + \frac{\ferrdown{z}}{2} \\
\label{eq:mu-fdiv-satisfies-properties-sub01}
  \frac{z2^{\emax+1}- 2^{1-p+\expf{z}+\emax+1} + 2^q}{(1-2^{-p})2^{\emax+1}}
    &< z+2^{\emin-p} \\
\notag
  \frac{z- 2^{1-p+\expf{z}} + 2^{-p+\emin}}{1-2^{-p}}
    &< z+2^{\emin-p} \\
\notag
  z - 2^{1-p+\expf{z}} + 2^{-p+\emin}
    &< (z+2^{\emin-p})(1-2^{-p}) \\
\notag
  z - 2^{1-p+\expf{z}} + 2^{-p+\emin}
    &<
      z+2^{\emin-p}-z2^{-p}-2^{\emin-2p} \\
\notag
  -2^{1-p+\expf{z}}
    &<
      -z2^{-p}-2^{\emin-2p} \\
\notag
  2^{\emin-p}
    &<
      2^{1+\expf{z}} -z \\
\notag
  2^{\emin-p}
    &<
      (2-m) 2^{\expf{z}} \\
\notag
  2^{\emin-p}
    &<
      \Bigl(2 - \bigl(2 - (2^{\emin-\expf{z}} 2^{1-p})\bigr)\Bigr) 2^{\expf{z}} \\
\notag
  2^{\emin-p}
    &<
      (2^{\emin-\expf{z}} 2^{1-p}) 2^{\expf{z}} \\
\notag
  2^{\emin-p}
    &<
      2^{\emin+1-p},
  \end{align}
where~\eqref{eq:mu-fdiv-satisfies-properties-sub000pre}
is equivalent to~\eqref{eq:mu-fdiv-satisfies-properties-sub000}
because of~\eqref{eq:fadd-2powq-is-exact-3} and because
$\ferrdown{z} = \fmin$, since $z$ is subnormal.
Moreover, \eqref{eq:mu-fdiv-satisfies-properties-sub000} is equivalent
to~\eqref{eq:mu-fdiv-satisfies-properties-sub01},
since $\ferrup{(z \fmul \fmax) \fadd 2^q} = 2^{1-p+\expf{z}+\emax+1}$.

In order to prove~\eqref{eq:ulpmmax2}
we need to prove that
\(
  \ulpmmax{\fdiv}(z)^+ \fdiv \fmax
    = \bigroundnear{\ulpmmax{\fdiv}(z)^+ / \fmax}
    > z
\).
By Definition~\ref{def:round-to-nearest-tail-to-even} it suffices to prove
that $\ulpmmax{\fdiv}(z)^+ / \fmax > z + \ferrdown{z}/2$.
We have that
\begin{align}
\notag
  \frac{\ulpmmax{\fdiv}(z)^+ }{\fmax}
    &= \frac{(z\fmul \fmax) + 2^q + 2^{1-p+\lexpf(z\fmul\fmax)}}
            {\fmax} \\
\label{eq:mu-fdiv-satisfies-properties-sub02}
    &= \frac{z2^{\emax+1} - 2^{1-p+\lexpf(z\fmul\fmax)} + 2^{1-p+\lexpf(z\fmul\fmax)} + 2^q}
            {\fmax} \\
\notag
    &= \frac{z + 2^{-p+\emin}}{1-2^{-p}} \\
\notag
    &> z + 2^{\emin-p} \\
\notag
    &= z + \ferrdown{z}/2,
\end{align}
where~\eqref{eq:mu-fdiv-satisfies-properties-sub02}
holds because of~\eqref{eq:mu-fdiv-satisfies-properties-sub00}.
Hence~\eqref{eq:ulpmmax2} is proved.

\item[$z = \float{1}{\emin-1}:$]
We first prove that, in this case, we have
\begin{equation}
\label{eq:mu-fdiv-satisfies-properties-sub03}
  \ulpmmax{\fdiv}(z) = \bigl(z \fmul \fmax\bigr) \fadd 2^q = z 2^{\emax+1}.
\end{equation}
By~\eqref{eq:mu-fdiv-satisfies-properties-sub00} we have that
\begin{align}
\notag
  \bigl(z \fmul \fmax\bigr) \fadd 2^q
    &= (z 2^{\emax+1})^- \fadd 2^q \\
\notag
    &= (2-2^{1-p})2^{\emax+1+\emin-2}\fadd 2^q \\
\notag
    &= \bigroundnear{(2-2^{1-p})2^{\emax+\emin-1}+ 2^q} \\
\notag
    &= \bigroundnear{(2-2^{1-p})2^{\emax+\emin-1} + 2^{1-p} 2^{\emin+\emax}} \\
\notag
    &= \scaleroundnear{\bigl((2-2^{1-p})+ 2^{1-p} + 2^{1-p}\bigr) 2^{\emin+\emax-1}} \\
\notag
    &= \bigroundnear{(1 + 2^{-p}) 2^{\emin+\emax}  } \\
\label{eq:mu-fdiv-satisfies-properties-sub04}
    &= \bigroundnear{2^{\emin+\emax} + \ferrdown{2^{\emin+\emax}}/2}\\
\notag
    &= 2^{\emin+\emax} \\
\notag
    &= z 2^{\emax+1},
\end{align}
where~\eqref{eq:mu-fdiv-satisfies-properties-sub04} holds
by Definition~\ref{def:round-to-nearest-tail-to-even}
as we have $\feven(z)$, and so is $z 2^{\emax+1} = 2^{\emin+\emax}$.

Then, in order to prove~\eqref{eq:ulpmmax1}, note that
the following inequalities are all equivalent:
\begin{align}
  \label{eq:mu-fdiv-satisfies-properties-sub05}
        \frac{\ulpmmax{\fdiv}(z) }{\fmax} &< z + \frac{\ferrdown{z}}{2} \\
\label{eq:mu-fdiv-satisfies-properties-sub06}
  \frac{ z 2^{\emax+1}}{(1-2^{-p})2^{\emax+1}} &< z+2^{\emin-p} \\
\notag
                        \frac{z}{1-2^{-p}} &< z+2^{\emin-p} \\
\notag
                                        z &< (z+2^{\emin-p})(1-2^{-p}) \\
\notag
                                        z &< z+2^{\emin-p}-z2^{-p}-2^{\emin-2p} \\
\notag
                                        0 &< 2^{\emin-p}-z2^{-p}-2^{\emin-2p} \\
\notag
                                        0 &< 2^{\emin-p}-2^{\emin-p-1}-2^{\emin-2p} \\
\notag
                                        0 &<2^{\emin-p-1}-2^{\emin-2p} \\
\notag
                                        0 &< 2^{-1}-2^{-p},
\end{align}
where~\eqref{eq:mu-fdiv-satisfies-properties-sub05} is equivalent to~\eqref{eq:mu-fdiv-satisfies-properties-sub06}
because of~\eqref{eq:mu-fdiv-satisfies-properties-sub03}.
Moreover, assuming $p>1$, the last inequality holds.

In order to prove~\eqref{eq:ulpmmax2} , we
need to prove that $\ulpmmax{\fdiv}(z)^+ \fdiv\fmax > z$.
By Definition~\ref{def:round-to-nearest-tail-to-even}
it suffices to prove that
$\ulpmmax{\fdiv}(z)^+ / \fmax > z + \ferrdown{z}/2$.
Indeed,
\begin{align}
\label{eq:mu-fdiv-satisfies-properties-sub07}
  \frac{\ulpmmax{\fdiv}(z)^+ }{\fmax}
    &= \frac{z 2^{\emax+1} + 2^q }{\fmax} \\
\notag
    &= \frac{z 2^{\emax+1} + 2^{1-p+\emin+\emax}}{2^{\emax}(2 - 2^{1-p})} \\
\notag
    &= \frac{z + 2^{\emin-p}}{1-2^{-p}} \\
\notag
    &> z + 2^{\emin-p} \\
\notag
    &= z + \ferrdown{z}/2,
\end{align}
where~\eqref{eq:mu-fdiv-satisfies-properties-sub07}
holds because of~\eqref{eq:mu-fdiv-satisfies-properties-sub03}.
Hence
\(
  \ulpmmax{\fdiv}(z)^+\fdiv \fmax
    = \bigroundnear{\ulpmmax{\fdiv}(z)^+/ \fmax}
    \geq z^+
\),
which proves~\eqref{eq:ulpmmax2}.

\item[$z = \float{1}{\expf{z}}$ with $\expf{z} < \emin - 1:$]
We first prove that, in this case,
\begin{equation}
\label{eq:mu-fdiv-satisfies-properties-sub10}
  \bigl(z \fmul \fmax\bigr) \fadd 2^q = z 2^{\emax+1} + 2^q.
\end{equation}
Applying~\eqref{eq:mu-fdiv-satisfies-properties-sub00}
we have that
\begin{align}
\notag
  \bigl(z \fmul \fmax\bigr) \fadd 2^q
    &= (z 2^{\emax+1})^- \fadd 2^q \\
\notag
    &= (2-2^{1-p}) 2^{\emax + 1 + \expf{z} - 1} \fadd 2^q \\
\notag
    &= \bigroundnear{(2-2^{1-p}) 2^{\emax + \expf{z}} + 2^q} \\
\notag
    &= \bigroundnear{(2-2^{1-p})2^{\emax + \expf{z}} + 2^{1-p} 2^{\emin-\expf{z}} 2^{ \emax+\expf{z}}}\\
\notag
    &= \scaleroundnear{
         \bigl(
           (2-2^{1-p}) + 2^{1-p} + 2^{1-p}
             + (2^{\emin - \expf{z}} - 2) 2^{1-p}
         \bigr) 2^{ \emax + \expf{z}}
       } \\
\notag
    &= \bigroundnear{
         (1 + 2^{-p} + (2^{\emin - \expf{z} - 1} - 1) 2^{1-p}) 2^{\emax + \expf{z} + 1}
       } \\
\notag
    &= \bigroundnear{
         (1 +(2^{\emin - \expf{z} - 1} - 1) 2^{1-p}) 2^{\emax + \expf{z} + 1}
           + 2^{-p}2^{\emax + \expf{z} + 1}
       } \\
\label{eq:mu-fdiv-satisfies-properties-sub11}
    &= (1 + 2^{1-p} + (2^{\emin - \expf{z} - 1} - 1)2^{1-p}) 2^{\emax + \expf{z} + 1} \\
\notag
    &= z 2^{\emax + 1} + 2^{\emin - \expf{z} - 1} 2^{1-p} 2^{\emax + \expf{z} + 1} \\
\notag
    &= z 2^{\emax + 1} + 2^{\emin+1-p+\emax}.
\end{align}
In order to appreciate why~\eqref{eq:mu-fdiv-satisfies-properties-sub11}
holds,
note first that, as $\expf{z} \geq \emin - p + 1$,
we have $1 + (2^{\emin - \expf{z} - 1} - 1) 2^{1-p} < 1+2^{-1}$.
This ensures that the floating-point number
$\bigl(1 + (2^{\emin - \expf{z} - 1} - 1) 2^{1-p}\bigr) 2^{\emax + \expf{z} + 1}$
is represented by a normalized significand of the form $1.0 b_2 \cdots b_p$.
Moreover, observe that $1 + (2^{\emin - \expf{z} - 1} - 1) 2^{1-p}$
--- and, consequently,
$\bigl(1 + (2^{\emin - \expf{z} - 1} - 1) 2^{1-p}\bigr) 2^{\emax + \expf{z} + 1} $ ---
is necessarily represented by an odd significand,
since the number that multiplies $2^{1-p}$ is odd.
Finally, note that
\[
  \frac{\ferrdown{(1 + (2^{\emin-\expf{z}-1} - 1) 2^{1-p}) 2^{ \emax+\expf{z}+1}}}{2}
    = 2^{-p} 2^{\emax+\expf{z}+1}
\]
and thus, by Definition~\ref{def:round-to-nearest-tail-to-even},
since
\(
  \fodd\bigl(
    \bigl(1 + (2^{\emin-\expf{z}-1} - 1) 2^{1-p}\bigr) 2^{ \emax+\expf{z}+1}
  \bigr)
\),
we can conclude that~\eqref{eq:mu-fdiv-satisfies-properties-sub11} holds.

Consider now the following sequence of equivalent inequalities:
\begin{align}
\label{eq:mu-fdiv-satisfies-properties-sub08}
  \frac{\ulpmmax{\fdiv}(z)}{\fmax}
    &< z + \frac{\ferrdown{z}}{2} \\
  \frac{(z 2^{\emax+1} + 2^{\emin+1-p+\emax})^-}{(1-2^{-p})2^{\emax+1}}
\label{eq:mu-fdiv-satisfies-properties-sub09}
    &< z+2^{\emin-p} \\
\label{eq:mu-fdiv-satisfies-properties-sub12}
  \frac{z 2^{\emax+1} + 2^{\emin+1-p+\emax} - 2^{1-p+\emax+\expf{z}+1}}
       {(1-2^{-p})2^{\emax+1}}
    &< z+2^{\emin-p} \\
\notag
  \frac{z + 2^{\emin-p} -2^{ 1-p+\expf{z}}}{1-2^{-p}}
    &< z+2^{\emin-p} \\
\notag
  z + 2^{\emin-p} -2^{1-p+ \expf{z}}
    &< (z+2^{\emin-p})(1-2^{-p}) \\
\notag
  z + 2^{\emin-p} -2^{ 1-p+\expf{z}}
    &< z+2^{\emin-p}-z2^{-p}-2^{\emin-2p} \\
\notag
  -2^{1-p+ \expf{z}}
    &< -z 2^{-p}-2^{\emin-2p} \\
\notag
  -2^{1-p+ \expf{z}}
    &< -2^{\expf{z} -p}-2^{\emin-2p}  \\
\label{eq:mu-fdiv-satisfies-properties-sub13}
  2^{\emin-2p}
    &< 2^{-p + \expf{z}} \\
\label{eq:mu-fdiv-satisfies-properties-sub14}
  2^{\emin-2p}
    &< 2^{\emin-2p+1},
\end{align}
where~\eqref{eq:mu-fdiv-satisfies-properties-sub08}
is equivalent to~\eqref{eq:mu-fdiv-satisfies-properties-sub09}
because of~\eqref{eq:mu-fdiv-satisfies-properties-sub10},
and~\eqref{eq:mu-fdiv-satisfies-properties-sub13}
is equivalent to~\eqref{eq:mu-fdiv-satisfies-properties-sub14}
because $\expf{z}\geq \emin-p+1$.
As for the equivalence between~\eqref{eq:mu-fdiv-satisfies-properties-sub09}
and~\eqref{eq:mu-fdiv-satisfies-properties-sub12}, note that
the exponent of $z 2^{\emax+1} + 2^{\emin+1-p+\emax}$ is $\emax + \expf{z} + 1$,
hence $\ferrup{z 2^{\emax+1} + 2^{\emin+1-p+\emax}} = 2^{1-p}2^{ \emax+\expf{z}+1}$.
Finally, assuming $p > 1$, the last inequality holds.

In order to prove~\eqref{eq:ulpmmax2}, we
need to prove that $\ulpmmax{\fdiv}(z)^+ \fdiv \fmax > z$.
By Definition~\ref{def:round-to-nearest-tail-to-even},
it suffices to prove that
$\ulpmmax{\fdiv}(z)^+ / \fmax > z + \ferrdown{z}/2$.
In this case we have that
\begin{align}
\label{eq:mu-fdiv-satisfies-properties-sub15}
  \frac{\ulpmmax{\fdiv}(z)^+}{\fmax}
    &= \frac{ z 2^{\emax+1} + 2^{\emin+1-p+\emax}}{\fmax} \\
\notag
    &= \frac{z+ 2^{-p+\emin}}{1-2^{-p}} \\
\notag
    &> z+2^{\emin-p} \\
\notag
    &= z+\ferrdown{z}/2,
\end{align}
where~\eqref{eq:mu-fdiv-satisfies-properties-sub15} holds
because of~\eqref{eq:mu-fdiv-satisfies-properties-sub10}.
Hence
\(
  \ulpmmax{\fdiv}(z)^+\fdiv \fmax
    = \bigroundnear{\ulpmmax{\fdiv}(z)^+/ \fmax}
   \geq z^+,
\)
which proves~\eqref{eq:ulpmmax2}.

\end{description}

For $z < 0$ we can reason as before choosing $y = -\fmax$.
\Halmos
\endproof

\begin{theorem}
$\ulpmmax{\fdiv}$ is well-defined and
satisfies~\eqref{eq:ulpmmax1} and~\eqref{eq:ulpmmax2}.
\end{theorem}
\proof{Proof.}
Immediate from Lemma~\ref{lem:mu-fdiv-satisfies-properties}
and Lemma~\ref{lem:mu-fdiv-satisfies-properties-sub}.
\Halmos
\endproof

\begin{proposition}
Let $z \in \Fset_{\fdiv}$ be nonzero.
If $z > 0$, then $\ulpmmax{\fdiv}(z^+) \geq \ulpmmax{\fdiv}(z)$;
on the other hand,
if $z < 0$, then $\ulpmmax{\fdiv}(z^-) \geq \ulpmmax{\fdiv}(z)$.
\end{proposition}
\proof{Proof.}
Assume for simplicity that $z > 0$.
We need to investigate the following critical cases on $z$:
\begin{description}

\item [$0 < z <(\fnormin)^-$
       and $z = \float{1}{\expf{z}}$ with $\expf{z}<\emin-1:$]
In this case the result holds because
\begin{align*}
  \ulpmmax{\fdiv}(z^+)
    &= \bigl((z + 2^{1-p+\emin}) \fmul \fmax\bigr) \fadd 2^q \\
    &\geq (z \fmul \fmax) \fadd 2^q \\
    &\geq \bigl((z \fmul \fmax) \fadd 2^q\bigr)^- \\
    &= \ulpmmax{\fdiv}(z).
\end{align*}

\item [$z = \float{1.1\ldots1}{\expf{z}}$ with $\expf{z}<\emin-2:$]
We need to show that $\ulpmmax{\fdiv}(z^+)\geq \ulpmmax{\fdiv}(z)$.
Note that,  by Definition~\ref{def:mu-fdiv}, we have%
\begin{align}
\label{eq-monotonicity-fdiv-0}
  \ulpmmax{\fdiv}(z^+)
    &= (\bigl((z^+ \fmul \fmax\bigr) \fadd 2^q)^- \\
\label{eq-monotonicity-fdiv-1}
    &= ((z^+ 2^{\emax+1}) + 2^q)^- \\
\label{eq-monotonicity-fdiv-2}
    &= (z^+ 2^{\emax+1}) + 2^q-2^{1-p+\emax+\expf{z}+2} \\
\label{eq-monotonicity-fdiv-3}
    &= (z + 2^{1-p+\emin}) 2^{\emax+1} + 2^q-2^{1-p+\emax+\expf{z}+2} \\
\notag
    &= (z 2^{\emax+1} + 2^{1-p+\emax+\emin} ) + 2^q-2^{1-p+\emax+\expf{z}+2}) \\
\label{eq-monotonicity-fdiv-5}
    &> z 2^{\emax+1}+ 2^q \\
\notag
    &> (z 2^{\emax+1})^- + 2^q \\
\label{eq-monotonicity-fdiv-6}
  &=\ulpmmax{\fdiv}(z),
\end{align}
where~\eqref{eq-monotonicity-fdiv-0} holds by Definition \ref{def:mu-fdiv}
and~\eqref{eq-monotonicity-fdiv-1}
holds by~\eqref{eq:mu-fdiv-satisfies-properties-sub10}.
In order to show that~\eqref{eq-monotonicity-fdiv-2} holds,
note that the exponent of $z^+ 2^{\emax+1} + 2^{\emin+1-p+\emax}$
is $\emax + \expf{z} + 2$;
hence $\ferrup{z^+ 2^{\emax+1} + 2^{\emin+1-p+\emax}} = 2^{1-p}2^{\emax+\expf{z}+2}$.
Eq.~\eqref{eq-monotonicity-fdiv-3} holds because $z$ is subnormal,
hence $\ferrdown{z} = \fmin$,
whereas~\eqref{eq-monotonicity-fdiv-5} holds
because we have assumed $\expf{z} < \emin - 2$.
Finally, \eqref{eq-monotonicity-fdiv-6} holds because
of~\eqref{eq:fadd-2powq-is-exact-3}.

\item [$z = (\fnormin)^-:$]
Namely, in this case,
$z = (2 - 2^{2-p}) 2^{\emin-1}$ and $z^+ = 2^{\emin}$.
We can thus write
\begin{align}
\notag
  \ulpmmax{\fdiv}(z)
    &= (z\fmul \fmax) + 2^q \\
\label{eq-monotonicity-fdiv-7}
    &= (z2^{\emax+1})^- + 2^q \\
\notag
    &= (2-2^{2-p}) 2^{\expf{z} + \emax+1} - 2^{1-p+\expf{z}+\emax+1}+2^q \\
\notag
    &= (2-2^{2-p}) 2^{\emin + \emax} - 2^q + 2^q \\
\notag
    &= (2-2^{2-p}) 2^{\emin + \emax} \\
\notag
    &= (2-2^{2-p}) 2^{\emax} 2^{\emin} \\
\notag
    &= (2-2^{2-p}) 2^{\emax} \fmul z^+ \\
\notag
    &< (2-2^{1-p}) 2^{\emax} \fmul z^+ \\
\notag
    &= \ulpmmax{\fdiv}(z^+),
\end{align}
where~\eqref{eq-monotonicity-fdiv-7} is justified
by~\eqref{eq:mu-fdiv-satisfies-properties-sub00}.

\end{description}
Hence, taking into account the monotonicity of $\fmul$ and $\fadd$,
we can conclude that $\ulpmmax{\fdiv}$ is monotone.
\Halmos
\endproof

In order to prove Theorem~\ref{th:upperbound} we need the following
intermediate result.

\begin{lemma}
\label{lem:dis-upperdiv}
Let $z \in \Fsetsub_{p,\emax}$ be such that $1^+ < |z| \leq \fmax$.
Then $\fmax \fdiv \upperdiv(z) < |z|$.
\end{lemma}
\proof{Proof.}
By Definition~\ref{def:upperbound_div}, we have to prove that
$\fmax \fdiv \bigl(\fmax \fdiv |z|^{-\,-}\bigr) < |z|$
for $1^+ < |z| \leq \fmax$.
Assume by simplicity that $z > 0$.
The case $z < 0$ can be obtained by considering the absolute value of $z$.

We have the following cases on $z$:
\begin{description}

\item[$z = \float{1.0 \cdots 01}{\expf{z}}:$]
In this case, since $1^+ < |z|$, then  $\expf{z}>0$.
We have $z^{-\,-}= (2 - 2^{1-p}) 2^{\expf{z}-1}$ and thus
\begin{align*}
  \fmax \fdiv z^{-\,-}
    &= \scaleroundnear{\frac{(2 - 2^{1-p}) 2^{\emax}}{(2 - 2^{1-p}) 2^{\expf{z}-1}}} \\
    &= \roundnear{2^{\emax-\expf{z}+1}} \\
    &= 2^{\emax-\expf{z}+1}, \\
\intertext{%
therefore
}
  \fmax \fdiv \bigl(\fmax\fdiv |z|^{-\,-}\bigr)
    &= \scaleroundnear{\frac{(2 - 2^{1-p}) 2^{\emax}}{2^{\emax-\expf{z}+1}}} \\
    &= \bigroundnear{(2 - 2^{1-p}) 2^{\expf{z}-1}} \\
    &= (2 - 2^{1-p}) 2^{\expf{z}-1} \\
    &< \float{1.0 \cdots 01}{\expf{z}} \\
    &= z.
\end{align*}

\item[$z = \float{1.0 \cdots 00}{\expf{z}}:$]
We have $z^{-\,-} = (2 - 2^{2-p}) 2^{\expf{z}-1}$ and thus
\begin{align}
\notag
  \fmax \fdiv z^{-\,-}
    &= \scaleroundnear{\frac{(2 - 2^{1-p}) 2^{\emax}}{(2 - 2^{2-p}) 2^{\expf{z}-1}}} \\
\label{lem:dis-upperdiv:eq0}
    &= \scaleroundnear{\frac{2 - 2^{2-p} + 2^{1-p}}{2 - 2^{2-p}}} 2^{\emax-\expf{z}+1} \\
\notag
    &= \scaleroundnear{1 + \frac{2^{1-p}}{2 - 2^{2-p}}} 2^{\emax-\expf{z}+1} \\
    &= (1+2^{1-p})2^{\emax-\expf{z}+1}.\label{lem:dis-upperdiv:eq1}
\end{align}
Eq.~\eqref{lem:dis-upperdiv:eq0} holds because the multiplication
by $2^{\emax-\expf{z} +1}$ can give rise neither to an
overflow --- because $z \geq 2$ and thus $\fmax \fdiv z^{-\,-} < \fmax$ ---
nor to an underflow ---
because $z \leq 2^{\emax}$ and thus $\fmax \fdiv z^{-\,-} \gg \fmin$.
Moreover, Eq.~\eqref{lem:dis-upperdiv:eq1} holds because
\[
  1 + \frac{2^{1-p}}{2 - 2^{2-p}}
    < 1 + 2^{-p} + 2^{1-p}
    = 1^+ + \ferrdown{1^+}/2
\]
and
\[
  1 + \frac{2^{1-p}}{2 - 2^{2-p}}
    > 1 + \frac{2^{1-p}}{2}
    = 1 + 2^{-p}
    = 1 + \ferrdown{1}/2
    = 1^+ - \ferrup{1^+} /2.
\]
Hence, by Definition~\ref{def:round-to-nearest-tail-to-even},
$\bigroundnear{1+\frac{2^{1-p}}{2 - 2^{2-p}}} = 1^+ = 1+2^{1-p}$.
We can thus write
\begin{align*}
  \fmax \fdiv \bigl(\fmax \fdiv |z|^{-\,-}\bigr)
    &=
      \scaleroundnear{\frac{(2 - 2^{1-p}) 2^{\emax}}
                           {(1+2^{1-p}) 2^{\emax-\expf{z}+1}}} \\
    &\leq \bigroundnear{(2 - 2^{1-p}) 2^{\expf{z}-1}} \\
    &= (2 - 2^{1-p})  2^{\expf{z}-1} \\
    &< \float{1.0 \cdots 00}{\expf{z}} \\
    &= z.
\end{align*}

\item[\text{$z \neq \float{1.0 \cdots 0}{\expf{z}}$ and
            $z \neq \float{1.0 \cdots0 1}{\expf{z}}:$}]
In this case
We have $z = \float{m}{\expf{z}}$ with $1 + 2^{2-p} \leq m \leq (2 - 2^{1-p})$
and thus
\begin{align}
\notag
  \fmax \fdiv \bigl(\fmax\fdiv |z|^{-\,-}\bigr)
    &= \scaleroundnear{
         \frac{(2 - 2^{1-p})2^{\emax}}
              {\scaleroundnear{\frac{(2 - 2^{1-p}) 2^{\emax}}
                                    {(m - 2^{2-p}) 2^{\expf{z}}}
               }
              }
       } \\
\label{lem:dis-upperdiv:eq2}
  &= \scaleroundnear{
       \frac{(2 - 2^{1-p}) 2^{\emax}}
            {\scaleroundnear{\frac{2 - 2^{1-p}}
                            {m - 2^{2-p}}
             } 2^{\emax-\expf{z}}
            }
     } \\
\label{lem:dis-upperdiv:eq3}
  &= \scaleroundnear{
       \frac{2 - 2^{1-p}}
            {\scaleroundnear{\frac{2 - 2^{1-p}}
                                  {m - 2^{2-p}}}}
     } 2^{\expf{z}},
\end{align}
where~\eqref{lem:dis-upperdiv:eq2} and~\eqref{lem:dis-upperdiv:eq3} hold
because the
multiplications by $2^{\emax-\expf{z} }$ and by $2^{\expf{z} }$,
respectively, can give rise neither to an overflow nor to an underflow,
since $m \geq 1 + 2^{2-p}$.
We are thus left to prove that
\begin{equation}
\label{eq:dis-significand}
  \scaleroundnear{\frac{2 - 2^{1-p}}
                    {\scaleroundnear{\frac{2 - 2^{1-p}}
                                          {m - 2^{2-p}}
                     }}
  } < m
\end{equation}
subject to $1 + 2^{2-p} \leq m \leq 2 - 2^{1-p}$.
We distinguish two cases on the value of
$\bigroundnear{\frac{2 - 2^{1-p}}{m - 2^{2-p}}}$:
\begin{description}

\item[$\bigroundnear{\frac{2 - 2^{1-p}}{m - 2^{2-p}}} \geq \frac{2 - 2^{1-p}}{m - 2^{2-p}}:$]
Thus
\begin{align*}
  \scaleroundnear{\frac{2 - 2^{1-p}}
                    {\scaleroundnear{\frac{2 - 2^{1-p}}
                                          {m - 2^{2-p}}
                     }}
  }
    &\leq \scaleroundnear{\frac{2 - 2^{1-p}}
                               {\frac{2 - 2^{1-p}}{m - 2^{2-p}}}
                         } \\
    &= \roundnear{m - 2^{2-p}} \\
    &= m - 2^{2-p} \\
    &< m,
\end{align*}
and~\eqref{eq:dis-significand} holds.

\item[$\roundnear{\frac{2 - 2^{1-p}} {m - 2^{2-p}}}< \frac{2 - 2^{1-p}} {m - 2^{2-p}}:$]
By Definition~\ref{def:round-to-nearest-tail-to-even} we know that
\begin{equation}
\label{lem:dis-upperdiv:0}
  \scaleroundnear{\frac{2 - 2^{1-p}}{m - 2^{2-p}}}
    + \frac{\ferrdown{\frac{2 - 2^{1-p}}{m - 2^{2-p}}}}{2}
      > \frac{2 - 2^{1-p}}{m - 2^{2-p}}.
\end{equation}
Since $\ferrdown{\frac{2 - 2^{1-p}}{m - 2^{2-p}}} = 2^{1-p}$,
from~\eqref{lem:dis-upperdiv:0} we obtain
\begin{equation}
\label{lem:dis-upperdiv:1}
  \bigroundnear{\frac{2 - 2^{1-p}}{m - 2^{2-p}}}
    \geq
      \frac{2 - 2^{1-p}}{m - 2^{2-p}} - 2^{-p}.
\end{equation}

Hence, applying~\eqref{lem:dis-upperdiv:1}, we have:
\begin{align}
  \scaleroundnear{\frac{2 - 2^{1-p}}
                       {\scaleroundnear{\frac{2 - 2^{1-p}}{m - 2^{2-p}}
                        }}
  }
\notag
    &\leq
      \scaleroundnear{\frac{2 - 2^{1-p}}
                        {\frac{2 - 2^{1-p}}{m - 2^{2-p}} - 2^{-p}}
      } \\
\notag
    &=
      \scaleroundnear{\frac{(2 - 2^{1-p}) (m - 2^{2-p})}
                           {2 - 2^{1-p} -2^{-p}(m - 2^{2-p})}
      } \\
\label{eq:dis-eq-significand1}
    &\leq
      \scaleroundnear{\frac{(2 - 2^{1-p}) (m - 2^{2-p})}
                           {2 - 2^{1-p} -2^{-p}(2)}
      } \\
\notag
    &=
      \scaleroundnear{\frac{(2 - 2^{1-p}) m - 2^{3-p} +2^{3-2p}}
                        {2 - 2^{1-p} -2^{1-p}}
      } \\
\notag
    &=
      \scaleroundnear{\frac{(2 - 2^{1-p}) m - 2^{3-p} +2^{3-2p}}
                        {2 - 2^{2-p} }
      } \\
\notag
    &=
      \scaleroundnear{\frac{(2 - 2^{2-p}+2^{1-p}) m - 2^{3-p} +2^{3-2p}}
                        {2 - 2^{2-p}}
      } \\
\notag
    &=
      \scaleroundnear{m + \frac{2^{1-p}m}{2 - 2^{2-p}}
                        - \frac{2^{3-p}}{2 - 2^{2-p}}
                        + \frac{2^{3-2p}} {2 - 2^{2-p}}
      } \\
    &\leq
      \bigroundnear{m + 2^{1-p} m - 2^{3-p} + 2^{3-2p}}
        \label{eq:dis-eq-significand2} \\
    &\leq
      \bigroundnear{m + 2^{2-p} - 2^{3-p} + 2^{3-2p}}
        \label{eq:dis-eq-significand3} \\
    &=
      \bigroundnear{m + 2^{2-p}(1-2) + 2^{3-2p}}
        \notag \\
    &=
      \bigroundnear{m - 2^{2-p}+  2^{3-2p}}
        \notag \\
    &\leq
      \roundnear{m- 2^{1-p}} \notag\\
    &=
      m^- \notag\\
    &<
      m.
        \notag
 \end{align}
Note that~\eqref{eq:dis-eq-significand1} and~\eqref{eq:dis-eq-significand3}
hold because $m \leq (2 - 2^{1-p}) < 2$,
whereas~\eqref{eq:dis-eq-significand2} holds because $(2 - 2^{1-p}) > 1$.
\end{description}
In any case \eqref{eq:dis-significand} holds and this concludes the proof.
\end{description}
\Halmos
\endproof

\begin{theorem}
Let $\Fset''_{\fdiv} = \Fsetsub_{p,\emax}$
and $\bar{\Fset}''_{\fdiv} = \Fset^+_{p,\emax}$.
Let
\(
  \fund{\ulpmmaxp{\fdiv}}{\Fset''_{\mathord{\fdiv}}}
                         {\bar{\Fset}''_{\mathord{\fdiv}}}
\)
be a function satisfying~\eqref{eq:ulpmmaxp1new}.
Then, for $0 < |z| \leq 1^+$ or $z = +\infty$,
$\ulpmmaxp{\fdiv}(z) \leq \upperdiv(z)$;
moreover, for $1^+ < |z| \leq \fmax$,
$\ulpmmaxp{\fdiv}(z) < \upperdiv(z)$.
\end{theorem}

\proof{Proof.}
Recall that, by definition, $\ulpmmaxp{\fdiv}$ satisfies~\eqref{eq:ulpmmaxp1new}
and, thus, for each $z \in \Fset''_{\fdiv} \setminus \{ -0, +0, -\infty \}$
there exists $x \in \bar{\Fset}''_{\mathord{\fdiv}}$
such that $x \fdiv \ulpmmaxp{\fdiv}(z) = z$.
There are two cases on $z$:
\begin{description}

\item[\text{$z = +\infty$ or $0 < |z| \leq 1^+:$}]
As we have $\upperdiv(z) = \fmax$, we just have to show that
$\ulpmmaxp{\fdiv}(z) \neq +\infty$.
Indeed, if $\ulpmmaxp{\fdiv}(z) = +\infty$,
then $x \fdiv \ulpmmaxp{\fdiv}(z)$ can only give $\pm 0$
(if $-\fmax \leq x \leq \fmax$) or NaN (if $x = \pm\infty$),
so that~\eqref{eq:ulpmmaxp1new} cannot be satisfied.

\item[$ < |z| \leq \fmax:$]
Assume, towards a contradiction,
that $\upperdiv(z) \leq \ulpmmaxp{\fdiv}(z)$
for some $z$ such that $1^+ < |z| \leq \fmax$.
Hence, as $\fdiv$ is antitone in its second argument,
$\fmax \fdiv \ulpmmaxp{\fdiv}(z) \leq \fmax\fdiv \upperdiv(z)$.
By Lemma~\ref{lem:dis-upperdiv}, $\fmax \fdiv \upperdiv(z) < z$,
hence we also have $\fmax \fdiv \ulpmmaxp{\fdiv}(z) < z$.
This contradicts the hypothesis that $\ulpmmaxp{\fdiv}$
satisfies~\eqref{eq:ulpmmaxp1new}.
In fact, as $\fdiv$ is monotone in its first argument,
$x \fdiv \ulpmmaxp{\fdiv}(z) = z$ would require $x > \fmax$
or, equivalently $x = +\infty$.
But $+\infty \fdiv \ulpmmaxp{\fdiv}(z)$ is either equal to $\pm\infty$,
if $\ulpmmaxp{\fdiv}(z)\leq \fmax$, or NaN, if $\ulpmmaxp{\fdiv}(z) > \fmax$.
This concludes the proof.
\end{description}
\Halmos
\endproof
\end{APPENDIX}
}{ 
} 

\end{document}